\documentclass[11pt,a4paper]{article}

\usepackage[T1]{fontenc}        %
\usepackage{times,latexsym}     %
\usepackage{microtype}          %
\usepackage{inconsolata}        %

\usepackage{amsmath}
\usepackage{algorithm}
\usepackage{algpseudocode}

\usepackage[acceptedWithA]{tacl2021v1}

\usepackage{graphicx}
\usepackage{subcaption}         %
\usepackage{tikz}               %
\usetikzlibrary{patterns}       %
\usepackage{pgfplots}
\pgfplotsset{compat=1.18}       %

\usepackage{url}                %

\usepackage{natbib}

\usepackage{pifont}             %

\usepackage{booktabs}           %
\usepackage{listings}           %

\usepackage{xcolor}             %
\definecolor{color1}{HTML}{FF0000} %
\definecolor{color2}{HTML}{00FF00} %
\definecolor{color3}{HTML}{0000FF} %
\definecolor{color4}{HTML}{808080} %
\definecolor{bblue}{HTML}{4F81BD}
\definecolor{rred}{HTML}{C0504D}
\definecolor{ggreen}{HTML}{9BBB59}
\definecolor{ppurple}{HTML}{9F4C7C}

\usepackage[most]{tcolorbox}    %

\usepackage[normalem]{ulem}     %
\newcommand{\magentauline}[1]{{\color{magenta}\uline{{\color{black}#1}}}}

\usepackage{xspace,mfirstuc,tabulary}

\newif\iftaclinstructions
\taclinstructionsfalse %
\iftaclinstructions
  \renewcommand{\confidential}{}
  \renewcommand{\anonsubtext}{(No author info supplied here, for consistency with
  TACL-submission anonymization requirements)}
  \newcommand{\instr}
\fi

\iftaclpubformat

\else

\fi

\newcommand{\secsymbol}{\S}     %

\title{\textit{NERsocial}: Efficient Named Entity Recognition Dataset Construction for Human-Robot Interaction Utilizing \textit{RapidNER}}

\author{
        Jesse Atuhurra, Hidetaka Kamigaito, Hiroki Ouchi, Hiroyuki Shindo, Taro Watanabe 
        \\ 
        Division of Information Science, NAIST, Japan 
        \\ 
        \texttt{ \{atuhurra.jesse.ag2, kamigaito.h, hiroki.ouchi, shindo,  taro.watanabe\}@naist.ac.jp} 
        }
\date{}

\begin{document}
\maketitle
\begin{abstract}
Adapting named entity recognition (NER) methods to new domains poses significant challenges. We introduce \textit{RapidNER}, a framework designed for the rapid deployment of NER systems through efficient dataset construction. \textit{RapidNER} operates through three key steps: (1) extracting domain-specific sub-graphs and triples from a general knowledge graph, (2) collecting and leveraging texts from various sources to build the \textit{NERsocial} dataset, which focuses on entities typical in human-robot interaction, and (3) implementing an annotation scheme using Elasticsearch (ES) to enhance efficiency. \textit{NERsocial}, validated by human annotators, includes six entity types, 153K tokens, and 99.4K sentences, demonstrating \textit{RapidNER}'s capability to expedite dataset creation.
\end{abstract}
\section{Introduction}
\label{sec:Introduction}
The natural language processing (NLP) field has grown substantially in recent years, surpassing expectations and creating many new applications. NLP applications are manifestations of NLP tasks, such as information extraction, text generation, language modeling, and the like. 

Yet, despite all the progress, adapting existing information extraction systems for NER to new domains and entity types remains a challenge, a fact that is compounded by the need to develop new datasets representative of the new domain, followed by fine-tuning NER classifiers to correctly detect new entity types in the new domain. 
\begin{figure}[!t]
\centering
\includegraphics[width=7.5cm]{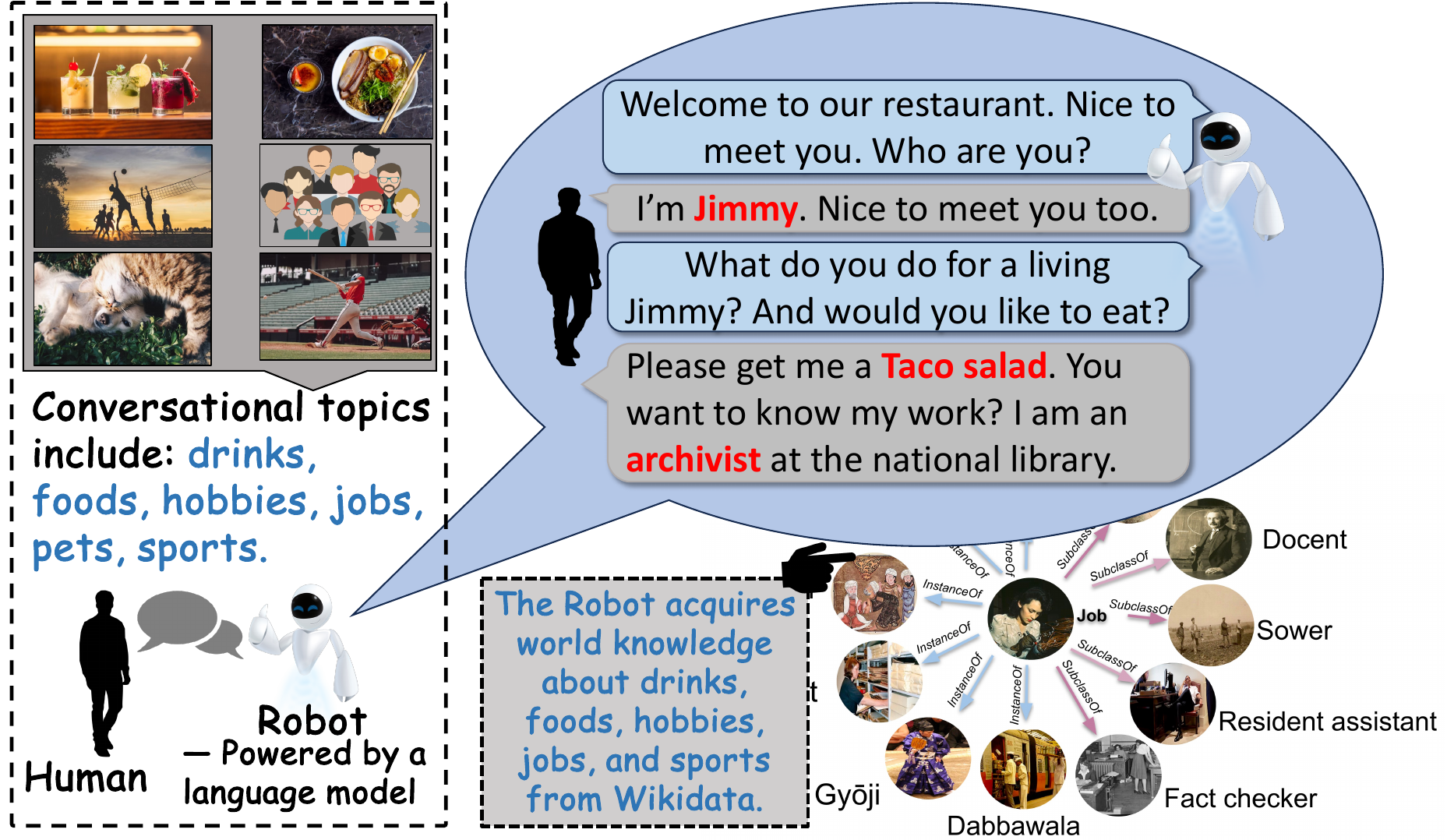}
\caption{Our NER dataset aims to support dialogue between humans and robots. New entity types are \textbf{Drink, Food, Hobby, Job, Pet, Sport}. We utilize Wikidata to acquire information about entity types.}
\label{fig:introJob}
\end{figure}

\begin{figure*}[!t]
\centering
\includegraphics[width=0.85\textwidth]{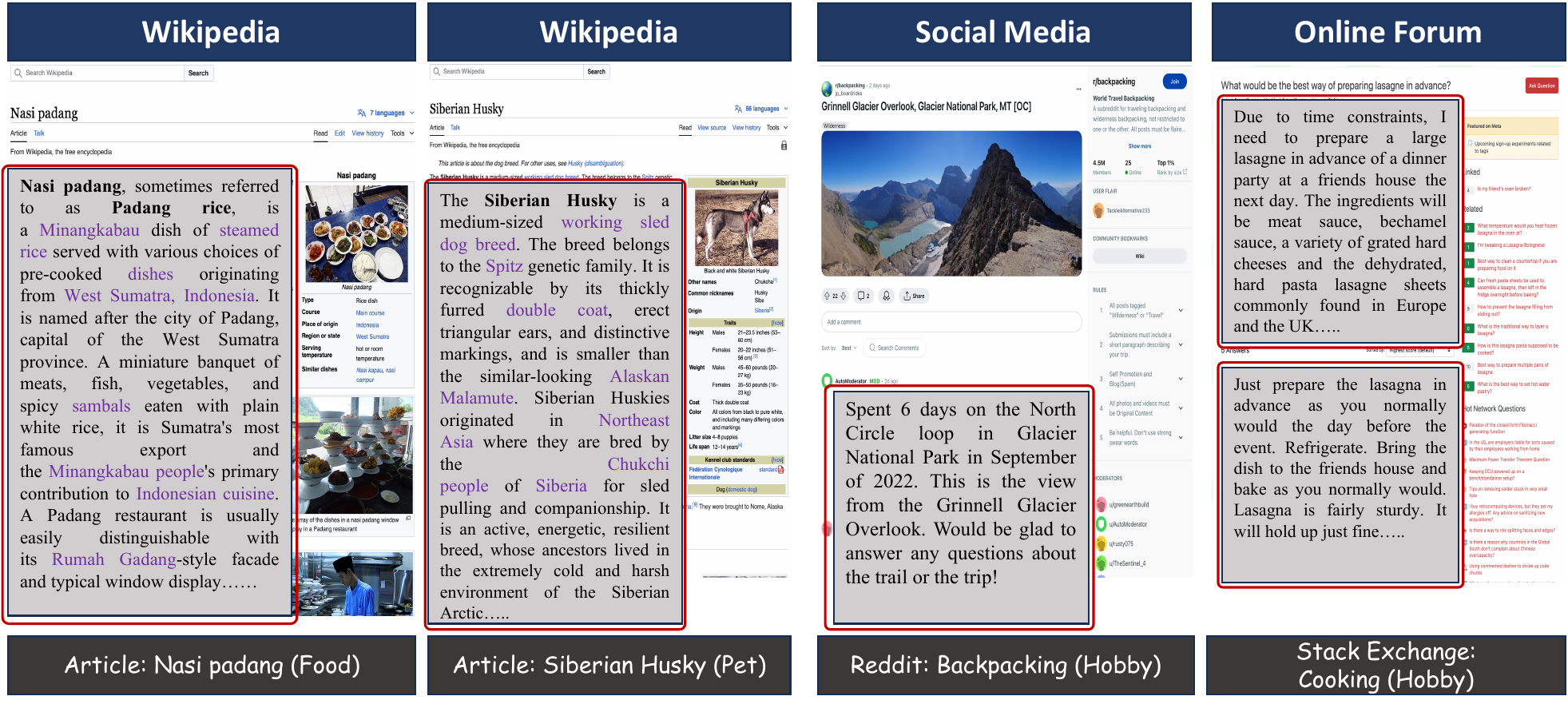}
\caption{We collected the texts from three sources: \textbf{Wikipedia}, \textbf{online forums} (Stack Exchange) and \textbf{social media} (Reddit). The red box indicates sections containing the texts that we are interested in, for each textual source.}
\label{fig:SourcesOfTexts}
\end{figure*}

Recognizing the need to create NER datasets for new domains and entity types efficiently, we make two succinct contributions: 1) We propose a framework \emph{RapidNER} to annotate NER datasets by utilizing off-the-shelf tools. Specifically, we exploit the search functions inside \textit{ES} and successfully develop an annotation method for NER data. 2) We focus on human-robot interaction (HRI) applications and develop a new dataset named \emph{NERsocial} comprising the entity types that are suitable for interaction between humans and a social robot\footnote{One use-case is that the robot's been deployed as a tutor for kids aged 10 years and below; when the teacher is absent.} (Figure \ref{fig:introJob}). Consequently, we sought texts that represent the natural conversational style embedded in human dialogue. Our textual sources include social media (Reddit) and online forums (Stack Exchange). Aiming for diversity in texts, we included Wikipedia as a complimentary textual source, which provides a rich contextual narrative for all entities. Figure \ref{fig:SourcesOfTexts} shows all the textual sources.

\emph{NERsocial} is not the first attempt by NER researchers to design new specifications of NE labels to develop NER datasets suited for new domains.  We follow the footsteps of domain-specific and unconventional efforts, such as \citet{epure2023}, to create our NER dataset \emph{NERsocial}. \emph{NERsocial} consists of six entity types\footnote{The six entity types were sufficient to build the first prototype of a social robot interacting with young kids. More entity types are iteratively added based on use-case.}  \textbf{drinks, foods, hobbies, jobs, pets, and sports} which are typical of \textit{social interactions} in HRI, and Figure \ref{fig:introJob} shows an example of a human-robot dialogue. While creating \emph{NERsocial}, we needed to \textit{acquire} knowledge about each entity type. Consequently, we sought to answer the research question \textit{how can we quickly gather entity information relevant to unseen entities, such as pets?} Hence, we chose the KG approach to gather such entity knowledge in the form of KG triples which guided the development of our NER dataset. This process is summarized in Figure~\ref{fig:IntroDatasetProcess}. In addition, KG triples enabled us to solve another problem, that is, \textit{how to increase the coverage and diversity of each NE?}  Concretely, KG triples guided the extraction of a large number of diverse pages (articles) from Wikipedia. In turn, the pages were the source of sentences used to construct the dataset coupled with the texts gathered from social media and online forums. The last challenge we faced was: \textit{how to annotate the thousands of sentences with less human effort?} We utilized the search functions inside Elasticsearch to develop an annotation scheme that took only a few seconds to \textit{mark} spans of entity mentions, \secsymbol\ref{subsec: Annotation Design}.
\section{Related work}
\label{sec:Related work}
Our work focuses on entities specific to HRI and unseen in previous NER datasets. Popular NE include: \textit{Person, Location, Organization, Product, Event, Building, Product}, among others; and NER datasets include: CoNLL2003~\cite{tjong-kim-sang-2002-introduction}, WikiGold~\cite{balasuriya-etal-2009-named}, OntoNotes~\cite{weischedel2013ontonotes}, WNUT2017~\cite{derczynski-etal-2017-results}, FEWNERD~\cite{ding-etal-2021-nerd}, and I2B2~\cite{Stubbs2015AnnotatingLC}. Though the above entity types and datasets are sufficient for many applications, utilizing the same datasets for unique domains such as HRI is difficult because HRI includes many ``rare entities'' such as \textit{pets, leisure activities, popular cuisines, weather\&seasons, etc.}. Due to this domain transfer problem, \citet{epure2023} developed an NER dataset for music recommendations in a conversational assistant. 
Researchers have also developed many NER datasets to support a plethora of unique applications. Exemplar NER applications and datasets include: biomedical applications, GENIA \cite{kim-etal-2003-genia}, NCBI \cite{dogan2014ncbi}, conversational music recommendation, MusicRecoNER \cite{epure2023}, disaster response systems, HarveyNER \cite{chen-etal-2022-crossroads}, taxonomic entities in biology, TaxoNERD \cite{le-guillarme-2022-taxonerd}, inter alia.  Evidently, in all these applications, it is common practice to develop a dataset characteristic of entity types relevant to the application, followed by fine-tuning a deep learning-based NER classifier to recognize entities. The common sources for texts include Wikipedia, news articles, social media posts TwitterNER7 \cite{ushio2022named},  online forum posts and comments from video streaming platforms like YouTube, WNUT \cite{derczynski-etal-2017-results}.
Datasets created from Wikipedia, social media and posts from online forums are relevant to our work. For example, WikiGold \cite{balasuriya-etal-2009-named}, WikiANN \cite{pan-etal-2017-cross}, FewRel 2.0 \cite{gao-etal-2019-fewrel} were constructed from either Wikipedia or Wikidata. However, WNUT is most similar in nature to our dataset. Hence, we compared WNUT to our dataset in subsequent sections.
Lastly, we have not yet found any similar works that employed ES to annotate NER datasets.
\section{Overview of \emph{RapidNER} Framework}
\label{sec:Overview of the framework}
\begin{algorithm}[!t]
\footnotesize
\caption{: RapidNER Framework}
\begin{algorithmic}[1]
\State \textbf{Input:} Knowledge Graph $G$, Set of Data sources $D$, \\
Set of Conversational Topics $C$, \\ Annotation Tool $A$
\State \textbf{Output:} NER Dataset $\mathcal{D}_{NER}$

\For{each data source $d \in D$}
    \State $\mathcal{D}_{data} \gets \emptyset$ \Comment{Initialize dataset}
    \For{each topic $c \in C$}
        \State $\mathcal{T}_c \gets \text{Gather triples in } G \text{ for topic } c$ \Comment{Extract sub-graph and triples}
        \State $\mathcal{H}_t \gets \text{Extract head entities from } \mathcal{T}_c$ \Comment{Create list of hypernyms}
        \State $\mathcal{W}_t \gets \text{Get Wiki articles for titles in } \mathcal{H}_t$ \Comment{Gather articles from Wikipedia }
        \State $\mathcal{S}_t \gets \text{Extract sentences from } \mathcal{W}_t$
        \For{each sentence $s \in \mathcal{S}_t$}
            \State $s_{annotated} \gets A.\text{annotate}(s, \mathcal{H}_t)$ %
            \State $s_{BIO} \gets \text{Convert spans in } s_{annotated}$
            \Statex \text{ to BIO tags} %
            \State $\mathcal{D}_{data} \gets \mathcal{D}_{data} \cup \{(s, s_{BIO})\}$
        \EndFor
    \EndFor
    \State $\mathcal{D}_{NER} \gets \mathcal{D}_{NER} \cup \mathcal{D}_{data}$
\EndFor

\State \textbf{return} $\mathcal{D}_{NER}$
\end{algorithmic}
\label{Algorithm:RapidNER framework}
\end{algorithm}
To address the problem of scarcity of labeled NER data for applications such as human-robot interaction (or any other application), we developed a framework to quickly collect and annotate texts, resulting in a new NER dataset. The proposed framework is called \emph{RapidNER}, comprising three main steps as described in Algorithm \ref{Algorithm:RapidNER framework}. 

\emph{First.} We select the knowledge graph, $G$; the source of textual data, that is, the set of all data sources $D$ from which to collect the texts. In this work, we mainly focused on HRI applications. Hence, we further defined a precise set of conversational topics, $C$, for HRI. $C$ represents the NE. 

\begin{figure*}[!t]
\centering
\includegraphics[width=0.85\linewidth]{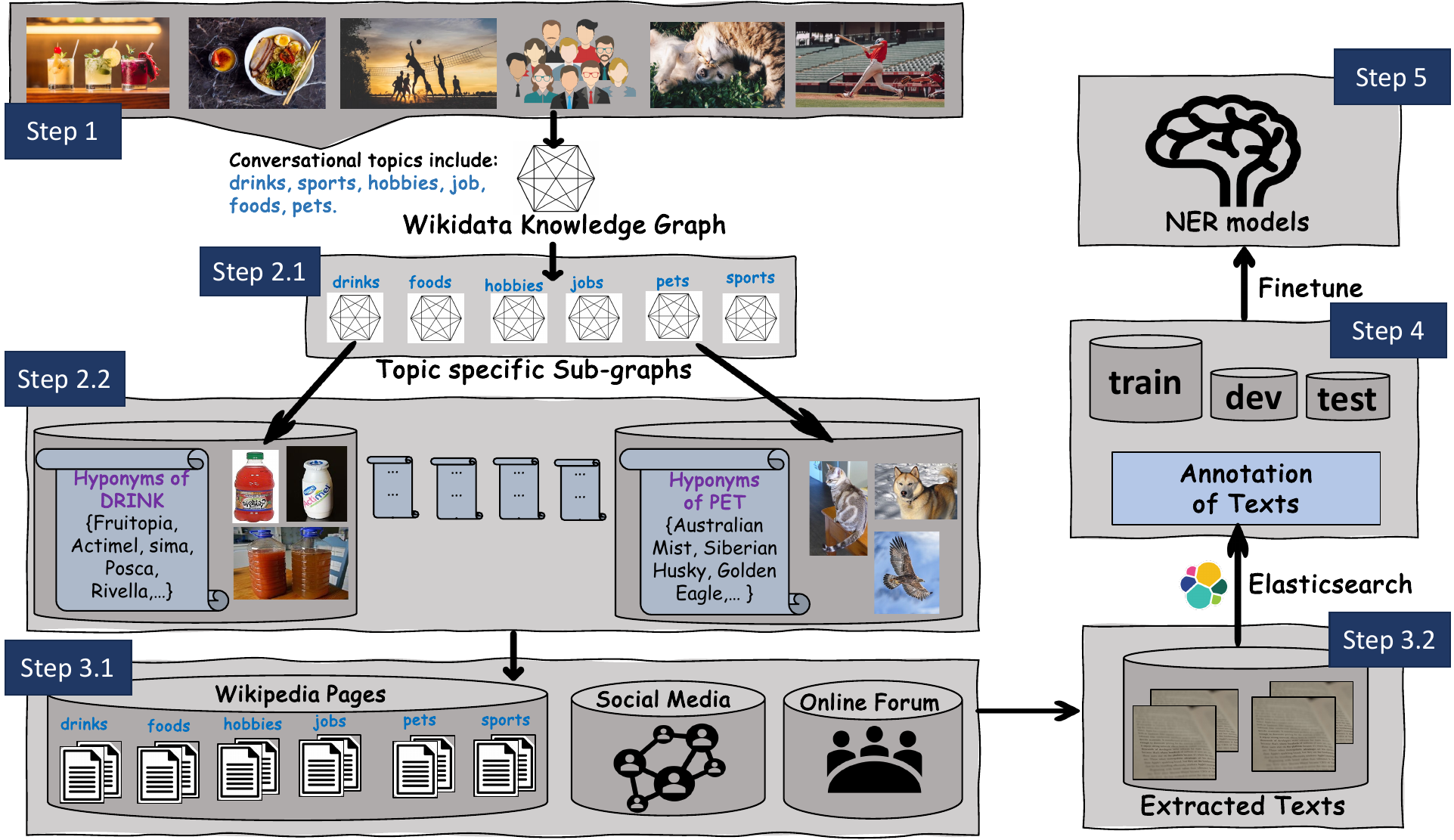}
\caption{
The construction process of \emph{NERsocial}. We gathered millions of triples from Wikidata and used the triples to collect Wikipedia articles. For each Wikipedia article, we extracted paragraphs from the \texttt{introduction} sections, and split them into sentences. Additionally, we collected conversational texts from Reddit and Stack Exchange. For each sentence, we annotated spans of text containing \texttt{entity mentions} with the help of ES. Human annotators verified text-span annotations before the text-spans were converted into NE labels.
}
\label{fig:IntroDatasetProcess}
\end{figure*}

\emph{Second.} From the chosen graph, i.e., Wikidata or ${G}_{Wiki}$, we extract topic-specific sub-graphs (see Figure~\ref{fig:IntroDatasetProcess} and Figure~\ref{fig:WikiSubgraphFood}) as follows. ${G}_{Wiki}$ has many properties (i.e., relations, $\mathcal{R}$), but we employ two properties \emph{instance-of} and \emph{subclass-of} to gather triples $\mathcal{T}_c$ for each topic $c$. 
Then, we extract the head entities $\mathcal{H}_t$ of triples in $\mathcal{T}_c$ to build a collection of hyponyms for each $c$\footnote{We imagine $c$ is a hypernym sharing a relation $r$ with the hyponym. $r$ exists in the triples.}. Lastly, we remove any duplicates in $\mathcal{H}_t$. Entries in $\mathcal{H}_t$ correspond to entity mentions of NE.

\emph{Third.} For each entry in $\mathcal{H}_t$, we find the corresponding Wikipedia article and extract text from that article. Texts from many articles are accumulated, split into sentences, and then prepared for annotation. We extend data collection to all the data sources. For all sentences, we deploy ES to mark all spans of text matching with any entries in $\mathcal{H}_t$.
In this way, we annotate thousands of sentences in a matter of seconds. Next, the annotations are fully human-verified. Annotations of spans of text are converted into BIO tags, creating a novel NER dataset, consisting of pairs of sentences and their BIO tags. 

Our new ES-based annotation scheme significantly reduces the time-and-cost required for labeling texts, hence the name \emph{RapidNER}.

\textit{Ultimately, RapidNER consists of three main components: a knowledge graph (Wikidata), sources of textual data (Wikipedia, Reddit, etc.), and an annotation tool (ES)}. This process is summarized in Algorithm \ref{Algorithm:RapidNER framework}.
\section{Construction of \emph{NERsocial} }
\label{sec:Construction of NERsocial}
To construct NERsocial, we utilized the framework introduced in \secsymbol\ref{sec:Overview of the framework}. We leverage the wealth of knowledge in ${G}_{Wiki}$ and the large number of Reddit posts, Stack Exchange posts, and articles in Wikipedia to build a NER dataset quickly for HRI. We focus on conversational topics or entity types that are typical of social interactions. We describe the process of constructing the dataset in the following sections.
\subsection{Overview of Dataset Construction}
\label{sec:construction overview}
To begin with, we acquired KG knowledge related to six NE, namely, \textbf{Drink, Food, Hobby, Job, Pet, Sport}, in the form of KG triples.  Then, we utilized the triples to (1) collect hyponyms, thus \emph{entity mentions}, for each NE and (2) guide the retrieval of Reddit\&Stack Exchange forums and pages from Wikipedia based on \textit{Page IDs}. Figure~\ref{fig:IntroDatasetProcess} (step 1 to step 4) shows the detailed process.
\subsection{Source Corpus}
\label{Source Corpus}
\textbf{Reddit Posts.} The rich conversational parlance embedded within Reddit posts made it an attractive source of texts for our dataset. Reddit inhibits back and forth interactions between users about a shared topic of interest, housed under a\textit{ subreddit}. Therefore, we collected Reddit posts as follows: 1) created a list of subreddits relevant to each entity type\footnote{For example, a collection of hobbies is available at \url{https://www.reddit.com/r/Hobbies/comments/zqwx6c/the_hobby_master_list_and_their_subreddit/}.} 2) leveraged the API to extract texts from the shortlist of subreddits 3) filtered out non-UTF8 characters, XML tags, URLs, emojis, and insignificant punctuation marks. 4) split paragraphs into single sentences using the \textit{nltk} tokenizer. Under, each subreddit, we prioritized posts that are most popular and thus convey high user engagement. Moreover, all the chosen subreddits in our study contain more than 4M members.  
\\ \textbf{Stack Exchange Posts.} In the same vein as extracting Reddit posts, we created a shortlist of forums on Stack Exchange relevant to each entity type. We access the API and collect posts, followed by a text filtering process described above.  
\\ \textbf{Wikipedia Texts.} We adopt a collection of Wikipedia articles and Wikidata \textit{statements} contained in KDWD\footnote{KDWD can be found here~\url{https://www.kaggle.com/datasets/kenshoresearch/kensho-derived-wikimedia-data}} dataset. The KDWD dataset consists of  6,985 ${G}_{Wiki}$ \texttt{properties}, 51M Wikidata \texttt{items} and 141M Wikidata \texttt{statements}. Moreover, 5,343,564 Wikipedia \texttt{pages} are included. \footnote{KDWD released the following files: \texttt{property.csv, statements.csv, item.csv, page.csv, link\_annotated\_text.jsonl}}. We leverage Wikidata \texttt{properties}, \texttt{items} and \texttt{statements}, in combination with Wikipedia \texttt{pages} to create a new NER dataset.

\begin{figure}[!t]
\centering
\includegraphics[width=7.0cm]{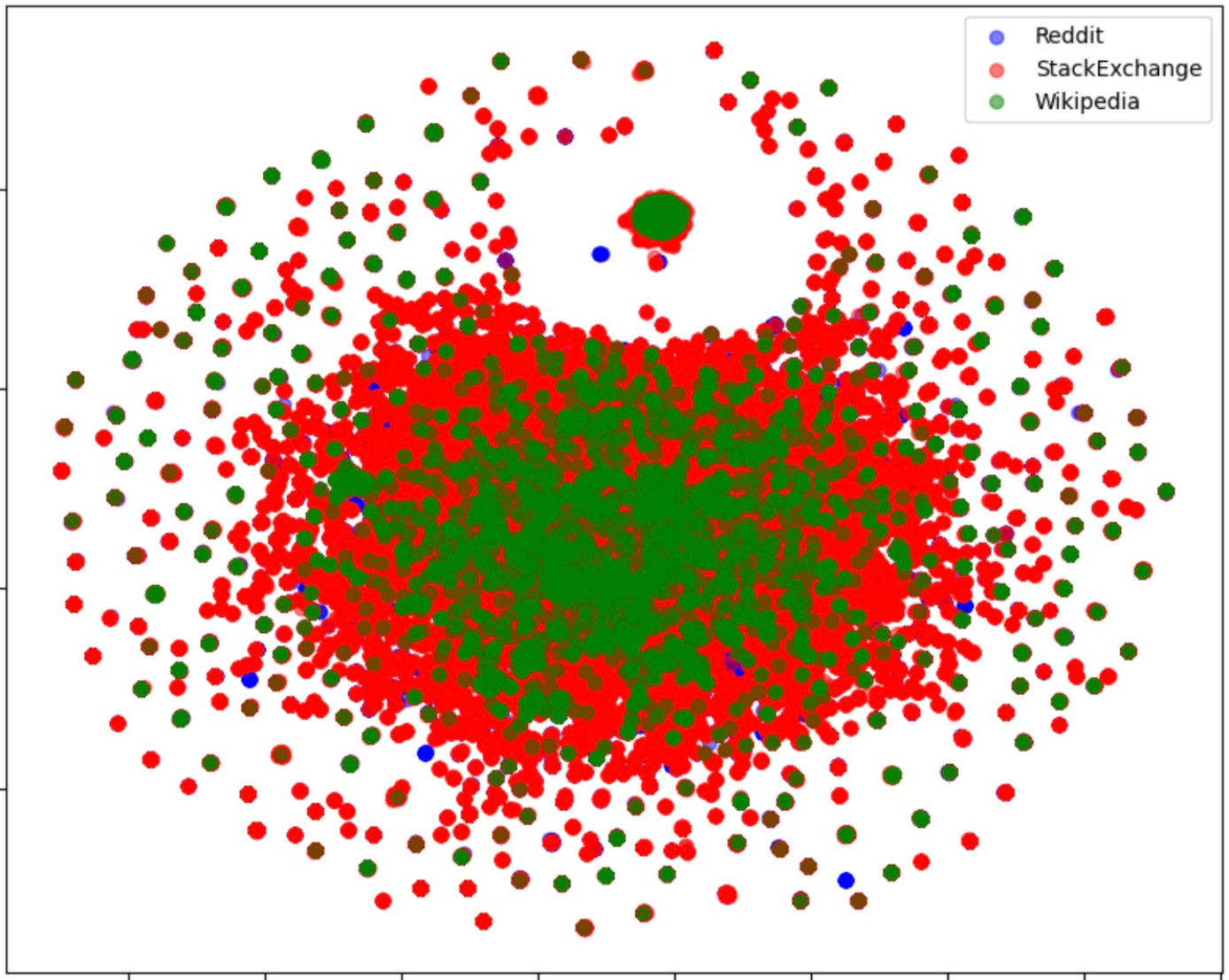}
\caption{The UMAP visualization shows the diversity of texts from three data sources: Reddit, Stack Exchange, and Wikipedia (best seen in color).}
\label{fig:dataSourceVisualization}
\end{figure}

Figure \ref{fig:dataSourceVisualization} is a UMAP visualization illustrating the diversity of texts from three data sources, Reddit, Stack Exchange, and Wikipedia, contained in our dataset. To create this visualization, we leverage \textit{nltk} \footnote{NLTK version is 3.8.1 is used} to tokenize the texts and to remove all \textit{stopwords}. Then, we deploy \textit{TF-IDF} to vectorize the texts, creating embeddings. Finally, we deployed UMAP and set \textit{\#neighbors} to 15, \textit{min distance} to 0.1, \textit{\#components} to 2, and \textit{random state} to 42 to visualize the embeddings from three textual sources.

\subsection{Wikidata Properties}
\label{subsec:Wikidata properties}
${G}_{Wiki}$ has 6,985 \textit{properties}\footnote{All properties provided in KDWD \texttt{property.csv} file.}. Table~\ref{Table: Wikidata properties} defines the two properties used in this study: \emph{instance-of \& subclass-of}.
These properties, in turn, describe \textit{relations} between two \textit{items} in the KDWD dataset, as shown in Table \ref{Table: Examples of Wiki-KG triples}. 
We selected the two properties because, in a triple described by either \emph{instance-of} or \emph{subclass-of} in Wikidata, we can strictly `set' head entity as the hyponym and tail entity as the hypernym\footnote{For illustration, consider these two triples inside Wikidata: \texttt{Fruitopia is instance-of drink brand} and \texttt{Fruitopia is subclass-of drink}. The head entity is \texttt{Fruitopia}, and the tail entity is \texttt{drink brand}/ \texttt{drink}.} \footnote{Fruitopia's page in Wikidata \url{https://www.wikidata.org/wiki/Q263424}}. In this study, the head entities, i.e.,  hyponyms constitute the entity mentions, yet the tail entities, i.e., hypernyms, constitute the named entities. This design allows us to gather many hyponyms for each hypernym (Step 2.2 in Figure \ref{fig:IntroDatasetProcess}), analogous to \textit{hypernym expansion}. The collection of hyponyms is used for (i) retrieving articles from Wikipedia (Step 3.1 in Figure \ref{fig:IntroDatasetProcess}), and (ii) annotation purposes (Step 4 in Figure \ref{fig:IntroDatasetProcess}).
Further  discussion of \textit{Wikidata properties} is shown in the Appendix \ref{Appendix:WikipropertiesBoth}.
\subsection{Selection of Wikidata Triples}
\label{sec: Collection of  Wiki-KG triples}
\textbf{Definition of a triple.} A \textit{knowledge graph triple} is an ordered triple $(h, r, t)$, where $h$ is the \textit{head entity}, $r$ is the \textit{relation}, and $t$ is the \textit{tail entity}. \\
Mathematically, a \textit{knowledge graph} can be represented as a directed graph $G = (V, E)$, where $V$ is the set of vertices (nodes) representing entities, while $E$ is the set of edges representing relationships between entities. Each edge $e \in E$ is an ordered pair $(h, r, t)$, where $h \in V$ is the head entity; $r \in R$ is the relation, with $R$ being the set of all possible relations; and $t \in V$ is the tail entity.
\\ \textbf{Wikidata Triples.} Given ${G}_{Wiki}$ triples $\{(h_1, r_1,t_1), ..(h_n, r_n,t_n)\}$ where $h_i$, $r_i$, $t_i$ denote the head entity, relation, and tail entity respectively, we utilize two relations/properties \emph{instance-of \& subclass-of} to collect triples $\mathcal{T}_c$ for each topic $c$.
In KDWD, there are 141M triples similar to those shown in Table~\ref{Table: Examples of Wiki-KG triples}. We used  \emph{instance-of \& subclass-of} to collect 26M and 1.7M triples, respectively\footnote{That is to say, P31 exists in many more triples in ${G}_{Wiki}$ than P279.}.
\begin{table}[!t]
\small
\centering
\begin{tabular}{*2l}
\toprule
\textbf{ Edge} & \textbf{Target} \\
\textbf{ Property} & \textbf{Item ID} \\
\midrule
instance-of   & 36,906,466\\
subclass-of & 3,695,190 \\
\bottomrule
\end{tabular}
\caption{Examples of ${G}_{Wiki}$ triples defined by \textit{instance-of} and \textit{subclasss-of}.
}
\label{Table: Examples of Wiki-KG triples}
\end{table}
\begin{figure*}[!t]
\centering
\includegraphics[width=11cm ]{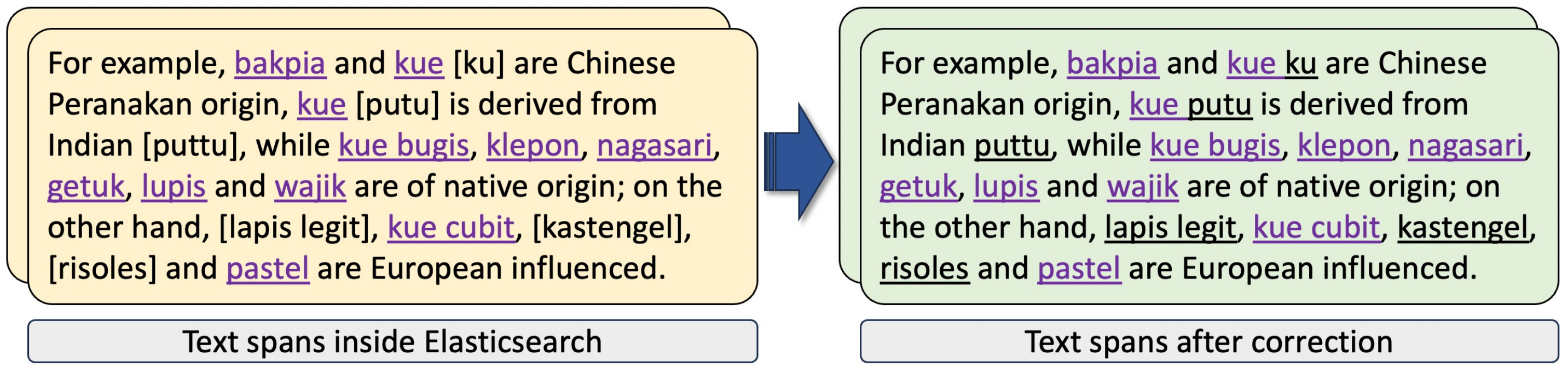}
\caption{Text-span annotations inside ES (underlined text in figure). Some text-spans were incorrectly annotated (marked inside \texttt{[ ]} symbols above). For example, \textbf{Food} mentions were completely missed (i.e., \texttt{puttu}, \texttt{lapis legit}, \texttt{kastengel}, \texttt{risoles}), or a part of the entity mention was left outside the text span (i.e., in \texttt{kue ku}, \texttt{kue putu}). When manually checking the correctness of text-spans annotated with ES, we corrected these spans of text.}
\label{fig:ElasticsearchExample}
\end{figure*}
\subsection{Collection of Wiki-KG Dictionaries}
\label{sec: Collection of Wikidata Dictionaries}
As mentioned in \secsymbol\ref{sec:Overview of the framework}, \secsymbol\ref{subsec:Wikidata properties}, and \secsymbol\ref{sec: Collection of  Wiki-KG triples}, we extracted the head entities $\mathcal{H}_t$ from triples in $\mathcal{T}$. The head entities are hyponyms representing entity mentions for each topic, i.e., NE. 
Next, we created a dictionary for each NE, containing the hyponyms.  We deployed these dictionaries to obtain Wikipedia articles and to annotate spans of text using ES.
In Table~\ref{Table: Size of Dictionaries}, we report the size of dictionaries obtained during dataset curation, for each NE. Examples of dictionary entries for each NE are shown in Table~\ref{Table:DictionariesforNEsTwo}. In addition, we describe steps to obtain dictionaries for each entity type in Appendix \ref{Appendix:Collection of Wikidata Dictionaries}.
\subsection{Selection of Wikipedia Paragraphs}
\label{sec: Selection of Wikipedia Paragraphs}
Whereas it was easier to collect posts from Reddit and Stack Exchange, we followed a more carefully designed process to gather Wikipedia paragraphs.

For each NE, we use the dictionary entries from \secsymbol\ref{sec: Collection of Wikidata Dictionaries}, and their unique \textit{Page IDs} to obtain corresponding articles, from a collection of 5.3M articles in English Wikipedia (more in Appendix \ref{sec: Selection of Wikipedia Paragraphs}). After retrieving the Wikipedia article, we filtered out all other sections to keep only the \textit{introduction} section for that article (we discuss this decision in detail in Appendix \ref{Appendix:filterWikipediaArticles}). 
We gathered all paragraphs in the \textit{introduction} section. Next, we split each paragraph into sentences. The collection of sentences formed the text corpus from which the NER dataset was constructed (see Step 3.2 in Figure \ref{fig:IntroDatasetProcess}). %
\begin{table}[!t]
\small
\centering
\begin{tabular}{l}
\toprule
\textbf{Examples of annotated Sentences } \\
\midrule
\magentauline{Caffè latte}$^{\texttt{DRINK}}$ often shortened to\\
just \magentauline{latte}$^{\texttt{DRINK}}$ in English, is a \magentauline{coffee}$^{\texttt{DRINK}}$  \\
drink of Italian origin made with \magentauline{espresso}$^{\texttt{DRINK}}$ and \\
steamed \magentauline{milk}$^{\texttt{DRINK}}$. Variants include the chocolate-\\
flavored \magentauline{mocha}$^{\texttt{DRINK}}$ or replacing the \magentauline{coffee}$^{\texttt{DRINK}}$ \\
with another beverage base such as \\
\magentauline{masala chai}$^{\texttt{DRINK}}$ (spiced Indian tea), \\
\magentauline{mate}$^{\texttt{DRINK}}$, \magentauline{matcha}$^{\texttt{DRINK}}$, \magentauline{turmeric}$^{\texttt{DRINK}}$ or \\
\magentauline{rooibos}$^{\texttt{DRINK}}$; alternatives to \magentauline{milk}$^{\texttt{DRINK}}$, \\
such as \magentauline{soy milk}$^{\texttt{DRINK}}$ or \magentauline{almond milk}$^{\texttt{DRINK}}$,\\
are also used.\\
\bottomrule
\end{tabular}
\caption{An annotated case from \emph{NERsocial}. The Sentences contain several mentions of \texttt{DRINK} that we annotated.
}
\label{Table: Annotation process for NEs}
\end{table}
\subsection{Annotation Scheme}
\label{subsec: Annotation Design}
We propose an annotation scheme based on ES Throughout the construction of \emph{NERsocial}, annotation of \textit{entity mentions} within each sentence is guided by the dictionaries gathered in \secsymbol\ref{sec: Collection of Wikidata Dictionaries}. We prepared thousands of sentences for annotation by removing any XML\&HTML tags, URLs, and non-UTF8 characters. Then, we imported the sentences into ES. Table \ref{Table: Annotation process for NEs} shows an annotated case from \emph{NERsocial}.
ES details are shown in Appendix \ref{section:Elasticsearch}.
\\ \textbf{Humans verify text-spans.} After annotating spans of text corresponding to \textit{entity mentions} within sentences, we applied human checks to ensure the text-spans were accurate. Moreover, incorrect spans were corrected while missed entity mentions were annotated\footnote{Misssed spans only occur when the dictionary does not contain the entry matching that span of text.} (illustrated in Figure \ref{fig:ElasticsearchExample}).

Recall that our annotation scheme utilizes dictionary entries gathered from a reputable knowledge source, ${G}_{Wiki}$. Most annotations were correct. Hence, minimal incorrect annotations and minimal cost of human verification.
\\ \textbf{Text-span to BIO-tag conversion.} We implemented a Python program to assign the appropriate BIO tags to all marked spans of text in each sentence. The pairs of sentence and BIO tags constitute our new NER dataset, dubbed \emph{NERsocial}. Hereafter, \emph{NERsocial} is split into train/dev/test sets to fine-tune a NER model (Step 4 in Figure \ref{fig:IntroDatasetProcess}).
\begin{table}[!t]
\footnotesize
\centering
\begin{tabular}{*4c}
\toprule
 & \textbf{Annotator} & \textbf{Annotator} & \textbf{Annotator} \\
 & \textbf{A\&B}      & \textbf{A\&C}      & \textbf{B\&C}\\
\midrule
C.Kappa  & 88.3\%   & 92.9\%    & 90.6\%    \\
\bottomrule
\end{tabular}
\caption{\label{Table: inter annotator agreement}
Agreement between pairs of annotators, measured by Cohen's Kappa.} 
\end{table}
\begin{table*}[!t]
\footnotesize
\centering
\resizebox{0.8\linewidth}{!}{
\begin{tabular}{*4r}
\toprule
{\bf Entity Type} &  {\bf \#Entity Tokens}  & {\bf  \#Entities} & {\bf \#Sentences} \\
{} &  {Wiki / Reddit / StackExch}  & {Wiki / Reddit / StackExch}  & {Wiki / Reddit / StackExch}  \\
\midrule
\textbf{Drink}   & 2,531 / 2,604 / \textbf{12,353} & 1,457 / 2,573 / \textbf{12,100} & 1,111 / 1,974 / \textbf{9,674} \\
\textbf{Food}    & 8,651 / \textbf{21,281} / 17,415 & 4,967 / \textbf{18,290} / 15,609 & 1,785 / \textbf{10,000} / \textbf{10,000}  \\
\textbf{Hobby}   & 2,268 / 4,640 / \textbf{13,446} & 1,624 / 4,601 / \textbf{13,446} & 1,147 / 3,521 / \textbf{10,000} \\
\textbf{Job}     & 4,773 / \textbf{14,599} / 12,195 & 3,150 / \textbf{12,207} / 11,241 & 1,479 / 4,061 / \textbf{10,000} \\
\textbf{Pet}     & 3,518 / 5,094 / \textbf{12,823} & 2,013 / 4,899 / \textbf{12,692} & 1,309 / \textbf{10,000}/ \textbf{10,000 }\\
\textbf{Sport}   & 3,089 / 5,211 / \textbf{6,611} & 2,115 / 4,940 / \textbf{6,153} & 1,049 / 3,967 / \textbf{5,274} \\
\midrule
\textbf{NERsocial (total)}   & {\bf 153, 102} & {\bf 134,074} & {\bf 99,448}\\
\midrule
\textbf{WNUT*}  & 6,180 & 3,899 & 5,690 \\
\bottomrule
\end{tabular}
}
\caption{\emph{NERsocial} and \emph{WNUT} statistics. For \emph{NERsocial}, we report the token size, number of entities and sentences for each entity type. }
\label{Table: NERsocial statistics}
\end{table*}
\begin{table}[!t]
    \centering
    \small
    \begin{tabular}{rrr}
    \toprule
         \textbf{Data source} & \textbf{Train/Valid/Test} & \textbf{Total \#Sent.} \\
         \hline
          Wikipedia & 6.2K / 0.8K / 0.8K & 7.8K \\
         Reddit      & 29.3K / 3.7K / 3.7K & 36.6K \\
         Stack Exch. & 44.0K / 5.5K / 5.5K & 54.9K \\
         \hline
         \textbf{NERsocial} & \textbf{79.5K /10K / 10K}  & \textbf{99.4K} \\
         \bottomrule
    \end{tabular}
    \caption{\emph{NERsocial} is created from three data sources. We show the \#sentences from each data source in train/valid/test splits during fine-tuning.}
    \label{tab:allNewNERdatasets}
\end{table}
\begin{table*}[!t]
\small
\centering
\begin{tabular}{*8r}
\toprule
{\bf Datasets} &  {\bf \#Tokens}  & {\bf  \#Entities} & {\bf \#Entity Types} & {\bf \#Sentences}  \\
\midrule
OntoNotes~\cite{weischedel2013ontonotes}   & 2067K   & 161.8K  & 18   & 103.8K  \\
WikiGold~\cite{balasuriya-etal-2009-named}    & 39K     & 3.6K    &  4   & 1.7K     \\
CoNLL2003~\cite{tjong-kim-sang-2002-introduction}   & 301.4K  & 35.1K   & 4    & 22.1K   \\
WNUT2017~\cite{derczynski-etal-2017-results}      & 86.1K   & 3.1K    & 6    & 4.7K     \\
I2B2~\cite{Stubbs2015AnnotatingLC}        & 805.1K  & 28.9K   & 23   & 107.9K    \\
FEW-NERD~\cite{ding-etal-2021-nerd}    & 4601.K  & 491.7K  & 66   & 188.2K \\
\midrule
{\bf NERsocial} & {\bf 153K} & {\bf 134K} & {\bf 6} & {\bf 99.4K} \\
\bottomrule
\end{tabular}
\caption{\label{Table: NERsocial Comparison} Comparison between \emph{NERsocial} and widely used NER datasets. We report all values for OntoNotes, WikiGold, CoNLL2003, WNUT2017, I2B2 and FEW-NERD from~\cite{ding-etal-2021-nerd}.} 
\end{table*}
\section{Dataset Analysis}
\textbf{Diversity.} The 99.4K sentences annotated throughout this process were gathered from three data sources. During annotation, we limit the number of sentences fetched for each entity type to 10K for each textual source, such as Reddit. We limit number of sentences from any Wikipedia page to ten. 
\\ \textbf{Statistics.} The characteristics of \emph{NERsocial} are shown in Table \ref{Table: NERsocial statistics}. Overall, there are 153,102 entity tokens, 134,074 entities, and 99,448 sentences for six entity types. We gathered most of the entity information from Stack Exchange. Table~\ref{Table: NERsocial Comparison} shows a comparison between \emph{NERsocial} and common NER datasets: OntoNotes, WikiGold, CoNLL2003, WNUT17, I2B2 and FEW-NERD. \\
\textbf{Data quality.} The inter-annotator agreement (IAA), as measured by Fleiss Kappa, is 90.6\% indicating a high degree of consistency. Pair-wise agreement for three annotators, measured by Cohen's kappa~\citep{cohen1960coefficient} is shown in Table~\ref{Table: inter annotator agreement}. \\
\textbf{Benefit of diverse data sources.} By extending our framework to \textit{Reddit} and \textit{StackExchange} texts, we annotated 36.6K sentences and 54.9K sentences from both sources, respectively. The sentence number is much larger than 7.8K sentences from Wikipedia texts. Additionally, the diversity of our dataset is enhanced, as shown in Figure \ref{fig:dataSourceVisualization}. 
\begin{table*}[t!]
\centering
\resizebox{0.95\linewidth}{!}{
\begin{tabular}{l|cccccc|cccccc}
\toprule
\textbf{Model} & \textbf{Drink} & \textbf{Food} & \textbf{Hobby} & \textbf{Job} & \textbf{Pet}  & \textbf{Sport} & \textbf{Corp}  & \textbf{C. Work} & \textbf{Grp} & \textbf{Loc} & \textbf{Per} & \textbf{Pdt}\\
\midrule
\textbf{BERT-base}       & 96.71 & 95.24 & 95.67 & 96.73 & 98.65 & 96.65 & 16.00 & 25.53 & 22.22 & 52.56 & 70.27 & 33.33\\
\textbf{RoBERTa-base}    & 96.68 & 94.29 & 95.14 & 96.45 & 98.45 & 96.54 & 42.42 & 31.20 & 39.62 & 64.90 & 77.69 & 43.41 \\
\textbf{DeBERTa-v3-base} & 97.03 & 94.86 & 94.53 & 97.12 & 98.31 & 95.91 & 16.67 & 37.79 & 35.51 & 55.25 & 78.94 & 35.66 \\
\bottomrule
\end{tabular}
}
\caption{F1-scores (\%) per entity type in \textit{NERsocial} and \textit{WNUT}. }
\label{Table:NERsocialwnutF1scores}
\end{table*}
\begin{table}
\centering
\resizebox{0.495\textwidth}{!}{
    \begin{tabular}{l|cc|cc}
    \toprule
     & \textbf{ID} &  &  \textbf{OOD} &   \\
     \hline
     & \textbf{NERsocial} & \textbf{WNUT}  & \textbf{NERsocial} & \textbf{WNUT} \\
     \hline
     \textbf{BERT-base}      & 96.41 & 53.04 & 2.42 & 8.9 \\
     \textbf{RoBERTa-base}   & 95.95 & 60.53 & 2.20 & 13.10 \\
     \textbf{DeBERTa-v3-base}&96.12 & 58.90 &  1.46 & 13.57 \\ 
     \bottomrule
    \end{tabular}
    }
    \caption{F1-scores (\%) in cross-dataset evaluation setting.}
\label{Table:CrossDatasetEvaluation}
\end{table}
\section{Experiments and Results}
\label{sec:Experiments&Results}
To measure the effectiveness of our dataset, we fine-tuned NER models to recognize the newly proposed entity types using the training set of \textit{NERsocial} and evaluated the models on the testing set.
\subsection{Implementation Details}
\label{sec:Implementation Details}
\textbf{NER models.}
Following common practice in supervised NER, we fine-tuned three pre-trained transformer models available at Hugging Face, by splitting our dataset into training data (80\%), validation data (10\%), and testing data (10\%). These are the NER models used in this work:
\\ \textbf{BERT-base}~\citep{devlin-etal-2019-bert}. We deploy \textit{bert-base-uncased}\footnote{\url{https://huggingface.co/google-bert/bert-base-uncased}}, a 110M parameter model. 
\\ \textbf{RoBERTa-base} \cite{liu2019roberta}. We adapt \textit{RoBERTa-base}\footnote{\url{https://huggingface.co/FacebookAI/roberta-base}} and fine-tune it for NER. 
\\ \textbf{DeBERTa-v3-base} \cite{he2021deberta}. We also utilized \textit{deberta-v3-base}\footnote{\url{https://huggingface.co/microsoft/deberta-v3-base}} in our experiments. 
\paragraph{Experimental Setup.} All experiments are conducted with one NVIDIA A100 80GB GPU. We split \emph{NERsocial} into train/dev/test sets based on an 8:1:1 ratio. Then, we fine-tuned the BERT-base, RoBERTa-base, and DeBERTa-v3-base using the train set for 10 epochs and reported the performance averaged over three runs.  We set the batch size to 64, learning rate to $1e-05$, sequence length to 128, and an Adam optimizer. We use three standard NER metrics to evaluate performance: precision, recall, and F1-score. 
\subsection{Results}
\label{subsec:Results}
We observe that BERT-base, RoBERTa-base, and DeBERTa-v3-base achieve competitive results when fine-tuned on \emph{NERsocial}. The average F1-scores are 96.41\%, 95.98\% and 96.12\%, respectively. Overall, models performed best on \textit{pet} and F1-scores are above 98\%, while the least F1-scores were obtained on \textit{food}, i.e., 94.29\%. The F1-scores for each entity type are shown in Table \ref{Table:NERsocialwnutF1scores}, all of which are above 94\%.

A closer look at the three textual sources reveals that NER models fine-tuned on texts from Stack Exchange achieve the highest F1-scores. For instance, the F1-scores for DeBERTa-v3-base after fine-tuning on texts from Wikipedia, Reddit, and Stack Exchange are 83.29\%, 95.97\%, and 98.07\%, respectively, which emphasizes the relevance of texts from interactive platforms (e.g., Reddit and Stack Exchnage) in creating \textit{NERsocial}. More ablations in \secsymbol\ref{subsec:AblationStudy}.

Lastly, due to \textit{RapidNER}'s efficiency, we created a large NER dataset, \textit{NERsocial},  enabling all the models to recognize the new entity types with high F1-scores, as mentioned above. 
\subsection{Comparison to \textit{WNUT 2017} Dataset}
\label{ComparisonToWNUTdata}
Entity types in \textit{WNUT} include: \textit{person, location, corporation, product, creative work \& group} yet \textit{NERsocial} includes \textit{drinks, foods, hobbies, jobs, pets \& sports}. There is some overlap in the entity types between the two datasets, i.e., \textit{product}, a broader category in \textit{WNUT}, may include \textit{drink \& food} in \textit{NERsocial}. Similarly, \textit{creative work} in \textit{WNUT} may include \textit{hobby} in \textit{NERsocial}. Whereas the entity types in \textit{WNUT} are more general, \textit{NERsocial} includes a set of highly specific entity types. 
Texts used to construct \textit{WNUT} originate from Twitter, YouTube,  Stack Exchange, and Reddit. Meanwhile, we constructed \textit{NERsocial} from Wikipedia, Stack Exchange, and Reddit texts. \textit{WNUT} contains 5.69K sentences, yet \textit{NERsocial} contains 99.4K sentences.

Due to similar textual sources for the two datasets, we investigated the generalizability and transferability of models fine-tuned on one dataset to the other. \textit{First}, we establish baselines on both datasets separately. On \textit{WNUT} data, BERT-base, RoBERTa-base, and DeBERTa-v3-base achieve 53.04\%, 60.53\% and 58.90\%, respectively. However,  the same models achieve 96.41\%, 95.95\% and 96.12\%, respectively. We observe a huge gap in absolute F1-scores among the baseline results, and all models perform better on \textit{NERsocial} (See Table \ref{Table:CrossDatasetEvaluation}). 
\textit{Second}, we assess the entity-level performance for all models across specific entity types in both datasets, and all the results are shown in Table \ref{Table:NERsocialwnutF1scores}. Overall, NER models achieve consistently high F1 scores across all entities in \textit{NERsocial}. Conversely, the F1-scores for entities in \textit{WNUT} are much lower, except for \textit{Person} entity type.
\textit{Third,} we conduct a cross-dataset evaluation, and the goal is to assess the performance drop or improvement for models fine-tuned on one dataset and evaluated on the other, a further indication for the similarity or difference among the two datasets. The evaluation results indicate that models fine-tuned on \textit{NERsocial} and evaluated on \textit{WNUT} (BERT-base 8.9\%, RoBERTa-base 13.10\%, and DeBERTa-v3-base 13.57\%) perform better than their counterparts, fin-tuned on \textit{WNUT} and evaluated on \textit{NERsocial} (BERT-base 2.42\%, RoBERTa-base 2.20\%, DeBERTa-v3-base 1.46\%). 
These findings indicate that there is higher transferability from \textit{NERsocial} to \textit{WNUT}; than from \textit{WNUT} to \textit{NERsocial}, to illustrate, 13.57\% vs 1.46\% for DeBERTa-v3-base. (Also in Table \ref{Table:CrossDatasetEvaluation}). 
\section{Ablation Study}
\label{subsec:AblationStudy}
We ablate the textual data used to fine-tune. We fine-tune DeBERTa-v3-base using texts from Wikipedia, social media, and online forums, separately. Then, we examine differences in performance across data sources (see results in Figure \ref{fig:DomainTransferInDeBERTa}). Overall, texts from social media and online forums provide the best fine-tuning results for each entity type in \textit{NERsocial}. Moreover, all models benefit from fine-tuning with NER data aggregated from all three textual sources. 
See detailed ablations in Appendix \ref{Appendix:DomainTransfer}, and the comprehensive summary or ablations in Table \ref{Table:AblationsForDataSources}.
\section{Conclusion}
In this paper, we presented \emph{RapidNER}, a novel framework that showcases an effective approach to creating NER datasets at a large scale through the integration of subgraph extraction, diverse text sources, and ES-based annotation. The resulting \emph{NERsocial} dataset provides a comprehensive resource for developing NER systems in human-robot interaction contexts. Our methodology significantly reduces the time and effort needed for dataset construction, highlighting its potential for broader applications in new domains.

\section*{Ethical Considerations}
\textbf{Misuse and Dual-Use.} There is a possibility for misuse. Given our framework, the ease of creating NER datasets for new domains could be exploited for rogue purposes, such as surveillance, targeted advertising, misinformation, and the like. To ascertain ethical usage of our framework, safeguards and guidelines should be established.
\\ \textbf{Transparency and Accountability.} Researchers using this framework need to be transparent about the capabilities and inherent biases present, and be prepared to take responsibility for all applications or algorithms developed from the datasets generated using this framework.

\section*{Limitations}
\textbf{Representation and Bias.} Our dataset is developed using the content available on Reddit, Stack Exchange, Wikipedia and Wikidata. These platforms inherit demographic and cultural biases emanating from their contributor communities. For that matter, the skewed representation could lead to certain named entities being underrepresented, especially those from non-Western or marginalized communities. The undesirable consequence might be that applications developed using this dataset could have limited universality and fairness.
\\ \textbf{NER vs Entity Linking.} Because NER and entity linking (EL) are often treated as separate tasks, we focused on creating a new dataset containing new named entities. We did not include entity linking, i.e., the disambiguation of those named entities with a knowledge base. Therefore, the EL part will be considered in the future work of this project.
\\ \textbf{Quality and Reliability of Sources.} The accuracy and reliability of all textual information varies with the users who create it. Our current framework does not have an in-built mechanism to verify the truthfulness or historical reliability of the data extracted.
\\ \textbf{Generalizability.} There is an unlimited number of possible entities of interest, yet our study includes just a small fraction of possible entities. The effectiveness of our framework in creating datasets for entities outside the ones tested is not fully established and requires more comprehensive studies.
\\ \textbf{Dynamic Content.} Reddit, Stack Exchange, and Wikipedia are dynamic, with content continually being updated, corrected, and expanded. This could lead to the data drift problem, and inconsistency of datasets generated at different times, affecting the reproducibility of subsequent studies.
\\ \textbf{Language and Cultural Bias.} Due to reliance on Reddit, Stack Exchange, Wikipedia and Wikidata, our framework is exposed to these platforms' biases, including under-representation of non-English languages and non-Western perspectives. 
Such biases can affect the diversity and inclusiveness of the datasets created using our framework.
\\ \textbf{Reliance on Elasticsearch.} The heavy reliance on ES may not be accessible to all researchers, limiting the reproducibility of the study. However, by detailing our annotation process, we hope other researchers will easily adapt this method to construct their NER datasets. 

\section*{Acknowledgment}
This work was done during the development of HARU under the mentorship of Eric Nichols at Honda Research Institute Japan (HRI-JP). 
We thank him for overseeing the creation of datasets, providing access to compute resources and the robot during Jesse's internship at HRI-JP.

\newpage
\bibliographystyle{acl_natbib}
\bibliography{main}

\newpage
\appendix

\section{Definitions of Entity Types} 
Our goal is to support the social robot's ability to interact with humans when the entity types shown in Table \ref{Table:Examples_daily_conversations} occur during interaction; for example, the question \textit{``Do you like Fanta?''} corresponds to the \textit{Drink} entity type.

Below are the definitions of entity types given to the annotators. Annotators were hired based on a minimal understanding of linguistics, and we compensated them at a market rate. All annotators are graduate students, two of whom are male and one is female.
\paragraph{Drink.} This entity type is used for sequences of words denoting explicit mentions of beverages, soft drinks, alcoholic and non-alcoholic drinks. 
\paragraph{Food.} This entity type is used for sequences of words denoting explicit mentions of food, culinary practices, or eating habits. 
\paragraph{Hobby.}This entity type is used for sequences of words denoting explicit mentions of leisure activities or hobbies.
\paragraph{Job.} This entity type is used for sequences of words denoting explicit mentions of professions, employment opportunities, or job-related activities. 
\paragraph{Pet.} This entity type is used for sequences of words denoting explicit mentions of animals traditionally kept as pets. 
\paragraph{Sport.} This entity type is used for sequences of words denoting explicit mentions of physical activities recognized as sports, including both team sports and individual physical activities. 
\begin{table}[!ht]
\footnotesize
\centering
\begin{tabular}{ll}
\toprule
\textbf{Entity Type} & \textbf{Example Question} \\
\hline
\textbf{Drink}  & ``Do you like Fanta?''\\
\textbf{Food}   & ``What is your favorite cuisine?'' \\
\textbf{Hobby}  & ``What do you do in your free time?'' \\
\textbf{Job}    & ``What do you do for a living?'' \\
\textbf{Pet}    & ``Do you like pets?''\\
\textbf{Sport}  & ``What's your favorite sport?'' \\
\bottomrule
\end{tabular}
\caption{Exemplar occurrences of the entity types in human dialogue.}
\label{Table:Examples_daily_conversations}
\end{table}

\begin{figure*}[!t]
\centering
\includegraphics[width=14.8cm]{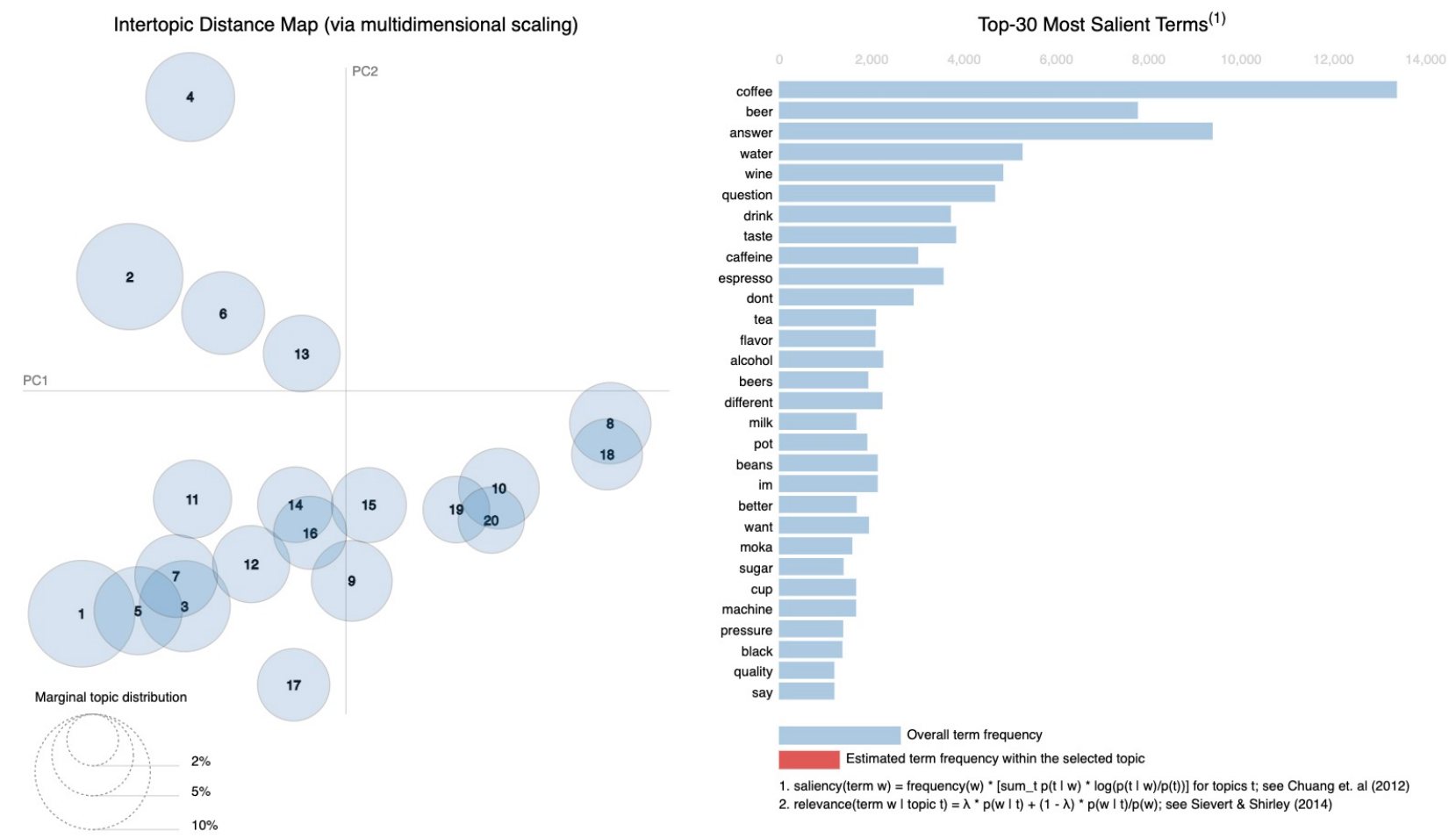}
\caption{The LDA visualization shows the distance between topic-clusters, and the thirty most common terms under the \textit{Drink} entity type in our dataset, among all text sources.}
\label{fig:drinkTopicsVisualized}
\end{figure*}

We obtained a diverse collection of textual information for each of the entity types described above. Figure \ref{fig:drinkTopicsVisualized} is an LDA visualization showing the thirty most prominent \textit{keywords} from the \textit{drink} texts,  and the texts are aggregated from all data sources: Reddit, Stack Exchange, and Wikipedia. Moreover, we can see the distance between clusters of topics in the \textit{drink} information, and most topics are closely related to each other.
\section{Reddit Posts}
\label{Reddit Posts}
Tables \ref{Table:listOfAllSubreddits}, \ref{Table:listOfAllSubbredditsHobbyPart1}, and \ref{Table:listOfAllSubbredditsHobbyPart2} contain the collection of \textit{subreddits} from which we collected texts while constructing the dataset. For the \textit{hobby} entity type, we adapted a comprehensive list from \url{https://www.reddit.com/r/Hobbies/comments/zqwx6c/the_hobby_master_list_and_their_subreddit/} but several channels contained zero discussions. Therefore, we needed to use a much larger collection of channels than for the other entity types.  
\begin{table*}[!t]
\centering
\footnotesize  %
\begin{tabular}{p{0.95\textwidth}}  %
\toprule
\textbf{Collection of Subreddits used in our dataset} \\
\midrule

\textbf{Entity Type: \textcolor{purple}{Drink}} \\
Coffee, tea, espresso, Kombucha, starbucks, BreakfastFood, Smoothies, Canning, restaurant, UberEATS, nespresso, dehydrating, TeaPorn, RateMyTea, teaexchange, teasales, TeaPictures, boba, wine, winemaking \\
\midrule

\textbf{Entity Type: \textcolor{purple}{Food}} \\
food, FoodPorn, EatCheapAndHealthy, foodhacks, MealPrepSunday, slowcooking, Cooking, keto, recipes, GifRecipes, HealthyFood, Baking, 15minutefood, budgetfood, StupidFood, cookingforbeginners, grilling, Breadit, Pizza, WeWantPlates, ramen, AskCulinary, Cheap\_Meals, easyrecipes, EatCheapAndVegan, seriouseats, burgers, IndianFood, steak, drunkencookery, BBQ, mexicanfood, castiron, pasta, KoreanFood, instantpot, Cheese, veganrecipes, WhatShouldICook, Sourdough, CulinaryPlating, grilledcheese, chinesefood, sushi, eatsandwiches, cakedecorating, Old\_Recipes, budgetcooking, MeatlessMealPrep, tonightsdinner, Foodforthought, doordash, streeteats, PutAnEggOnIt, filipinofood, cookingtonight, Freefood, finedining, Sandwiches, trailmeals, fastfood, RedditInTheKitchen, deliciouscompliance, AskBaking \\
\midrule

\textbf{Entity Type: \textcolor{purple}{Job}} \\
jobs, NotMyJob, WorkOnline, findapath, careeradvice, freelance, GetEmployed, medicine, KitchenConfidential, recruitinghell, AskHR, FinancialCareers, TalesFromTheFrontDesk, graphic\_design, datascience, cscareerquestions, Entrepreneur, careerguidance, compsci, Filmmakers, ecommerce, Carpentry, ITCareerQuestions, electricians, uberdrivers, Construction, forhire, sales, overemployed, engineering, EngineeringStudents, EngineeringPorn, MechanicalEngineering, AskEngineers, StonerEngineering, redneckengineering, ElectricalEngineering, civilengineering, CIVIL\_ENGINEERING, ReverseEngineering, ChemicalEngineering, EngineeringJobs, ECE, EngineeringDocs, AerospaceEngineering, EngineeringResumes, SoftwareEngineering, StructuralEngineering, dataengineering, SocialEngineering, ComputerEngineering, audioengineering, bioengineering \\
\midrule

\textbf{Entity Type: \textcolor{purple}{Pet}} \\
aww, AnimalsBeingDerps, AnimalsBeingBros, AnimalsBeingJerks, rarepuppers, cats, Awwducational, FunnyAnimals, Eyebleach, likeus, dogs, natureismetal, WhatsWrongWithYourDog, AnimalsBeingGeniuses, StartledCats, Zoomies, Dogtraining, corgi, PuppySmiles, WhatsWrongWithYourCat, Catloaf, holdmycatnip, awwwtf, parrots, Catswhoyell, Chonkers, CatsStandingUp, wildlifephotography, CatsOnKeyboards, catfaceplant, DogsAndPlants, BackYardChickens, Awww, woof\_irl, Rabbits, goldenretrievers, cute, pitbulls, blackcats, Dachshund, hitmanimals, birding, snakes, Thisismylifemeow, RATS, Superbowl, spiders, happycowgifs, MasterReturns, trashpandas \\
\midrule

\textbf{Entity Type: \textcolor{purple}{Sport}} \\
sports, nba, nfl, soccer, PremierLeague, CFB, running, MMA, baseball, CollegeBasketball, Boxing, hockey, nhl, ufc, snowboarding, skiing, Cricket, Basketball, climbing, soccer, tennis, NASCAR, bicycling, Bundesliga, golf, EASportsFC, SquaredCircle, worldcup, futbol, formuladank, football, fantasybball, bjj, MLS, cycling, FantasyPL, 49ers, championsleague, rugbyunion, GrandPrixRacing, powerlifting, olympics, theocho, skateboarding, LaLiga, cowboys, chelseafc, bouldering, footballmanagergames, Gunners, ipl, LigaMX, motogp, martialarts, sixers, MTB, INDYCAR, AFL, eagles, leafs, chicagobulls, NYYankees, fantasybaseball, seriea, rally, discgolf \\
\bottomrule
\end{tabular}
\caption{A list of subreddits used to construct \textit{NERsocial}.}
\label{Table:listOfAllSubreddits}
\end{table*}
\begin{table*}[!t]
\centering
\tiny
\begin{tabular}{p{0.95\textwidth}}  %
\toprule
\textbf{Collection of \textbf{Hobby} Subreddits used in our dataset} \\
\midrule
\textbf{Target entity: \textcolor{purple}{Hobby}.}  3DPrinting, functionalprint, FixMyPrint, 3dprinter, 3dprintingdeals, PrintedMinis, sketchup, AcroYoga, acting, Actingclass, ActingNerds, ActionFigures, AnimeFigures, ArticulatedPlastic, aerospace, AerospaceEngineering, airhockey, Planespotting, FlightSpotting, airsoft, airsoftmarket, airsoftcirclejerk, Speedsoft, LeftWingAirsoft, airsoftgore, GasBlowBack, Airsoft\_UK, airsoftmarketcanada, airsoftcanada, A\_irsoft, animation, animationcareer, 2DAnimation, antkeeping, ants, antscanada, Antiques, ArtefactPorn, UnknownArtefact, whatsthisworth, whatisthisthing, ThriftStoreHauls, Aquascape, PlantedTank, Aquariums, Archaeology, Archeology, Archery, TraditionalArchery, bowhunting, Bowyer, archeryexchange, Art, artcollecting, ArtFundamentals, ArtHistory, ArtistLounge, ArtTherapy, artstore, ArtBuddy, ArtCrit, ArtProgressPics, artbusiness, ArtEd, Artadvice, Artists, ArtPorn, HungryArtists, beginnerastrology, Advancedastrology, AstrologyChartShare, astrology, astrologyreadings, AskAstrologers, AstrologyCharts, astrologymemes, LetsFuckWithAstrology, Zodiac, Astronomy, askastronomy, telescopes, astrophotography, astrophysics, audiophile, diyaudio, BudgetAudiophile, audiophilemusic, audio, audiojerk, vintageaudio, AutoDetailing, Detailing, auto\_racing, Racecars, motorsports, projectcar, classiccars, AxeThrowing, Axecraft, throwing, basejumping, SkyDiving, bmx, BmxStreets, bmxcruiser, Bikeporn, bmxracing, dirtjumping, bicycling, xbiking, backgammon, backpacking, WildernessBackpacking, BackpackingDogs, Backpackingstoves, BackpackingPictures, backpackingfood, AdvancedBackpacking, UltralightBackpacking, PHikingAndBackpacking, lightweight, Ultralight, solotravel, PNWhiking, Shoestring, UltralightCanada, wanderlust, ULHikingUK, VIRGINIA\_HIKING, outdoorgear, adventures, badminton, BadmintonWorld, Baking, BALLET, balletpics, Flats, ballroom, DanceSport, baseball, mlb, collegebaseball, fantasybaseball, BaseballGloves, InternationalBaseball, baseballgifs, baseballunis, baseballstats, Basketball, nba, BasketballTips, CollegeBasketball, basketballcoach, BBallShoes, fantasybball, fantasybasketball, basketballcards, BasketballShoes, basketballjerseys, NCAAW, batontwirling, beachvolleyball, beachcombing, beatbox, Beatboxing, BeautyQueens, Beekeeping, bees, beer, beerporn, CraftBeer, TheBrewery, beercirclejerk, Homebrewing, atlbeer, ctbeer, LABeer, chicagobeer, AustinBeer, ColumbusBeer, njbeer, AusBeer, ncbeer, bellringing, bells, allbenchmarks, overclocking, billiards, pool, biology, biologymemes, Biologyporn, biotech, molecularbiology, marinebiology, bioinformatics, Biochemistry, birdwatching, birding, whatsthisbird, blacksmithing, Blacksmith, Bladesmith, knifemaking, Axecraft, Blogging, blogs, blogsnark, longboarding, snowboarding, surfing, skimboarding, boostedboards, skateboarding, ElectricSkateboarding, boardgames, boardgame, boardgamescirclejerk, BoardgameDesign, BoardGameExchange, Boardgamedeals, tabletopgamedesign, soloboardgaming, bodybuilding, naturalbodybuilding, femalebodybuilding, FunBodybuilding, powerbuilding, BodybuildingAdvice, TeenBodybuilding, bodybuildingpics, bonsaicommunity, BonsaiPorn, CannaBonsai, Bonsai\_Pottery, Bonchi, bookfolding, bookbinding, BookCollecting, bookhaul, bookshelf, bookporn, MangaCollectors, rarebooks, bookrepair, TheBindery, botany, Bowling, bowlingalleyscreens, Boxing, amateur\_boxing, boxingworkoutlogs, boxingcirclejerk, boxingdiscussion, bjj, brazilianjiujitsu, Breadit, Bread, breadmaking, Sourdough, BreadMachines, Breakdancing, bridge, bulletjournal, BasicBulletJournals, bujo, digitalbujo, Butterflies, butterfly, button, Calisthenic, CalisthenicsCulture, HybridCalisthenics, calisthenicsparks, bodyweightfitness, Calligraphy, BrushCalligraphy, Scribes, calligraffiti, ArabicCalligraphy, camping, CampingGear, CampingandHiking, CampingPorn, Outdoors, carcamping, StealthCamping, hammockcamping, TruckCampers, motocamping, canoecamping, WinterCamping, CampAndHikeMichigan, wildcampingintheuk, candlemaking, Candles, luxurycandles, goosecreekcandles, candy, CandyMakers, CandyMaking, canoeing, canoecamping, canoewithaview, canyoneering, carspotting, spotted, exoticspotting, StreetviewCarSpotting, ClassicCarSpotting, ECU\_Tuning, cardistry, playingcards, PlayingCardsMarket, cardistry, tradingcardcommunity, sportscards, pokemoncards, pokemoncardcollectors, soccercard, footballcards, baseballcards, basketballcards, hockeycards, caving, Ceramics, Pottery, checkers, cheerleaders, Cheerleading, cheesemaking, Cheese, chemistry, chemhelp, chemistryhelp, OrganicChemistry, ChemicalEngineering, chemistrymemes, ALevelChemistry, chemistryporn, chess, chessbeginners, ChessPuzzles, chessporn, chessvariants, ComputerChess, AnarchyChess, chessmemes, chessopenings, climbing, climbergirls, climbharder, bouldering, ClimbingPorn, Indoorclimbing, ClimbingGear, tradclimbing, ClimbingPartners, RockClimbing, urbanclimbing, iceclimbing, alpinism, ClimbingCircleJerk, Mountaineering, DIYclothes, roasting, CoffeeRoasting, coffee\_roasters, coincollecting, coins, AncientCoins, Silverbugs, CoinClub, EuroCoins, Colorguard, Coloring, AdultColoring, Coloringbookspastime, coloringtherapy, comicbookcollecting, MangaCollectors, ComicBookPorn, competitiveeating, composting, Cooking, cookingforbeginners, cookingtonight, AskCulinary, cookingvideos, cookingtips, CookingForOne, Cornhole, cosplay, CosplayNation, cosplayers, CosplayandModeling, couponing, coupons, crafts, CraftingWorld, crafting, craftsnark, craftit, somethingimade, creativewriting, cribbage, Cricket, CricketShitpost, IndiaCricket, EnglandCricket, crochet, crocheting, Amigurumi, crochetpatterns, GeekyCrochet, CrochetBlankets, croquet, CrossStitch, cross\_stitch, StitchersofReddit, Stitchy, Embroidery, crossword, crosswords, NYTCrossword, crypto, cryptography, netsec, codes, Crystals, Crystalsforbeginners, crystalgrowing, Crystalsforsale, Curling, cycling, bicycling, ladycyclists, BicyclingCirclejerk, bikecommuting, CyclingFashion, ukbike, Velo, gravelcycling, bicycletouring, bicycleculture, vancouvercycling, londoncycling, NYCbike, torontobiking, RedditPHCyclingClub, FixedGearBicycle, ausbike, Cyclingmemesofficial, CyclingMSP, CyclingTech, DJs, Beatmatch, djing, DJSetups, Turntablists, DJsCirclejerk, dance, dancing, poledancing, SalsaDancing, SwingDancing, JustDance, bachata, westcoastswing, Darts, DartsTalk, Debate, HomeDecorating, interiordecorating, InteriorDesign, DesignMyRoom, Decorating, deltiology, PostCardExchange, diamondpainting, dioramas, TerrainBuilding, Miniworlds, discgolf, discgolfcirclejerk, discgolfgirls, discexchange, DiscReleases, discdyeing, electronics, AskElectronics, diyelectronics, Arduino, ElectronicsRepair, arduino, ElectronicsSalvage, ECE, engraving, Laserengraving, ephemera, Equestrian, Horses, esports, ExhibitionDrill, dfsports, fantasysports, fantasyfootball, fantasybball, fantasyhockey, fantasybaseball, FantasyPL, farming, urbanfarming, verticalfarming, dairyfarming, Fencing, HistoricalFencing, FengShui, Fieldhockey, FigureSkating, iceskating, cinematography, TrueFilm, Filmmakers, movies, criterion, Shortfilms, ExperimentalFilm, videography, boxoffice, FishFarming, Aquaculture, aquaponics, Fishing, FishingForBeginners, Fishing\_Gear, Fishingmemes, troutfishing, SurfFishing, kayakfishing, CarpFishing, IceFishing, flyfishing, FishingAustralia, FlyFishingCircleJerk, bassfishing, FishingOntario, fishkeeping, Aquariums, bettafish, Goldfish, shittyaquariums, PlantedTank, fishtank, Koi, nanotank, corydoras, Fitness, FitAndNatural, xxfitness, FitnessGirls, bodybuilding, bodyweightfitness, homegym, fitness30plus, FitnessGuides, fatlogic, PetiteFitness, gymsnark, FitnessMaterialHeaven, loseit, nattyorjuice, fitness40plus, stopdrinkingfitness, crossfit, veganfitness, fitnessonline, fitpregnancy, FitnessOver50, vrfit, fitness50plus, Fitnessguideshare, flagfootball, FlowerArranging, flowers, FloralDesign, Cutflowers, Floristry, houseplants, foraging, foraginguk, fossicking, FossilHunting, FossilPorn, fossilid, Paleontology, fossils, FreestyleFootball, Frisbee, ultimate, fruit, FruitTree, woodworking, BeginnerWoodWorking, Carpentry, palletfurniture, gaming, pcgaming, Gaming4Gamers, GirlGamers, gamingsetups, truegaming, gamingsuggestions, GamingDetails, retrogaming, gamingnews, gamingpc, GamingLaptops, Gamecollecting, gamingmarket, linux\_gaming, gardening, vegetablegardening, GuerrillaGardening, SquareFootGardening, UrbanGardening, GardenWild, IndoorGarden, homestead, GardeningAustralia, OrganicGardening, AustinGardening, GardeningUK, whatsthisplant, houseplants, containergardening, garden, HawaiiGardening, indoorgardening, NativePlantGardening, GardeningPNW, GardeningWhenItCounts, GardeningInventions, Genealogy, Ancestry, GeneticGenealogyNews, 23andme, AncestryDNA, geocaching, geocache, geology, geologyporn, geologycareers, Geologymemes, whatsthisrock, rockhounds, GeologySchool, GhostHunting, Ghosts, Paranormal, Ghoststories, GingerbreadHouses, glassblowing, lampwork, gogame, Prospecting, Gold, golf, GolfSwing, golfclassifieds, ProGolf, GongFuTea, Gongoozling, Graffiti, blackbookgraffiti, graffhelp, GraffitiTagging, GraffitiStickers, groundhopping, gunsmithing, DIYGuns, Gymnastics, CollegeGymnastics, WomensGymnastics, GymnasticsWorld, hacking, HowToHack, cybersecurity, masterhacker, Hacking\_Tricks, HackingTechniques, learnhacking, hardwarehacking, ethicalhacking, hamradio, Handball, herbalism, Herblore, Horses, horseracing, HomeImprovement, homeimprovementideas, Home, \\
\bottomrule
\end{tabular}
\caption{A list of subreddits for \textbf{Hobby} used to construct \textit{NERsocial}. Part I.}
\label{Table:listOfAllSubbredditsHobbyPart1} 
\end{table*}
\begin{table*}[!t]
\centering
\tiny
\begin{tabular}{p{0.95\textwidth}}  %
\toprule
\textbf{Collection of \textbf{Hobby} Subreddits used in our dataset} \\
\midrule
\textbf{Target entity: \textcolor{purple}{Hobby}.}  
homeautomation, hometheater, homeowners, Homebrewing, HBL, TheBrewery, brewing, Horses, horsebackriding, Horseshoes, Hooping, Hunting, bowhunting, Huntingdogs, elkhunting, coyotehunting, CanadaHunting, Duckhunting, Waterfowl, Californiahunting, TexasHunting, HuntingAlberta, HuntingAustralia, HuntingGearChat, HuntingRecipes, GhostHunting, TreasureHunting, Arrowheads, hurling, HydroDip, Hydroponics, hydro, SemiHydro, aquaponics, hockeyplayers, hockey, nhl, collegehockey, hockeyjerseys, hockeygoalies, iceboating, inlineskating, Inline\_Cafe, AggressiveInline, inline, BladerNews, Entomology, InsectCollections, whatsthisb, Insectcollecting, Instruments, NativeInstruments, Luthier, UnusualInstruments, InstrumentPorn, guitar, piano, drums, bass, violin, trumpet, saxophone, trombone, ukelele, harmonica, uke, fiddle, orchestra, brass, woodwinds, percussion, bagpipes, banjo, accordion, mandolin, harmonium, sitar, didgeridoo, bouzouki, invention, jewelrymaking, jewelry, jewelers, jewelrylove, resin, Jigsawpuzzles, jogging, slowjogging, C25K, running, Journaling, JournalingIsArt, ArtJournaling, notebooks, judo, SurpriseJudo, juggling, jiujitsu, Ju\_Jutsu, Kabaddi, prokabaddi, karaoke, singing, gokarts, Karting, Kayaking, kayakfishing, Kendama, Kendo, kites, kitesurfing, Kiteboarding, knives, Knife\_Swap, knifeclub, chefknives, knifemaking, Bladesmith, ThrowingKnife, knitting, casualknitting, KnitHacker, knittinghelp, YarnAddicts, GeeKnitting, yarnporn, MachineKnitting, LoomKnitting, Yarnswap, Kitting, Drunkknitting, nailbinding, tatting, BobbinLace, knots, macrame, Kombucha, findascoby, LARP, Lapidary, Leathercraft, Leatherworking, LeatherClassifieds, LeatherTutorials, Leatherworkers, lego, legodeal, Legomarket, legostarwars, legotechnic, legos, legomodular, Legodimensions, LEGOtrains, LegoTechniques, legoleaks, Letterboxing, linguistics, asklinguistics, badlinguistics, linguisticshumor, languagelearning, lockpicking, Locksmith, lomography, longboarding, longboardingDISTANCE, LongboardBuilding, Machinists, machining, manufacturing, MachinistPorn, MachinePorn, engineering, EngineeringPorn, AskEngineers, macrame, Magic, magick, occult, magnetfishing, Mahjong, Makeup, MakeupEducation, Makeup101, MakeupRehab, MakeupLounge, UnconventionalMakeup, makeuptips, makeupflatlays, MakeupAddiction, MakeupAddicts, beauty, BeautyGuruChatter, RandomActsofMakeup, makeupfreebies, makeupdupes, MakeupForMen, makeupartists, makeuporganization, Makeup\_Reviews, MakeupNews, manga, mangaswap, MangaCollectors, MangaFrames, manganews, MangaArt, mangadeals, Mangamakers, mangacoloring, MangaMemes, manhwa, Manhua, Marbles, marblerun, marchingband, CMB, Bandmemes, drumcorps, band, martialarts, MMA, MuayThai, kravmaga, aikido, MilitaryMartialArts, taekwondo, wma, karate, kungfu, judo, bjj, StreetMartialArts, TheMcDojoLife, kungfucinema, massage, MassageTherapists, MassageGuns, reflexology, learnmath, mathematics, math, matheducation, mathmemes, PhilosophyofMath, badmathematics, mazes, puzzles, mechanics, MechanicAdvice, AskMechanics, medical, medicine, medical\_advice, medlabprofessionals, medicalschoolanki, medicalschool, premed, MedicalPhysics, MedicalHelp, MedicalMeme, Meditation, MeditationPractice, audiomeditation, meditationscience, transcendental, TheMindIlluminated, streamentry, memorypalace, memory, metaldetecting, BottleDigging, metalworking, Welding, Blacksmith, blacksmithing, meteorology, weather, climate, microbiology, microscopy, electronmicroscopy, MicroNatureIsMetal, MicroPorn, underthemicroscope, Minerals, rockhounds, Radioactive\_Rocks, mineralcollectors, whatsthisrock, MineralPorn, Model, modelmakers, ModelCars, modeltrains, Gunpla, 3Dmodeling, modelengineering, ModelShips, modelrailroads, modelplanes, modelexchange, modelrockets, modelrocketry, modeltanks, 3Dprinting, modelaircraft, Model, MODELING, CosplayandModeling, motorsports, motorcycles, motorcycle, MotoUK, bikesgonewild, motocamping, SuggestAMotorcycle, motorcyclesroadtrip, Ducati, vintagemotorcycles, RideitJapan, adventuremotorcycling, AdventureBike, MotoIRELAND, MTB, Bikeporn, bicycling, xbiking, Mountaineering, alpinism, climbing, museum, MuseumPros, MuseumPorn, Music, LetsTalkMusic, musicsuggestions, WeAreTheMusicMakers, MusicPromotion, musicproduction, indieheads, MusicInTheMaking, MusicFeedback, musictheory, MusicBattlestations, MusicRecommendations, musicians, musicmarketing, composer, musicology, MusicMatch, MusicForConcentration, musicals, edmproduction, MusicIndia, MusicaEnEspanol, MusicEd, MusicForRPG, MusicNews, musicindustry, musicbusiness, MusicProductionDeals, musichoarder, musicfestivals, MusicalTheatre, musicalcomedy, musicalscripts, musiccognition, musicmakers, musicprogramming, musiceducation, Musicandmathematics, mycology, MushroomGrowers, unclebens, Mushrooms, shroomers, shrooms, sporetraders, MycoBazaar, MycoBuySellTrade, PsilocybinMushrooms, MycologyandGenetics, shroomery, SporeTradersIndia, mushroom\_hunting, NailArt, Nails, Nailpolish, malepolish, lacqueristas, redditpolish, nail\_art, DipPowderNails, Needlepoint, netball, neuroscience, neuro, Neuropsychology, Outdoors, Outdoor, outdoorgear, paintball, painting, acrylicpainting, oilpainting, learnart, Art, WhatIsThisPainting, minipainting, AbstractArt, PaintingTutorials, DigitalPainting, Watercolor, PourPainting, paragliding, freeflight, paramotor, Parkour, Goatparkour, ParkourTeachers, penspinning, penspin, MeetPeople, Performance, PerformanceArt, fragrance, Perfumes, PerfumeExchange, fragranceclones, petsitting, philately, stamps, Phillumeny, philosophy, askphilosophy, PhilosophyMemes, AcademicPhilosophy, photography, AmateurPhotography, streetphotography, photocritique, itookapicture, AskPhotography, Beginning\_Photography, filmphotography, ExposurePorn, PhotographyProTips, PhotographyTutorials, wildlifephotography, postprocessing, Casual\_Photography, Photography\_Gear, PhotographyJobs, analog, LandscapePhotography, Cameras, WeddingPhotography, foodphotography, ToyPhotography, Physics, AskPhysics, PhysicsStudents, physicsmemes, QuantumPhysics, GamePhysics, MedicalPhysics, apphysics, CartoonPhysics, PhysicsHelp, TheoreticalPhysics, Pickleball, pilates, Posture, ClubPilates, Pins, EnamelPins, plastic, Instruments, NativeInstruments, UnusualInstruments, InstrumentPorn, podcast, podcasts, podcasting, PodcastClassifieds, PodcastGuestExchange, Poetry, OCPoetry, ShittyPoetry, poetry\_critics, poetryreading, PoetrySlam, poi, poker, Poker\_Theory, onlinepoker, PokerVids, poledancing, poledance, polo, waterpolo, pools, swimmingpools, postcrossing, Pottery, Ceramics, potteryporn, powerboats, powerlifting, powerbuilding, Iron, bodybuilding, weightlifting, strength\_training, weightroom, Weakpots, Jokes, flowerpressing, editing, Proverbs, biblereading, psychology, askpsychology, psychologystudents, PsychologyTalk, AcademicPsychology, ClinicalPsychology, IOPsychology, psychologymemes, psychologyresearch, schoolpsychology, positivepsychology, PublicSpeaking, puppetry, puppets, puzzles, Jigsawpuzzles, LogicPuzzles, mechanicalpuzzles, Pyrography, qigong, TrueQiGong, Quidditch, quilling, papercraft, quilting, modernquilts, quiltingblockswap, quiz, quiz, trivia, racing, racewalking, racquetball, radiocontrol, whitewater, rafting, rappelling, makinghiphop, hiphop101, books, suggestmeabook, ReadingGroup, bookscirclejerk, 52book, kindle, recipes, TheHighChef, cookbooks, Cooking, vinyl, hiphopvinyl, heavyvinyl, vinyljerk, Vinyl\_Jazz, finishing, reiki, energy\_work, renfaire, homerenovations, AusRenovation, Renovations, HomeImprovement, research, researchchemicals, SampleSize, gadgets, Futurology, robotics, shittyrobots, FRC, battlebots, FTC, RockBalancing, RockClimbing, climbharder, climbing, climbergirls, ClimbingPorn, rockpainting, rockhounds, RockhoundExchange, RockTumbling, whatsthisrock, rpg, RPGdesign, rpg\_gamers, DnD, rollerderby, RollerDerbyReddit, Rollerskating, rollerblading, ItsMySkateNight, rollerskate, Rubiks\_Cubes, Cubers, rugbyunion, RugbyTraining, USArugby, RugbyAustralia, CanadaRugby, springboks, superleague, rughooking, running, AdvancedRunning, runninglifestyle, BeginnersRunning, RunningShoeGeeks, trailrunning, XXRunning, C25K, SafariLive, sailing, SailboatCruising, Tallships, boatporn, sailingcrew, SandArt, scouting, scrapbooking, diving, scuba, scubaGear, scubadiving, freediving, Rowing, sculpting, miniaturesculpting, Sculpture, seaglass, Shelling, sewing, SewingForBeginners, freepatterns, vintagesewing, sewhelp, SewingTips, sewingpatterns, SewingMachinePorn, HandSew, SewingStations, shoemaking, Cordwaining, shogi, Shooting, shortwave, SWL, shuffleboard, singing, SingingTips, ratemysinging, redditsings, skateboarding, NewSkaters, OldSkaters, skateboardhelp, classicskateboarding, ElectricSkateboarding, Sketching, SketchingPrompts, skiing, Skigear, xcountryskiing, Skijumping, ski, skimboarding, jumprope, SkyDiving, Slackline, slacklining, Dogsledding, Sledding, slotcars, snorkeling, snowboarding, snowboardingnoobs, ShredditGirls, snowmobiling, snowshoeing, soapmaking, soccer, bootroom, WomensSoccer, CanadaSoccer, NWSL, soccercirclejerk, Softball, CollegeSoftball, Spearfishing, Speedskating, sportstacking, SportsMemorabilia, spreadsheet, excel, squash, stampcollecting, stamps, philately, StandUpComedy, Standup, stormchasing, tornado, writing, writingadvice, WritingPrompts, KeepWriting, StoryWriting, fantasywriters, writers, WritingHub, WritingResources, writinghelp, writingopportunities, writingcontests, writingcritiques, WritingStyle, writingprompt, writing\_gigs, Storytelling, Stretching, flexibility, sudoku, BeginnerSurfers, surfing, Survival, preppers, Survivalist, survivalism, UrbanSurvivalism, WildernessBackpacking, Swimming, OpenWaterSwimming, tabletennis, taekwondo, taichi, taijiquan, taoism, Tapestry, tarot, tarotpractice, Tarotpractices, tarotreadings, TarotReading, TarotDecks, Tarots, SecularTarot, TarotCards, tattoo, tattoos, TattooDesigns, shittytattoos, Best\_tattoos, DrawMyTattoo, traditionaltattoos, TattooArt, TattooApprentice, nerdtattoos, tattooscratchers, badtattoos, tattooadvice, sticknpokes, TattooRemoval, TattooVideos, TattooArtists, tattooflash, tattoocomparisons, tattooing, TattooIdeasDesigns, Tattoocoverups, Taxidermy, badtaxidermy, vultureculture, TaxidermyisMetal, tea, TeaPorn, RateMyTea, teaexchange, teasales, TeaPictures, boba, teaching, Teachers, TeachingUK, education, StudentTeaching, AustralianTeachers, teachinginkorea, teachinginjapan, TEFL, ScienceTeachers, ELATeachers, Internationalteachers, MusicEd, OnlineESLTeaching, Professors, tennis, GirlsTennis, TennisCourtPorn, 10s, tabletennis, terrariums, bioactive, ClosedTerrariums, Terrarium, Jarrariums, Mossariums, Thruhiking, AppalachianTrail, PacificCrestTrail, CDT, NCTrails, JMT, ticketbrokers, topiary, tourism, ThailandTourism, irishtourism, india\_tourism, travel, JapanTravel, TourismHell, ItalyTourism, IAmA\_Tourism, CostaRicaTravel, GuiderTravel, darktourism, koreatravel, braziltourism, TRADE, mechmarket, pokemontrades, TakeaPlantLeaveaPlant, hardwareswap, snackexchange, discexchange, AdoptMeTrading, comicswap, Knife\_Swap, Pen\_Swap, AquaSwap, makeupexchange, Miniswap, SteelbookSwap, watch\_swap, KitSwap, mangaswap, trains, TrainPorn, transit, uktrains, MelbourneTrains, nycrail, TrapShooting, travel, solotravel, TravelHacks, TravelPorn, travelpartners, TravelNoPics, Shoestring, digitalnomad, travelphotos, travelblogging, onebag, travelbuddies, travel\_deals, TravelMaps, Travel\_HD, travelhacking, travelblogs, TreasureHunting, triathlon, IronmanTriathlon, ultimate, unicycling, upcycling, urbanexploration, UrbanExploring, Urbex, virtualreality, VRGaming, Vive, vrfit, VRphysics, VRplugins, vegetablegardening, Veganic, WormFarming, projectcar, VideoEditing, editors, VideoEditingRequests, editing, buildapcvideoediting, videography, gamecollecting, RetroGamePorn, Shittygamecollecting, VideoGameCollection, gamedev, GamePhysics, IndieDev, gameDevClassifieds, indiegames, INAT, VideoGameDevelopers, VideoGameEditing, GameDevelopersOfIndia, gamedevscreens, classiccars, vintagejapaneseautos, vintagecars, VintageFashion, VintageTees, VintageClothing, vinyl, VoiceActing, VoiceActingJobs, RecordThisForFree, volleyball, VolleyballGirls, volleyballtraining, volunteer, walking, wargaming, wargames, computerwargames, watchmaking, waterpolo, WaxSealers, malegrooming, weaving, Welding, Weldingporn, WeldPorn, BadWelding, BadWelding, whittling, wildwhittlers, wine, winemaking, witchcraft, realwitchcraft, BabyWitch, occult, magick, Witch, Wicca, apprenticewitches, Woodcarving, Spooncarving, Carving, BeginnerWoodWorking, woodworking, turning, wrestling, prowrestling, Wreddit, WritingPrompts, yoga, YogaWorkouts, YogaTeachers, YogaChallenge, zoos, Zoo\_Pics, zoology, AnimalPorn, zumba \\
\bottomrule
\end{tabular}
\caption{A list of subreddits for \textbf{Hobby} used to construct \textit{NERsocial}. Part II.}
\label{Table:listOfAllSubbredditsHobbyPart2} 
\end{table*}
\section{Stack Exchange Posts}
We gathered texts from these \textit{sites}: \textit{cooking.stackexchange, pets.stackexchange, workplace.stackexchange, alcohol.stackexchange, beer.stackexchange, coffee.stackexchange, crafts.stackexchange, gardening.stackexchange, homebrew.stackexchange, photo.stackexchange, sports.stackexchange, fitness.stackexchange, bicycles.stackexchange}. The above list of \textit{sites} was sufficient to generate a massive corpus ready for annotation. 
\begin{figure*}[!t]
\centering
\includegraphics[width=14.8cm]{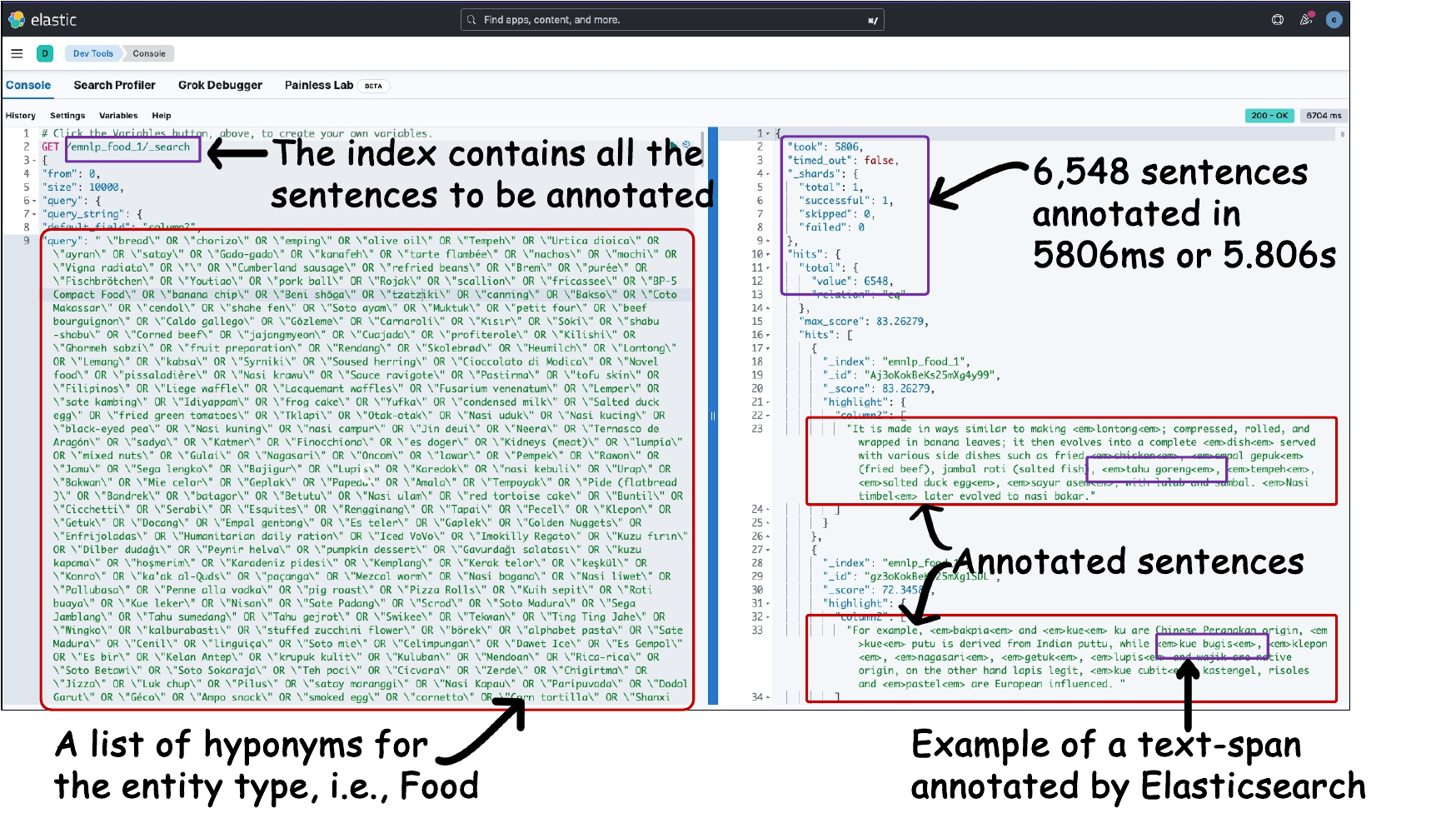}
\caption{To annotate the texts inside ES, thousands of sentences, and the list of hyponyms are uploaded. We introduced $<$em$>$ tags to make all spans of text that \textit{match} with any entry in the list of hyponyms on the left.  We can see anntoatetd spans of text on the right.}
\label{fig:ElasticAnnotationProcess}
\end{figure*}
\section{Annotation Scheme based on Elasticsearch}
\label{section:Elasticsearch}
With Elasticsearch (ES)\footnote{We used Elasticsearch version 8.14.0}., we sped up the annotation process, significantly reducing the burden incurred with manual annotation of sentences. In fact, our ES annotation scheme costs 0.9 milliseconds to annotate one sentence, yet a manual annotation method, such as the use of \textit{Doccano} costs 1 minute per sentence. This results in massive savings of time and labor costs during annotation.\\
\textbf{Import data into ES.} Thousands of sentences saved inside a \texttt{CSV file} were imported into ES. 
For each file uploaded to ES, an \texttt{index} is created in ES. It is vital to create an appropriate \texttt{mapping} for the data while uploading it to ES. We needed to use \texttt{dynamic mapping} which supports a \texttt{fast vector highlighter (fvh)}. By selecting the \texttt{Advanced} option at data import, we customized the \texttt{mappings} for the columns in the data. An example of such a \texttt{mapping} for the data containing \textbf{Food} sentences is shown below:
\begin{lstlisting} [basicstyle=\small] 
{
 "properties": { 
 "column1": { "type": "long"},
 "column2": { "type": "text", 
  "term_vector":"with_positions_offsets"}
    }
}
\end{lstlisting}
In this \texttt{mapping}, \texttt{column1} is of type \texttt{long} and it contains the unique ID for a sentence. However, \texttt{column2} is of type \texttt{text}, and it contains the actual sentences to be annotated. To apply the \texttt{fast vector highlighter} to \texttt{column2}, we set the argument \texttt{term\_vector} to \texttt{with\_positions\_offsets}. 
\paragraph{Fast vector highlighter (fvh).}
ES constitutes a unique feature called \texttt{fast vector highlighter} or \texttt{fvh}. This is a special \texttt{highlighter} in ES. The \texttt{fvh} highlighter behavior is enabled by setting \texttt{"term\_vector":"with\_positions\_offsets"} on the appropriate column of the data (in this case, column 2). The \texttt{fvh} highlighter makes sure that at the time of sentence retrieval, compound terms for example, \textit{Barton Premium Blend}, are not split into their constituent words \textit{Barton},  \textit{Premium}, and \textit{Blend}. The \texttt{fvh} highlighter made it possible to achieve perfect string matching.
\\
\textbf{ES Search Space.} Given a NE, we defined the search space for ES as the \textit{dictionary} terms obtained from Wiki-KG for that NE. These terms enabled ES to search, and retrieve sentences in which one, two or more terms \textit{matched} the search space of ES. In other words, we utilized ES as a search engine to return only relevant sentences. Besides, we can deduce that ES search space corresponds to \textit{entity mentions} for a specific NE.
\\
\textbf{Annotation with ES.} We employed the \texttt{fvh} highlighter for annotation in three ways. 
First, \textit{fvh} ensured that \textit{compound} terms were not split into separate words at search time. This is vital because it allowed us to annotate compound terms accurately. Second, via correct string matching between ES search terms and words in sentences, \texttt{fvh} enabled us to demarcate text-spans within each sentence that \textit{matched} with one or more terms within the search space. Third, we employed this feature to introduce special tags \texttt{<em>} to indicate such text-spans or \textit{entity mentions} in each sentence. In other words, ES was used for \textit{labeling/annotating} all \textit{entity mentions} in each sentence.
ES made it possible to mark text-spans for thousands of sentences at once. All annotations were thoroughly checked and corrected by annotators who possessed ample linguistic knowledge.
Figure \ref{fig:ElasticAnnotationProcess} shows different sections of an ES window during annotation. 
\section{Knowledge Graph to Domain-specific Sub-graphs}
\begin{figure}[!t]
\centering
\includegraphics[width=7.5cm]{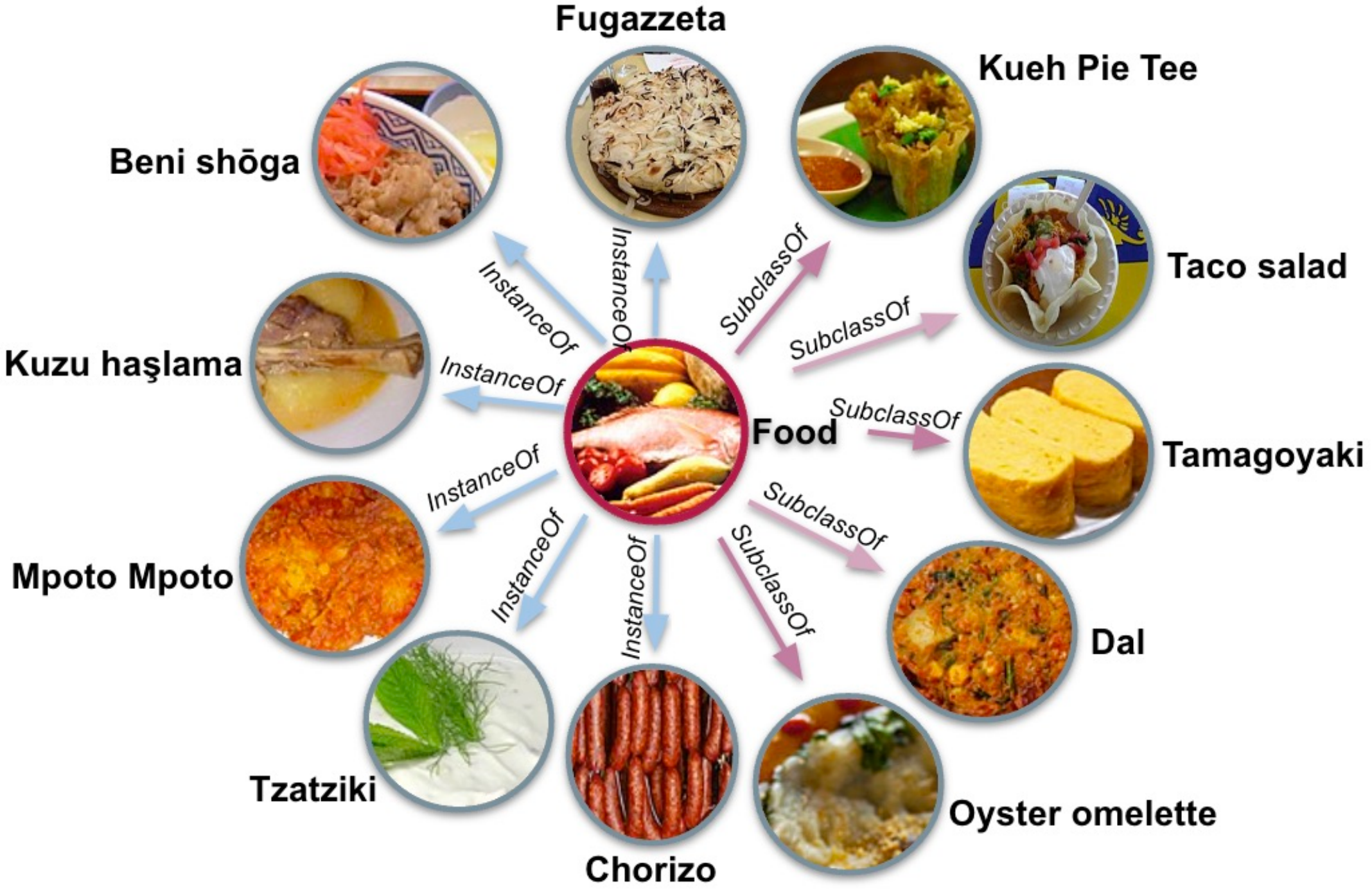}
\caption{We deploy \textit{instance-of}, and \textit{subclass-of} to extract sub-graphs of each entity type from Wikidata.}
\label{fig:WikiSubgraphFood}
\end{figure}

We show the detailed process to collect triples and extract sub-graphs like the one in Figure \ref{fig:WikiSubgraphFood}. 
\begin{table*}[!t]
\centering
\small
\begin{tabular}{ll}
\toprule
\textbf{Property} & \textbf{Description} \\
\midrule
instance-of & that class of which this subject is a particular example and member (subject typically\\
&  an individual member with a proper name label), different from P279 \\
\midrule
subclass-of & all instances of these items are instances of those items; this item is a class (subset) \\
& of that item. Not to be confused with P31 (instance-of).\\
\bottomrule
\end{tabular}
\caption{\label{Table: Wikidata properties}Descriptions of two ${G}_{Wiki}$ properties.}
\end{table*}
\subsection{Wikidata properties}
\label{Appendix:WikipropertiesBoth}
\begin{figure}[!ht]
\centering
\includegraphics[width=0.65\columnwidth]{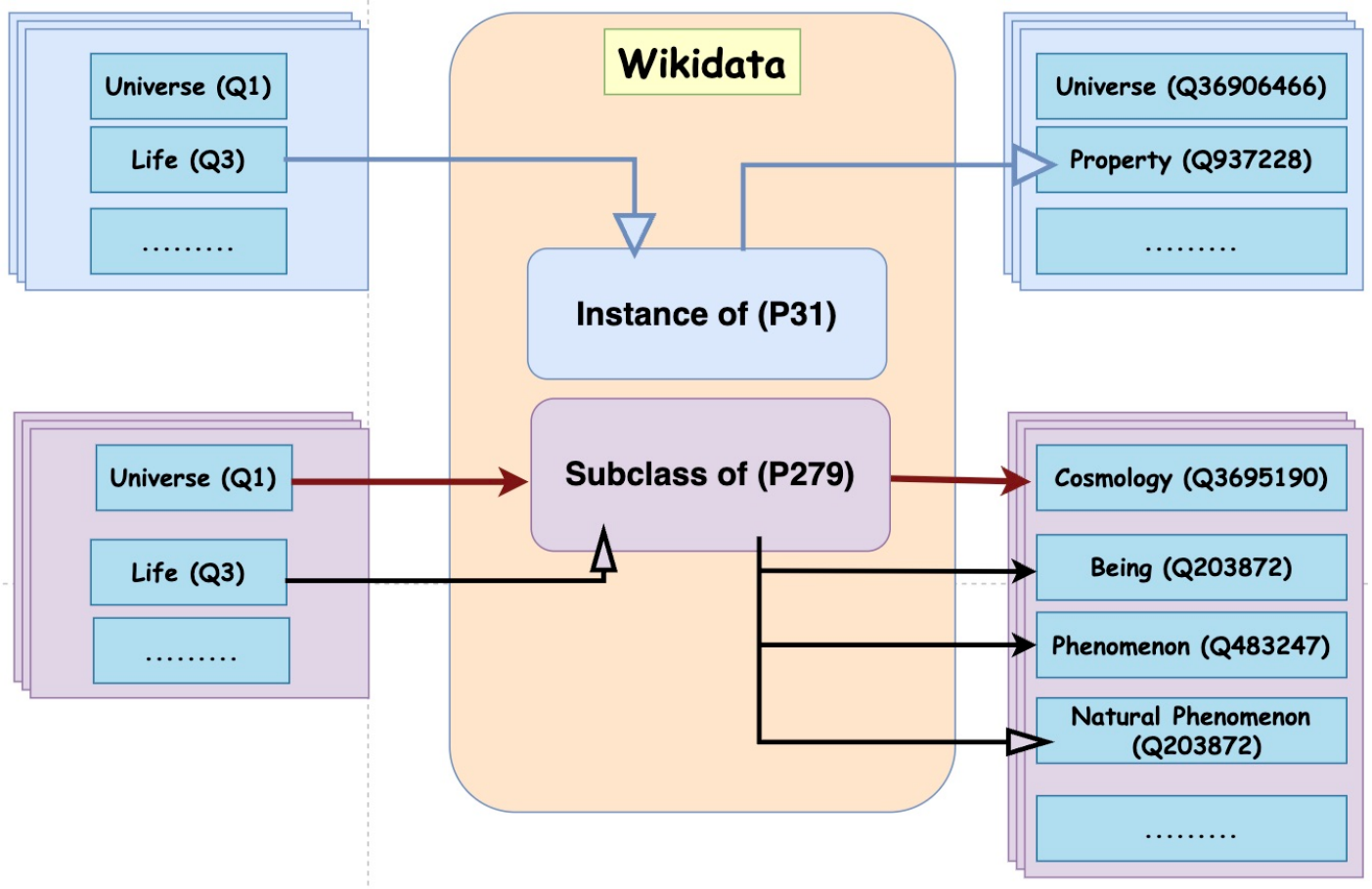}
\caption{Illustration of \textit{items, properties, and statements} in Wikidata. }
\label{fig:KDWD_WikipropertiesBoth}
\end{figure}

The illustration in Figure~\ref{fig:KDWD_WikipropertiesBoth} introduces main concepts in Wiki-KG. In this figure, the list of \textbf{items} and their unique Wiki-KG \textit{ID}s includes: \textit{Universe (Q1), Life(Q3), Cosmology (Q3695190), Property (Q937228), Being (Q203872), Phenomenon (Q483247), Natural Phenomenon (Q203872).} The list of Wiki-KG \textbf{properties} includes \textit{instance-of} and \textit{subclass-of}. The \textbf{statements} indicate relations between items in Wiki-KG e.g., \textit{``Life (Q3) is subclass-of (P279) a Natural Phenomenon (Q203872)''}. In general the statements, hereafter Wiki-KG triples, are of the form \texttt{[source item ID, edge property, target item ID]}.
In this work, we employed only two properties, namely: \textit{instance-of (i.e., P31)} and \textit{subclass-of (i.e., P279)}. Because, (i) the properties define a sufficiently large number of Wiki-KG relations. The relations allowed us to generate large dictionary sizes for each target NE. (ii) We aim to retrieve as many \texttt{child nodes} for each \texttt{parent node}. The \texttt{parent node} is the NE. \textit{subclass-of \& instance-of} permitted us to gather such \texttt{child nodes}, as illustrated in Figure \ref{fig:WikiSubgraphFood}. Both \textbf{properties} are described in Table~\ref{Table: Wikidata properties}. All properties are contained in KDWD's \texttt{property.csv} file. 
\subsection{Items in Wikidata}
\label{Appednix:WikidataItems}
Table \ref{Table: Sample of KDWD item file} shows definitions of some items in Wikidata. The definitions are precise, helping to distinguish between several items in Wikidata.
\begin{table*}[!t]
\footnotesize
\centering
\begin{tabular}{lll} 
\toprule
\textbf{Item ID} & \textbf{ Label} & \textbf{Description} \\
\midrule
1  &  Universe   & totality of space and all contents\\
2  &  Earth & third planet from the Sun in the Solar System \\
5  &   Human & any member of Homo sapiens, unique extant species of the genus Homo, from embryo to adult \\
\bottomrule
\end{tabular}
\caption{Sample of entries in \texttt{item.csv} file of KDWD dataset.}
\label{Table: Sample of KDWD item file}
\end{table*}
\section{Collection of Wiki-KG Dictionaries}
\label{Appendix:Collection of Wikidata Dictionaries}
After filtering out triples, the triples were used to gather dictionaries from Wiki-KG via 5 steps. All the triples are defined by two relations \textbf{P31\&P279}. We selected only those triples in which \textit{Target Item ID} correspond to \textit{Item ID} of the \textit{item} of interest. For all remaining triples, the \textit{Source Item ID}s were extracted. At last, via a \textit{join} operation with \texttt{item.csv} we got the \textit{English label}s. This collection of \textit{English label}s formed the dictionary for a NE. These procedures are detailed below.

In KDWD's \texttt{item.csv} file, there exists 51M items, similar to those shown in Table~\ref{Table: Sample of KDWD item file}. Each entry has three attributes, i.e., \textit{Item ID}, \textit{ English Label}, \textit{English Description}. Dictionaries were obtained as follows. \textbf{1)} Given a NE such as \textbf{Food}, we searched through the \texttt{item.csv} file by setting search condition \textit{``English Label == Food''}. This enabled us to get the \textit{Item ID} of \textbf{Food} as \textbf{2095}. \textbf{2)}  We filtered through the Wiki-KG triples (which were collected in the previous section), and retained only those triples in which the \textit{Target Item ID} is 2095. \textbf{3)}  We extracted all the \textit{Source Item ID} in each triple. 1,365 \textit{Source Item ID} were obtained for \textbf{P31} triples, while 2,884 \textit{Source Item ID} were obtained for \textbf{P279} triples. \textbf{4)}  All the \textit{Source Item ID} were collected and by matching them with the entries in \texttt{item.csv} file, we got the set of \textit{English Label} for all \textit{Source Item ID}. This set formed the dictionary for \textbf{Food}. Examples of terms in the dictionary for \textbf{Food} are shown in Table~\ref{Table:DictionariesforNEsTwo}. \textbf{5)}  We regarded the dictionary obtained for each NE as the set of \textit{entity mentions} for that NE. In short, this dictionary formed the search space/entity mentions for that NE in ES during annotation.

We report the size of dictionaries obtained during dataset curation, for each NE in Table~\ref{Table: Size of Dictionaries}. In addition, we describe unique steps used to obtain dictionaries for NE. For these NE: \textit{Drink, Food, Job}, \textbf{P31} and \textbf{P279} were sufficient to generate a large number of elements in their dictionaries. Yet, for \textit{Hobby, Pet, Sport}, we augmented the dictionaries due to a small number of examples obtained from Wiki-KG. 

The following describes dictionary augmentation for the NE.
\textbf{Hobby:} We added more \textit{Hobby} examples from a Wikipedia article\footnote{A Wikipedia list of hobbies \url{https://en.wikipedia.org/wiki/List_of_hobbies}}. \textbf{Job:} Due to \textit{Job} and \textit{Profession} being synonymous with each other, we use the union set of elements contained in the dictionaries of \textit{Job} and \textit{Profession}. \textbf{Pet:} The unique term \textit{Pet} returned no dictionary items from Wikidata so we combined several dictionaries e.g., for \textit{dogs, cats, horses, rodents, domesticated birds} to obtain a dictionary size sufficient to guide the annotation process for \textit{Pet}. \textbf{Sport:} To ensure no overlap between \textit{Hobby} and \textit{Sport}, all items under \textit{Hobby} identifiable as \textit{Sport} were deleted. Moreover, we couldn't get a significant dictionary size from Wikidata based on \textbf{P31} and \textbf{P279} so we added more \textit{Sport} examples from a Wikipedia article\footnote{A Wikipedia list of sports \url{https://en.wikipedia.org/wiki/List_of_sports}} and one sports website~\footnote{This website lists popular sports \url{https://trenzl.com/sports/}}.

Examples of dictionary elements for each NE are shown in Table~\ref{Table:DictionariesforNEsTwo}. Furthermore, Table \ref{Table: Size of Dictionaries} shows the size of each entity type's dictioanry.
\begin{table*}[!t]
\small
\centering
\begin{tabular}{*6c}
\toprule
\textbf{Named}  & \multicolumn{2}{c}{\textbf{KG Triple Size}} & \multicolumn{2}{c}{\textbf{{Dictionary} Size}} & \textbf{\#Wikipedia}\\
\textbf{Entity} & (26M) & (1.7M)    &    &  &\textbf{Pages}  \\
                & \textbf{P31} & \textbf{P279}  &  \textbf{P31/ P279} & \textbf{Total}   &   \\
\midrule
\textbf{Drink}   & 281   & 309    & 249 / 280 & 529 & 451\\
\textbf{Food}    &  1,365 & 2,884  & 321 / 2,683 & 3,004 & 2,613\\
\textbf{Hobby}  &  35 & 13 & 28 /12 & 40 & 24\\
\textbf{Job/Profession}  &  6,475  &  82   & 4,126 / 61 & 4,187 & 1,527\\
\textbf{Pet}  &  5  &  10   & 5 / 10 & 15 & 11\\
\textbf{Sport}  &  164  &  120   & 137 / 103 & 240 & 160\\
\bottomrule
\end{tabular}
\caption{
Size of dictionaries used to construct \emph{RapidNER}, and the number of Wikipedia pages. For NE with few pages gathered, we collected extra pages using authentic lists relevant to such NE in Wikipedia.
}
\label{Table: Size of Dictionaries}
\end{table*}
\begin{table*}[!t]
\centering
\footnotesize
\begin{tabular}{p{0.95\textwidth}}  %
\toprule
\textbf{Examples of hyponyms obtained from Wikidata based on \textit{instance-of} and \textit{subclass-of}. } \\
\midrule
\textbf{Entity Type: \textcolor{purple}{Drink}.} Fruitopia, 
Actimel, sima, Posca, Rivella, Aquafina, Brause, Dortmunder Hansa, Tereré, Cole Cold, Qoo, Chun Mee tea, Arabic coffee, Tequila Sunrise, Cajuína, mors, ochsenblut, White Horse, 
Paddy Whiskey, gorbatschow, Black drink, Apotekarnes Cola, Perú Cola, 
Blantons, Clamato, coco, sahti, Everclear, Dewars, \\
\midrule
\textbf{Entity Type: \textcolor{purple}{Food}.} Tempeh, ayran,  Urtica dioica,  satay, Gado-gado, kanafeh, 
tarte flambée, nachos, mochi, Vigna radiata, 
Cumberland sausage, refried beans 
Brem, purée, Fischbrötchen, Youtiao, pork ball, Rojak, scallion, fricassee, 
BP-5 Compact Food, banana chip,Beni shōga, tzatziki, canning, Bakso, Coto Makassar, 
cendol, shahe fen, Soto ayam, Muktuk, petit four, beef bourguignon, Caldo gallego, \\
\midrule
\textbf{Entity Type: \textcolor{purple}{Hobby}.} fishkeeping, amateur radio, Tyrosemiophilia, Sucrology 
Antenne Bergstraße, birdwatching, DXing, model building, garage kit, filmmaker 
, philatelist, Squirrel fishing, , book collecting, video blogger, 
Butterfly watching, Element collecting, MW DX, mermaiding, mineral collecting, 
Sneaker collecting, modelling with clay, sardana dancer, Cynophilia, \\
\midrule
\textbf{Entity Type: \textcolor{purple}{Job}.} mangaka, diplomat, 
elevator operator, machinist, kit manager, privateer, 
historian, sniper, referee, greenskeeper, veterinarian
Combat Medic Specialist, Zimmerpolier, invent 
Civilian employee, executioner, 
psychiatrist,  goldsmith, civil servant, psychologist, freelancer, merchant, 
chauffeur, chauffeuse, qadi, customs broker, Zollrevis, croupier, missionary, \\
\midrule
\textbf{Entity Type: \textcolor{purple}{Pet}.} Angey, Mrs. Chippy, Catmando, Tama, Dewey Readmore Books, 
Socks, F. D. C. Willard, Nora, Hodge, Humphrey, 
Towser, Wilberforce, India, Simon, Larry 
Winnie, Creme Puff, Sybil, Maru, Mr. Green Genes, 
Trim, Scarlett, Casper, Choupette, Dusty the Klepto Kitty, Fred the Undercover Kitty, 
Freya, Heed, Henri, Little Nicky, Luna the Fashion Kitty, Meow, Mike, Mr. Nuts, \\
\midrule
\textbf{Entity Type: \textcolor{purple}{Sport}.} Aerial dance, Indiaca,  playboating, synchronized diving, Hornussen, 
bull-leaping, extreme ironing, tamburello 
Camel racing, Buzkashi, motorcycle speedway, Ancient Greek boxing, Formula 5000, 
Chinlone, Woodball, artistic fencing, Indiaca, rally raid, Paralympic association football, 
bodyboarding, powerbocking, Water Jousting, Professional baseball, Brännboll, \\
\bottomrule
\end{tabular}
\caption{An overview of dictionaries extracted from Wikidata, for six named entities, \textbf{Drink, Food, Hobby, Job, Pet, and Sport}  during the construction of \emph{NERsocial}.}
\label{Table:DictionariesforNEsTwo}
\end{table*}
\section{Selecting Wikipedia Paragraphs}
\label{Appendix: Selection of Wikipedia Paragraphs}
Given the \textit{Item ID} for all \textit{Items} in the dictionary of NE such as \textbf{Food}, we employ those \textit{Item ID} and obtain the corresponding Wikipedia \textit{Page} for each \textit{Item} as follows.

In KDWD's \texttt{page.csv} file, there exists 5.3M entries in the form shown in Table~\ref{Table: Sample of KDWD pages file}. Each entry has four attributes, i.e., \textit{Page ID}, \textit{Item ID}, \textit{Title}, \textit{Views}. 5.3M entries correspond to the 5.3M articles in English Wikipedia as of November 2019.
The collection of \textit{Item ID} obtained from the previous section permitted us to acquire \textit{Page ID}s. Each \textit{Page ID} refers to a unique article within Wikipedia. Hence, we can obtain the Wikipedia page and its contents.

\section{Why we filter out other sections of the Wikipedia Article?}
\label{Appendix:filterWikipediaArticles}
\begin{figure*}[!t]
\centering
\includegraphics[width=\textwidth]{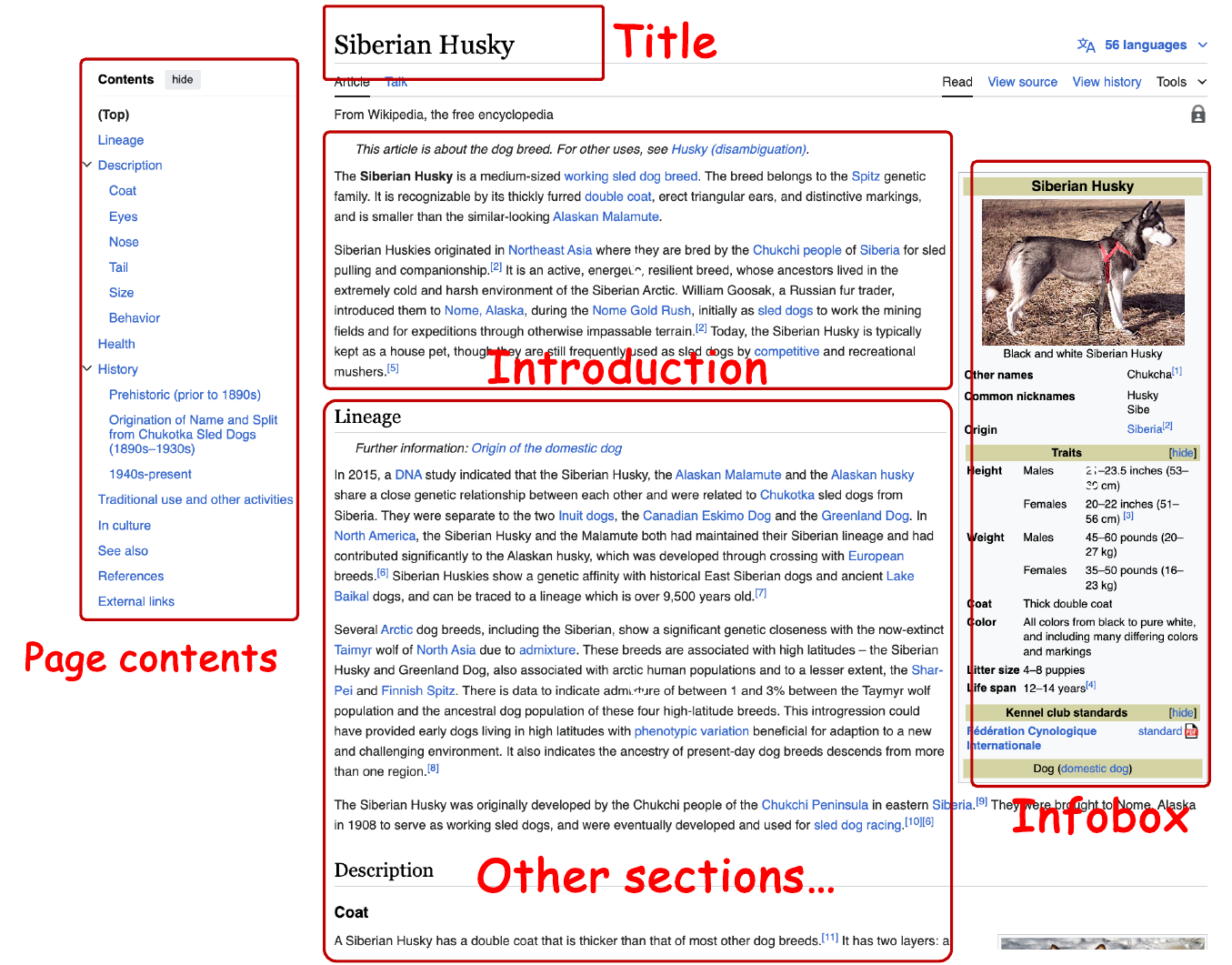}
\caption{Illustration of main sections of a Wikipedia article. We are interested in the \textbf{title} and \textbf{introduction} of the article. }
\label{fig:WikipediaPageSections}
\end{figure*}

We aim to extract the most concise information from Wikipedia. The \texttt{Wikipedia Manual of Style}\footnote{URL for the Wikipedia Manual of Style \url{https://en.wikipedia.org/wiki/Wikipedia:Manual_of_Style}.} recommends that \textit{``An article's content should begin with an introductory lead section – a concise summary of the article – which is never divided into sections''}. Hence, by design, the introductions are intended to capture the essence of a Wikipedia article. Moroever, Wikipedia introductions have previously been used as query-relevant summaries or ground truth for evaluating the quality of summaries in information retrieval and text summarization research \cite{svore-etal-2007-enhancing}. Lastly, we aim for a wider distribution of entities in our dataset. Hence, we selected the most precise section of the Wikipedia article, and this decision enabled us to collect more texts from millions of other articles. 
Figure \ref{fig:WikipediaPageSections} indicates the \textit{title} and \textit{introduction} sections which were paramount during dataset construction. 
\begin{table}[!t]
\small
\centering
\begin{tabular}{*4c}
\toprule
\textbf{Page ID} & \textbf{Item ID} & \textbf{Title} & \textbf{Views} \\
\midrule
12  &  6199   & Anarchism & 31335\\
25  &  38404   & Autism & 49693\\
\bottomrule
\end{tabular}
\caption{\label{Table: Sample of KDWD pages file}
Sample of entries in \texttt{page.csv} file of KDWD dataset. }
\end{table}

\begin{table*}[!t]
\centering
\resizebox{0.985\textwidth}{!}{
\begin{tabular}{|c||c|c|c||c|c|c||c|c|c||c|c|c||c|c|c||c|c|c||c|}
\hline
   &  \multicolumn{3}{|c|}{ {\bf Drink}} &  \multicolumn{3}{|c|}{ {\bf Food} }  &  \multicolumn{3}{|c|}{ {\bf Hobby} } &  \multicolumn{3}{|c|}{ {\bf Job} } &  \multicolumn{3}{|c|}{ {\bf Pet} } &  \multicolumn{3}{|c|}{ {\bf Sport} } & \\ \hline
        \textbf{NER}  &&&&&&&&&&&&&&&&&&&   \\ \hline
\textbf{Model} & \textbf{P} & \textbf{R} & \textbf{F1} & \textbf{P} & \textbf{R} & \textbf{F1} & \textbf{P} & \textbf{R} & \textbf{F1} & \textbf{P} & \textbf{R} & \textbf{F1} & \textbf{P} & \textbf{R} & \textbf{F1} & \textbf{P} & \textbf{R} & \textbf{F1} & \textbf{Avg. (F1)} \\ \hline
\multicolumn{20}{|c|}{ \textcolor{blue}{I. Models trained on Wikipedia texts.} } \\ \hline
\multicolumn{20}{|c|}{ \textbf{In domain setting.} } \\ \hline \hline
BERT-base & 81.34 & 79.93 & 80.63 & 72.81 & 81.37  & 76.85 & 74.06 & 84.4 & 78.96 &  83.56 & 87.32 & 85.39 &  86.11 & 90.45 & 88.23 & 85.22 & 85.90 & 85.56 & 81.61 \\ \hline
RoBERTa-base & 80.07 & 82.01 & 81.03 & 76.16 & 80.49 & 78.27 & 78.79 & 85.53 & 82.02 & 84.60 & 88.72 & 86.61 & 84.39 & 84.62 & 84.50 & 84.45 & 86.97 &  85.71 & 82.29 \\ \hline
DeBERTa-v3-base & 87.81 & 87.19 & 87.50 & 78.12 & 79.12 & 78.62 & 79.68 & 82.57 & 81.09 & 82.17 & 89.58 & 85.71 & 85.43 & 91.78 & 88.49 & 84.59 & 86.17 & 85.38 & 83.29 \\ \hline
\multicolumn{20}{|c|}{ OOD setting 1: Social Media } \\ \hline
BERT-base    & 20.01 & 24.50 & 22.03 & 37.74 & 59.44 & 46.17 & 26.58 &  15.56 & 19.63 &  62.18 & 76.42 & 68.57 & 22.69 & 16.65 & 19.21 & 68.78 & 74.62 & 71.58 & \textcolor{red}{49.76} \\ \hline
RoBERTa-base & 33.75 & 58.99 & 42.94 & 36.62 & 63.61 & 46.48 & 40.55 & 23.36 & 29.65 & 63.76 & 75.39 & 69.09 & 26.87 & 25.78 & 26.31 & 65.08 & 77.35 & 70.68 & \textcolor{red}{51.61} \\ \hline
DeBERTa-v3-base & 26.64  & 23.75 & 25.11 & 24.87 & 23.63 & 24.24 & 36.72 & 18.24 & 24.37 & 51.81  & 42.51 & 46.71 & 18.88 & 10.51 & 13.51 & 65.01 & 57.89 & 61.25 & \textcolor{red}{33.11} \\ \hline
\multicolumn{20}{|c|}{ OOD setting 2: Online Forum  } \\ \hline
BERT-base & 36.05 & 41.93 & 38.77 & 37.74 & 46.85 & 51.07 & 48.87 &  36.87 & 23.32 &  28.57 & 63.77 & 76.05 & 69.37 & 52.49 & 15.56 & 24.00 & 71.66 & 69.69 & \textcolor{red}{70.66} \\ \hline
RoBERTa-base & 49.01  & 75.56 & 59.46 & 49.00 & 55.57 & 52.08 & 45.91 & 24.12 & 31.62 & 66.93 & 75.06 & 70.76 & 62.10 & 26.62 & 37.26 & 70.65 & 76.87 &  73.63 & \textcolor{red}{54.27} \\ \hline
DeBERTa-v3-base & 46.21 & 47.72 & 46.96 & 39.08 & 32.94 & 35.75 & 35.16 & 19.66 & 25.22 & 63.29 & 57.09 & 60.03 & 65.79 & 22.83 & 33.89 & 68.93 & 57.40 & 62.64 & \textcolor{red}{42.56} \\ \hline
\multicolumn{20}{|c|}{ \textcolor{blue}{II. Models trained on Reddit texts.} }  \\ \hline
\multicolumn{20}{|c|}{ \textbf{In domain setting.} }  \\ \hline \hline
BERT-base & 95.39 & 92.79 & 94.07 & 95.57 & 97.37  & 96.46 & 96.17 & 93.10 & 94.62 &  96.36 & 97.31 & 96.83 &  97.89 & 99.09 & 98.49 & 98.67 & 97.96 & 98.31 & 96.65 \\ \hline
RoBERTa-base & 92.61  & 95.32 & 93.95 & 95.39 & 95.88 & 95.64 & 95.52 & 94.24 & 94.87  & 94.91 & 95.66 & 95.29 & 97.88 & 98.38 & 98.13 & 96.10 & 97.96 & 97.02 & 95.78\\ \hline
DeBERTa-v3-base & 93.03 & 96.29 & 94.64 & 93.65 & 95.96 & 94.79 & 93.82 & 95.27 & 94.54 & 97.27 & 97.23 & 97.25 & 96.74 & 99.29 & 97.99 & 96.02 & 98.47 & 97.23 & 95.97 \\ \hline
\multicolumn{20}{|c|}{ OOD setting 1: Wikipedia} \\ \hline
BERT-base    & 47.73 & 4.15 & 7.64 & 50.66 & 31.15 & 38.57 & 32.53 &  26.76 & 29.37 &  70.99 & 35.57 & 47.39 & 54.14 & 18.22 & 27.27 & 76.41 & 39.11 & 51.74 & \textcolor{red}{36.79} \\ \hline
RoBERTa-base & 43.34 & 09.13 & 15.09 & 59.95 & 32.86 & 42.45 & 38.50 & 33.95 & 36.08 & 71.05 & 40.06 & 51.23 & 61.28 & 26.63 & 37.13 & 71.35 & 52.54 & 60.52 & \textcolor{red}{42.68} \\ \hline
DeBERTa-v3-base & 11.52 & 05.77 & 07.69 & 25.69 & 24.07 & 24.85  &  26.18 & 36.08 & 30.35 & 45.05 & 26.41 & 33.30 & 28.66 & 23.24 & 25.67 & 50.41 & 45.96 &  48.08 & \textcolor{red}{28.76} \\ \hline
\multicolumn{20}{|c|}{ OOD setting 2: Online forum}  \\ \hline
BERT-base    & 47.73 & 4.15 & 7.64 & 50.66 & 31.15 & 38.57 & 32.53 &  26.76 & 29.37 &  70.99 & 35.57 & 47.39 & 54.14 & 18.22 & 27.27 & 76.41 & 39.11 & 51.74 & \textcolor{red}{36.79} \\ \hline
RoBERTa-base & 91.84 & 92.88 & 92.35 & 87.73 & 95.73 & 91.56 & 75.24 & 60.44 & 67.03 & 92.57  & 97.93 & 95.17 & 98.83 & 92.91 & 95.78 & 92.08 & 94.49 &  93.27 &  \textcolor{red}{89.17} \\ \hline
DeBERTa-v3-base & 87.08 & 95.04 & 90.89 & 86.47 & 95.28 & 90.66 & 73.04 & 60.92 & 66.43 & 94.54 & 97.11 & 95.81 & 96.19 & 96.05 & 96.12 & 92.12 & 94.98 & 93.53 & \textcolor{red}{88.57} \\ \hline
\multicolumn{20}{|c|}{ \textcolor{blue}{III. Models trained on Stack Exchange texts. } } \\ \hline
\multicolumn{20}{|c|}{ \textbf{In domain setting.} } \\ \hline \hline
BERT-base    & 97.72 & 99.16 & 98.44 & 96.96 & 98.26 & 97.61 & 98.12 &  95.91 & 97.00 &  98.65 & 98.56 & 98.61 & 99.69 & 99.84 & 99.77 & 96.25 & 98.61 & 97.42 & 98.16 \\ \hline
RoBERTa-base & 97.83 & 98.61 & 98.22 & 96.59 & 98.13 & 97.36 & 98.49 & 95.19 & 96.81 & 98.61 & 98.34 & 98.47 & 99.42 & 99.69 & 99.55 & 97.72 & 98.20 & 97.96 & 98.03 \\ \hline
DeBERTa-v3-base & 97.84 & 98.87 & 98.35 & 96.50 & 98.00 & 97.25 & 98.72 & 95.64 & 97.16 & 98.26 & 98.74 & 98.50 & 99.11 & 99.96 & 99.53 & 97.18 & 98.45 & 97.81 & 98.07 \\ \hline
\multicolumn{20}{|c|}{ OOD setting 1: Wikipedia}  \\ \hline 
BERT-base    & 51.98 & 14.51 & 22.69 & 55.06 & 29.19 & 38.15 & 39.69 &  33.59 & 36.39 &  75.01 & 31.26 & 44.13 & 58.65 & 23.91 & 33.97 & 67.87 & 46.00 & 54.84 & \textcolor{red}{39.16} \\ \hline
RoBERTa-base & 63.87 & 18.94 & 29.22 & 60.67 & 29.77 & 39.94 & 48.01 & 32.49 & 38.76 & 76.62 & 32.79 & 45.93 & 63.69 & 30.36 & 41.12 & 72.69 & 42.17 & 53.37 & \textcolor{red}{41.69} \\ \hline
DeBERTa-v3-base & 28.49 & 14.56 & 19.27 & 24.11 & 19.15 & 21.35 & 29.67 & 35.59 & 32.36 & 39.29 & 24.75 & 30.37 & 28.68 & 22.84  & 25.43 & 49.45 & 38.99 & 43.61 & \textcolor{red}{27.61} \\ \hline
\multicolumn{20}{|c|}{ OOD setting 2: Social media } \\ \hline 
BERT-base    & 86.65 & 91.21 & 88.87 & 92.31 & 90.06 & 91.17 & 72.22 &  51.83 & 60.35 &  91.87 & 91.76 & 91.81 & 94.23 & 97.39 & 95.78 & 93.01 & 96.53 & 94.73 & \textcolor{red}{89.71} \\ \hline
RoBERTa-base & 91.27 & 88.71 & 89.97 & 90.18 & 88.72 & 89.44 & 75.41 & 48.19 & 58.80 & 92.28 & 91.47 & 91.87 & 94.04 & 96.35 & 95.18 & 91.90 & 93.22 & 92.56 & \textcolor{red}{88.79} \\ \hline
DeBERTa-v3-base & 90.59 & 90.25 & 90.42 & 85.00 & 88.65 & 86.79 & 69.75 & 48.75 & 57.39 & 89.04 & 89.51 & 89.27 & 91.89 & 97.69 & 94.71 & 90.64 & 92.51 & 91.56 & \textcolor{red}{86.55} \\ \hline
\multicolumn{20}{|c|}{ \textcolor{blue}{IV. Models trained on combined StackExch+Wiki+Reddit texts. } } \\ \hline
\multicolumn{20}{|c|}{ \textbf{In domain setting.} } \\ \hline \hline
BERT-base       & 96.50 & 96.92 & 96.71 & 94.53 & 95.95 & 95.24 & 96.37 &  94.98 & 95.67 &  96.12 & 97.34 & 96.73 & 98.67 & 98.62 & 98.65 & 95.62 & 97.69 & 96.65 & 96.41  \\ \hline
RoBERTa-base    & 96.97 & 96.39 & 96.68 & 93.04 & 95.57 & 94.29 & 95.23 & 95.06 & 95.14 & 96.19 & 96.72 & 96.45 & 98.29 & 98.59  & 98.45 & 95.82 & 97.27 & 96.54 & 95.96    \\ \hline
DeBERTa-v3-base & 97.04 & 97.01 & 97.03 & 94.00 & 95.74 & 94.86 & 93.69 & 95.36 & 94.53 & 96.85 & 97.39 & 97.12 & 98.09 & 98.52 & 98.31 & 95.39 & 96.44 & 95.91 & 96.13    \\ \hline
\multicolumn{20}{|c|}{ OOD setting 1: Wikipedia}  \\ \hline 
BERT-base       & 97.94 & 97.55 & 97.74 & 97.26 & 97.56 & 97.41 & 97.80 & 98.10 & 97.95 &  97.65 & 98.49 & 98.07 & 99.31 & 98.18 & 98.74 & 97.56 & 98.15 & 97.86 & \textcolor{green}{97.86}   \\ \hline
RoBERTa-base    & 98.05 & 97.31 & 97.68 & 94.18 & 97.28 & 95.71 & 95.66 & 98.15 & 96.89 & 97.64 & 98.65 & 98.14 & 98.61 & 98.69 & 98.65 & 97.84 & 98.15 & 97.99 & \textcolor{green}{97.15}   \\ \hline
DeBERTa-v3-base & 87.37 & 95.95 & 91.46 & 90.23 & 91.14 & 90.68 & 91.39 & 96.74 & 93.99 & 94.65 & 96.71 & 95.67 & 92.54 & 97.32 & 94.87 & 91.06 & 94.98 & 92.98 & \textcolor{green}{92.92}   \\ \hline
\multicolumn{20}{|c|}{ OOD setting 2: Reddit}  \\ \hline 
BERT-base        & 99.12 & 98.96 & 99.04 & 99.22 & 99.45 & 99.33 & 98.79 &  98.53 & 98.66 &  99.60 & 99.69 & 99.65 & 99.71 & 99.86 & 99.78 & 99.37 & 99.62 & 99.49 & \textcolor{green}{99.40} \\ \hline
RoBERTa-base    & 98.98 & 97.04 & 98.00 & 98.45 & 98.82 & 98.64 & 98.16 & 97.85 & 98.00 & 99.51 & 99.32 & 99.41 & 99.53 & 99.65 & 99.59 & 98.89 & 99.39 & 99.14 & \textcolor{green}{98.90}   \\ \hline
DeBERTa-v3-base & 98.74 & 97.20 & 97.96 & 98.19 & 98.83 & 98.51 & 97.22 & 98.15 & 97.69 & 99.08 & 99.26 & 99.17 & 99.31 & 99.43 & 99.37 & 98.51 & 99.03 & 98.77 & \textcolor{green}{98.68}    \\ \hline
\multicolumn{20}{|c|}{ OOD setting 3: Stack Exchange}  \\ \hline 
BERT-base       & 99.52 & 99.78 & 99.65 & 99.59 & 99.85 & 99.72 & 99.69 &  99.15 & 99.42 &  99.75 & 99.87 & 99.81 & 99.96 & 99.92 & 99.95 & 99.62 & 99.85 & 99.74 & \textcolor{green}{99.71}   \\ \hline
RoBERTa-base    & 99.33 & 99.26 & 99.29 & 98.88 & 99.76 & 99.32 & 99.44 & 98.49 & 98.97 & 99.78 & 99.79 & 99.79 & 99.91 & 99.92 & 99.91 & 99.34 & 99.69 & 99.52 & \textcolor{green}{99.45}    \\ \hline
DeBERTa-v3-base & 99.31 & 99.12 & 99.21 & 98.55 & 99.62 & 99.08 & 99.13 & 98.74 & 98.93 & 99.69 & 99.79 & 99.74 & 99.88 & 99.88 & 99.88 & 99.38 & 99.48 & 99.43 & \textcolor{green}{99.35}    \\ \hline
\end{tabular}
}
\caption{Detailed ablation results for in-domain (ID) and out-of-domain (OOD) experiments for NER models fine-tuned on social media, online forum, and Wikipedia texts. P, R, and F1 stand for precision, recall, and F1-score, respectively.}
\label{Table:AblationsForDataSources}
\end{table*}
\section{Domain Transfer: Impact of Text-source on NER Performance}
\label{Appendix:DomainTransfer}
Recognizing the need for robust NER models that can generalize well to other domains, we investigated model sensitivity to domain shifts such as data sampled from different text-sources. We conducted experiments under both in-domain (ID) and out-of-domain(OOD) settings. Our experiments show that combining data from different sources, such as Wikipedia, social media and online forums, is a promising strategy to make the NER systems less sensitive to domain shifts; due to improved diversity. NER models showed an improved performance. 

In this experiment, we fine-tuned three NER models on Wikipedia text, and saved the model. Then, we conducted an OOD evaluation in which each model was initialized with the saved weights after fine-tuning. Results indicate a drop in F1-scores across all domains, for all three models (see Figure \ref{fig:DomainTransferAllModelsinWikipedia1}). The drop in performance indicates the difference in characteristics between text found inside Wikipedia and texts in social media and online forums.
\begin{figure*}[!t]
\centering
\begin{subfigure}{0.33\textwidth}
\centering
\begin{tikzpicture}
\begin{axis}[
    ybar,
    ymin=40,
    ymax=105,
    width=\textwidth,
    height=3cm,
    bar width=0.60cm,
    enlarge x limits=0.4,
    legend style={at={(0.5,1.05)}, anchor=south, legend columns=-1},
    ylabel={F1-score},
    ylabel style={ font=\tiny },
    y tick label style={ font=\tiny },
    xtick=data,
    xticklabels={Wikipedia, social media, online forums, SM+OF},
    x tick label style={rotate=45, anchor=east, font=\tiny, cells={align=left} },
    nodes near coords,
    every node near coord/.append style={rotate=45, anchor=south, font=\tiny},
    legend style={at={(0.5,1.05)}, anchor=south, legend columns=-1, font=\tiny, cells={align=left} },
    ]
\addplot[fill=color4, postaction={pattern=north east lines}] coordinates { (0, 83) (1, 49) (2, 48) (3, 47)  };
\draw[red, thick] (axis cs:0,83) -- (axis cs:3,83);
\draw[red, thick, -stealth, node font=\tiny] (axis cs:1,83) -- node[right] {-34\%} (axis cs:1,49);
\draw[red, thick, -stealth, node font=\tiny] (axis cs:2,83) -- node[right] {-35\%} (axis cs:2,48);
\draw[red, thick, -stealth, node font=\tiny] (axis cs:3,83) -- node[right] {-36\%} (axis cs:3,47);
\legend{BERT-base F1-scores}
\end{axis}
\end{tikzpicture}
\caption{BERT-base}
\label{fig:DomainTransfer_BERTbase3}
\end{subfigure}%
\begin{subfigure}{0.33\textwidth}
\centering
\begin{tikzpicture}
\begin{axis}[
    ybar,
    ymin=40,
    ymax=105,
    width=\textwidth,
    height=3cm,
    bar width=0.60cm,
    enlarge x limits=0.4,
    legend style={at={(0.5,1.05)}, anchor=south, legend columns=-1},
    ylabel={F1-score},
    ylabel style={ font=\tiny },
    y tick label style={ font=\tiny },
    xtick=data,
    xticklabels={Wikipedia, social media, online forums, SM+OF},
    x tick label style={rotate=45, anchor=east, font=\tiny, cells={align=left} },
    nodes near coords,
    every node near coord/.append style={rotate=45, anchor=south, font=\tiny},
    legend style={at={(0.5,1.05)}, anchor=south, legend columns=-1, font=\tiny, cells={align=left} },
    ]
\addplot[fill=color2, postaction={pattern=dots}] coordinates { (0, 82) (1, 52) (2, 54) (3, 53)  };
\draw[red, thick] (axis cs:0,82) -- (axis cs:3,82);
\draw[red, thick, -stealth, node font=\tiny] (axis cs:1,82) -- node[right] {-30\%} (axis cs:1,52);
\draw[red, thick, -stealth, node font=\tiny] (axis cs:2,82) -- node[right] {-28\%} (axis cs:2,54);
\draw[red, thick, -stealth, node font=\tiny] (axis cs:3,82) -- node[right] {-29\%} (axis cs:3,53);
\legend{RoBERTa-base F1-scores}
\end{axis}
\end{tikzpicture}
\caption{RoBERTa-base}
\label{fig:DomainTransfer_RoBERTabase2}
\end{subfigure}%
\begin{subfigure}{0.33\textwidth}
\centering
\begin{tikzpicture}
\begin{axis}[
    ybar,
    ymin=30,
    ymax=105,
    width=\textwidth,
    height=3cm,
    bar width=0.60cm,
    enlarge x limits=0.4,
    legend style={at={(0.5,1.05)}, anchor=south, legend columns=-1},
    ylabel={F1-score},
    ylabel style={ font=\tiny },
    y tick label style={ font=\tiny },
    xtick=data,
    xticklabels={Wikipedia, social media, online forums, SM+OF},
    x tick label style={rotate=45, anchor=east, font=\tiny, cells={align=left} },
    nodes near coords,
    every node near coord/.append style={rotate=45, anchor=south, font=\tiny},
    legend style={at={(0.5,1.05)}, anchor=south, legend columns=-1, font=\tiny, cells={align=left} },
    ]
\addplot[fill=color3, postaction={pattern=crosshatch}] coordinates { (0, 83) (1, 33) (2, 43) (3, 39)  };
\draw[red, thick] (axis cs:0,83) -- (axis cs:3,83);
\draw[red, thick, -stealth, node font=\tiny] (axis cs:1,83) -- node[right] {-50\%} (axis cs:1,33);
\draw[red, thick, -stealth, node font=\tiny] (axis cs:2,83) -- node[right] {-40\%} (axis cs:2,43);
\draw[red, thick, -stealth, node font=\tiny] (axis cs:3,83) -- node[right] {-44\%} (axis cs:3,39);
\legend{DeBERTa-v3-base F1-scores}
\end{axis}
\end{tikzpicture}
\caption{ DeBERTa-v3-base }
\label{fig:DomainTransfer_DeBERTav3base1}
\end{subfigure}
\caption{\textbf{Domain transfer from Wikipedia to other domains:} F1-scores for different models (BERT-base, RoBERTa-base, DeBERTa-v3-base) when transferred from Wikipedia (in-domain) to other domains: social media (Reddit), online forums (Stack Exchange), and a mix of both. SM stands for social media, while OF stands for online forums. \textit{Performance drops are marked in red}.}
\label{fig:DomainTransferAllModelsinWikipedia1}
\end{figure*}
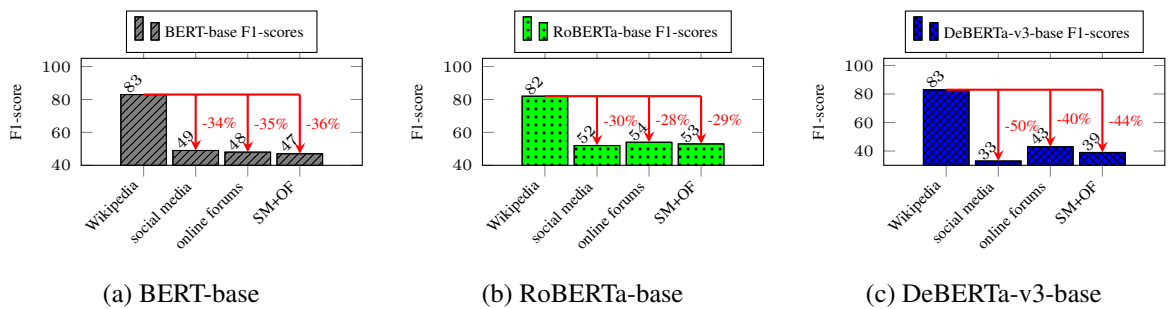

\begin{figure*}[!t]
\centering
\begin{subfigure}{0.33\textwidth}
\centering
\begin{tikzpicture}
\begin{axis}[
    ybar,
    ymin=95,
    ymax=101,
    width=\textwidth,
    height=3cm,
    bar width=0.60cm,
    enlarge x limits=0.4,
    legend style={at={(0.5,1.05)}, anchor=south, legend columns=-1},
    ylabel={F1-score},
    ylabel style={ font=\tiny },
    y tick label style={ font=\tiny },
    xtick=data,
    xticklabels={SM+OF+Wiki, social media, online forums, Wikipedia},
    x tick label style={rotate=45, anchor=east, font=\tiny, cells={align=left} },
    nodes near coords,
    every node near coord/.append style={rotate=45, anchor=south, font=\tiny},
    legend style={at={(0.5,1.05)}, anchor=south, legend columns=-1, font=\tiny, cells={align=left} },
    ]
\addplot[fill=color4, postaction={pattern=north east lines}] coordinates { (0, 96) (1, 98) (2, 98) (3, 98)  };
\draw[red, thick] (axis cs:0,96) -- (axis cs:3,96);
\draw[green, thick, -stealth, node font=\tiny] (axis cs:1,96) -- node[right] {+2\%} (axis cs:1,98);
\draw[green, thick, -stealth, node font=\tiny] (axis cs:2,96) -- node[right] {+2\%} (axis cs:2,98);
\draw[green, thick, -stealth, node font=\tiny] (axis cs:3,96) -- node[right] {+2\%} (axis cs:3,98);
\legend{BERT-base F1-scores}
\end{axis}
\end{tikzpicture}
\caption{BERT-base}
\label{fig:DomainTransfer_BERTbase1}
\end{subfigure}%
\begin{subfigure}{0.33\textwidth}
\centering
\begin{tikzpicture}
\begin{axis}[
    ybar,
    ymin=95,
    ymax=101,
    width=\textwidth,
    height=3cm,
    bar width=0.60cm,
    enlarge x limits=0.4,
    legend style={at={(0.5,1.05)}, anchor=south, legend columns=-1},
    ylabel={F1-score},
    ylabel style={ font=\tiny },
    y tick label style={ font=\tiny },
    xtick=data,
    xticklabels={SM+OF+Wiki, social media, online forums, Wikipedia},
    x tick label style={rotate=45, anchor=east, font=\tiny, cells={align=left} },
    nodes near coords,
    every node near coord/.append style={rotate=45, anchor=south, font=\tiny},
    legend style={at={(0.5,1.05)}, anchor=south, legend columns=-1, font=\tiny, cells={align=left} },
    ]
\addplot[fill=color2, postaction={pattern=dots}] coordinates { (0, 96) (1, 99) (2, 99) (3, 98)  };
\draw[red, thick] (axis cs:0,96) -- (axis cs:3,96);
\draw[blue, thick, -stealth, node font=\tiny] (axis cs:1,96) -- node[right] {+3\%} (axis cs:1,99);
\draw[blue, thick, -stealth, node font=\tiny] (axis cs:2,96) -- node[right] {+3\%} (axis cs:2,99);
\draw[blue, thick, -stealth, node font=\tiny] (axis cs:3,96) -- node[right] {+3\%} (axis cs:3,98);
\legend{RoBERTa-base F1-scores}
\end{axis}
\end{tikzpicture}
\caption{RoBERTa-base}
\label{fig:DomainTransfer_RoBERTabase3}
\end{subfigure}%
\begin{subfigure}{0.33\textwidth}
\centering
\begin{tikzpicture}
\begin{axis}[
    ybar,
    ymin=90,
    ymax=101,
    width=\textwidth,
    height=3cm,
    bar width=0.60cm,
    enlarge x limits=0.4,
    legend style={at={(0.5,1.05)}, anchor=south, legend columns=-1},
    ylabel={F1-score},
    ylabel style={ font=\tiny },
    y tick label style={ font=\tiny },
    xtick=data,
    xticklabels={SM+OF+Wiki, social media, online forums, Wikipedia},
    x tick label style={rotate=45, anchor=east, font=\tiny, cells={align=left} },
    nodes near coords,
    every node near coord/.append style={rotate=45, anchor=south, font=\tiny},
    legend style={at={(0.5,1.05)}, anchor=south, legend columns=-1, font=\tiny, cells={align=left} },
    ]
\addplot[fill=color3, postaction={pattern=crosshatch}] coordinates { (0, 96) (1, 98) (2, 99) (3, 93)  };
\draw[red, thick] (axis cs:0,96) -- (axis cs:3,96);
\draw[green, thick, -stealth, node font=\tiny] (axis cs:1,96) -- node[right] {+2\%} (axis cs:1,98);
\draw[green, thick, -stealth, node font=\tiny] (axis cs:2,96) -- node[right] {+3\%} (axis cs:2,99);
\draw[red, thick, -stealth, node font=\tiny] (axis cs:3,96) -- node[right] {-3\%} (axis cs:3,93);
\legend{DeBERTa-v3-base F1-scores}
\end{axis}
\end{tikzpicture}
\caption{ DeBERTa-v3-base }
\label{fig:DomainTransfer_DeBERTav3base2}
\end{subfigure}
\caption{\textbf{Combined domain-data boosts performance.} Improved F1-scores for different models (BERT-base, RoBERTa-base, DeBERTa-v3-base) when fine-tuned on all domain data. SM stands for social media, while OF stands for online forums. }
\label{fig:DomainTransferAllModelsinWikipedia2}
\end{figure*}
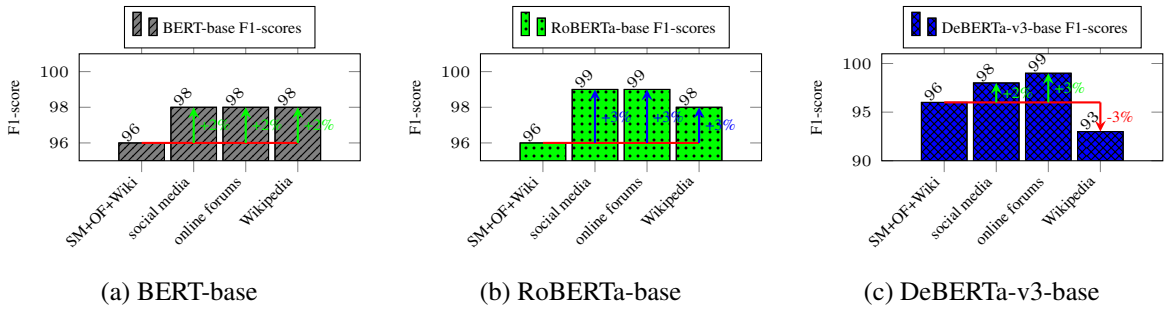
\begin{figure*}[!t]
\centering
\begin{subfigure}{0.33\textwidth}
\centering
\begin{tikzpicture}
\begin{axis}[
    ybar,
    ymin=20,
    ymax=105,
    width=\textwidth,
    height=4cm,
    bar width=0.60cm,
    enlarge x limits=0.4,
    legend style={at={(0.5,1.05)}, anchor=south, legend columns=-1},
    ylabel={F1-score},
    ylabel style={ font=\tiny },
    y tick label style={ font=\tiny },
    xtick=data,
    xticklabels={social media, Wikipedia, online forums, Wiki+OF},
    x tick label style={rotate=45, anchor=east, font=\tiny, cells={align=left} },
    nodes near coords,
    every node near coord/.append style={rotate=45, anchor=south, font=\tiny},
    legend style={at={(0.5,1.05)}, anchor=south, legend columns=-1, font=\tiny, cells={align=left} },
    ]
\addplot[fill=color3, postaction={pattern=crosshatch}] coordinates { (0, 96) (1, 29) (2, 86) (3, 79)  };
\draw[red, thick] (axis cs:0,96) -- (axis cs:3,96);
\draw[red, thick, -stealth, node font=\tiny] (axis cs:1,96) -- node[right] {-67\%} (axis cs:1,29);
\draw[red, thick, -stealth, node font=\tiny] (axis cs:2,96) -- node[right] {-10\%} (axis cs:2,86);
\draw[red, thick, -stealth, node font=\tiny] (axis cs:3,96) -- node[right] {-17\%} (axis cs:3,79);
\legend{DeBERTa-v3-base w/ social media}
\end{axis}
\end{tikzpicture}
\caption{\textbf{One.} social media data.}
\label{fig:DomainTransfer_BERTbase2}
\end{subfigure}%
\begin{subfigure}{0.33\textwidth}
\centering
\begin{tikzpicture}
\begin{axis}[
    ybar,
    ymin=20,
    ymax=105,
    width=\textwidth,
    height=4cm,
    bar width=0.60cm,
    enlarge x limits=0.4,
    legend style={at={(0.5,1.05)}, anchor=south, legend columns=-1},
    ylabel={F1-score},
    ylabel style={ font=\tiny },
    y tick label style={ font=\tiny },
    xtick=data,
    xticklabels={online forums, Wikipedia, social media, Wiki+SM},
    x tick label style={rotate=45, anchor=east, font=\tiny, cells={align=left} },
    nodes near coords,
    every node near coord/.append style={rotate=45, anchor=south, font=\tiny},
    legend style={at={(0.5,1.05)}, anchor=south, legend columns=-1, font=\tiny, cells={align=left} },
    ]
\addplot[fill=color3, postaction={pattern=crosshatch}] coordinates { (0, 98) (1, 28) (2, 87) (3, 48)  };
\draw[red, thick] (axis cs:0,98) -- (axis cs:3,98);
\draw[red, thick, -stealth, node font=\tiny] (axis cs:1,98) -- node[right] {-60\%} (axis cs:1,28);
\draw[red, thick, -stealth, node font=\tiny] (axis cs:2,98) -- node[right] {-11\%} (axis cs:2,87);
\draw[red, thick, -stealth, node font=\tiny] (axis cs:3,98) -- node[right] {-50\%} (axis cs:3,48);
\legend{DeBERTa-v3-base w/ Online Forum}
\end{axis}
\end{tikzpicture}
\caption{\textbf{Two.} online forums data.}
\label{fig:DomainTransfer_RoBERTabase1}
\end{subfigure}%
\begin{subfigure}{0.33\textwidth}
\centering
\begin{tikzpicture}
\begin{axis}[
    ybar,
    ymin=90,
    ymax=105,
    width=\textwidth,
    height=4cm,
    bar width=0.60cm,
    enlarge x limits=0.4,
    legend style={at={(0.5,1.05)}, anchor=south, legend columns=-1},
    ylabel={F1-score},
    ylabel style={ font=\tiny },
    y tick label style={ font=\tiny },
    xtick=data,
    xticklabels={Wiki+SF+OF, social media, online forums, Wikipedia},
    x tick label style={rotate=45, anchor=east, font=\tiny, cells={align=left} },
    nodes near coords,
    every node near coord/.append style={rotate=45, anchor=south, font=\tiny},
    legend style={at={(0.5,1.05)}, anchor=south, legend columns=-1, font=\tiny, cells={align=left} },
    ]
\addplot[fill=color3, postaction={pattern=crosshatch}] coordinates { (0, 96) (1, 99) (2, 99) (3, 92)  };
\draw[green, thick, -stealth, node font=\tiny] (axis cs:1,96) -- node[right] {+3\%} (axis cs:1,99);
\draw[green, thick, -stealth, node font=\tiny] (axis cs:2,96) -- node[right] {+3\%} (axis cs:2,99);
\draw[red, thick, -stealth, node font=\tiny] (axis cs:3,96) -- node[right] {-4\%} (axis cs:3,92);

\legend{DeBERTa-v3-base w/ ALL data}
\end{axis}
\end{tikzpicture}
\caption{ \textbf{Three.} Combined data.} 
\label{fig:DomainTransfer_DeBERTav3base3}
\end{subfigure}
\caption{ We ablate the source of data (social media, online forums, combined texts) used to fine-tune DeBERTa-v3-base for NER. Then, we conduct OOD experiments to measure the ability of the fine-tuned model to generalize to unseen domains. \textit{DeBERTa-v3-base performance in OOD setting improves when finetuned with a combination of all data sources; see (c) on the right. } SM stands for social media, while OF stands for online forums. }
\label{fig:DomainTransferInDeBERTa}
\end{figure*}
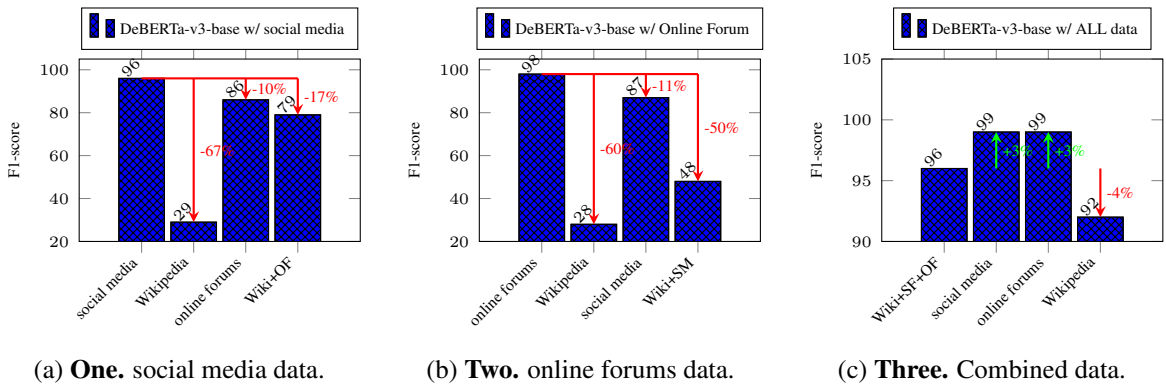
\begin{figure*}[!t]
\centering
\includegraphics[width=13cm]{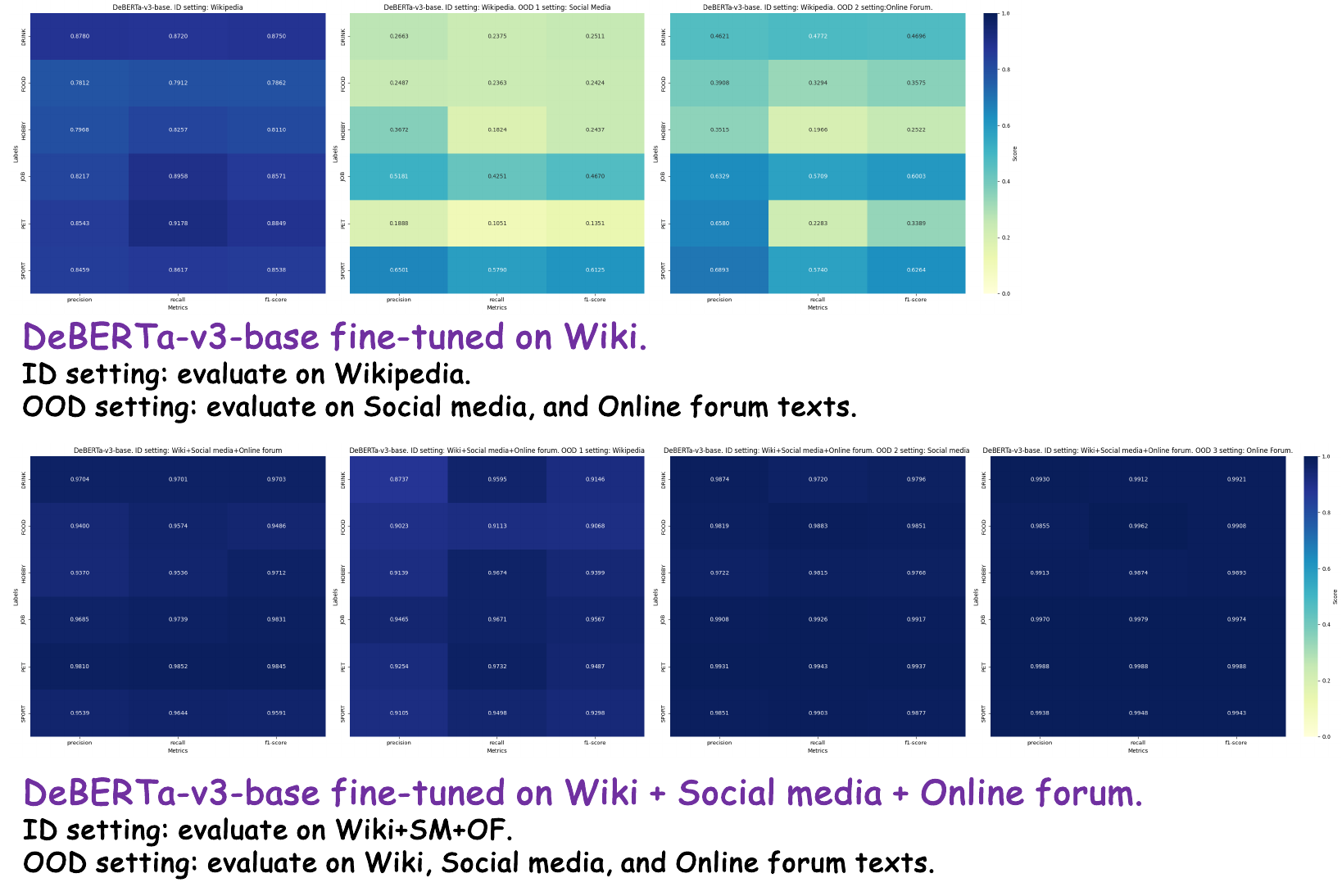}
\caption{Heatmaps showing the Precision, Recall, and F1-score across named entities; for both ID and OOD settings. The NER model is DeBERTa-v3-base.}
\label{fig:DeBERTabasev3Viz}
\end{figure*}
\begin{figure*}[!t]
\centering
\includegraphics[width=13cm]{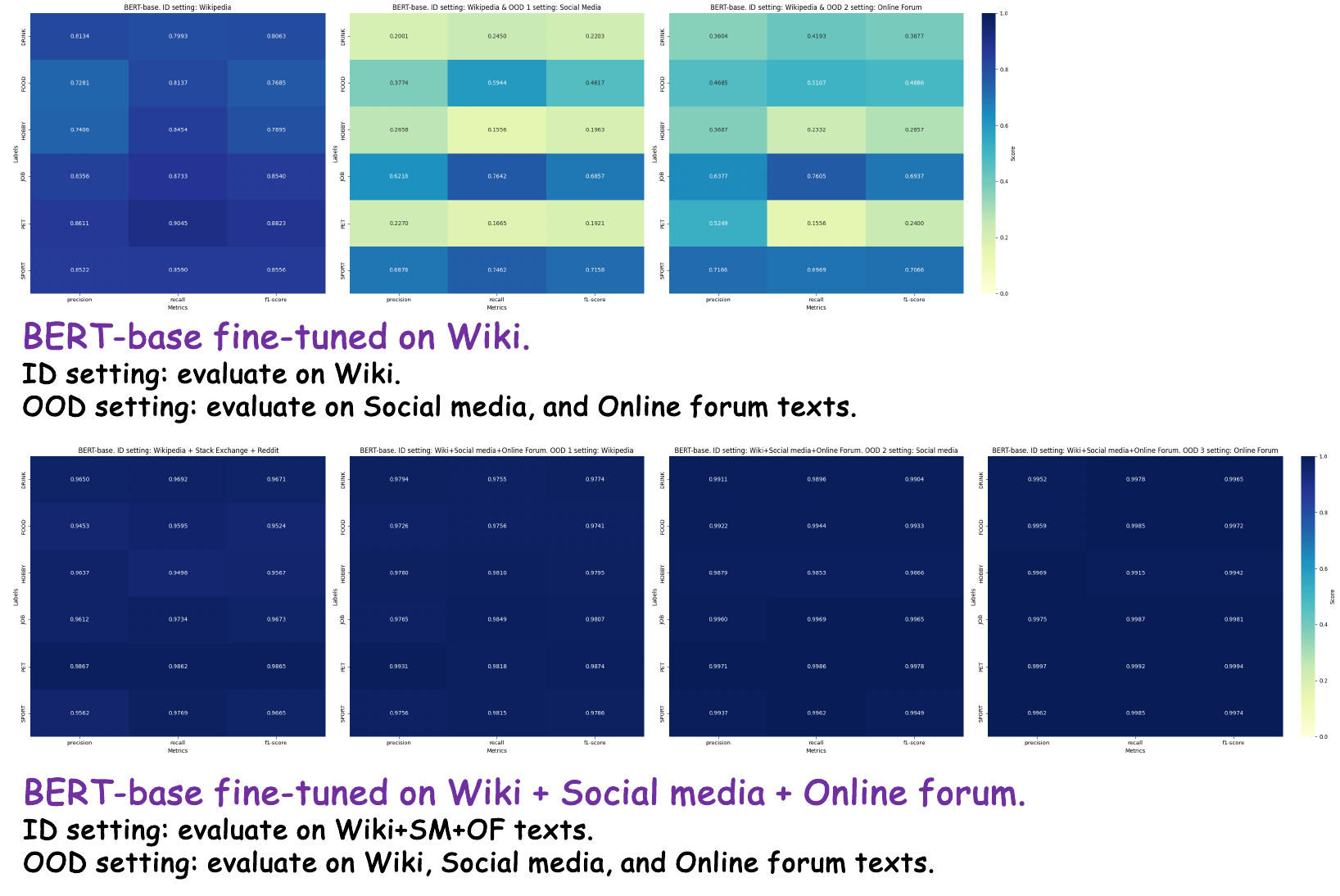}
\caption{Heatmaps showing the Precision, Recall, and F1-score across named entities; for both ID and OOD settings. The NER model is BERTa-base.}
\label{fig:BERTbaseViz}
\end{figure*}
\begin{figure*}[!t]
\centering
\includegraphics[width=13cm]{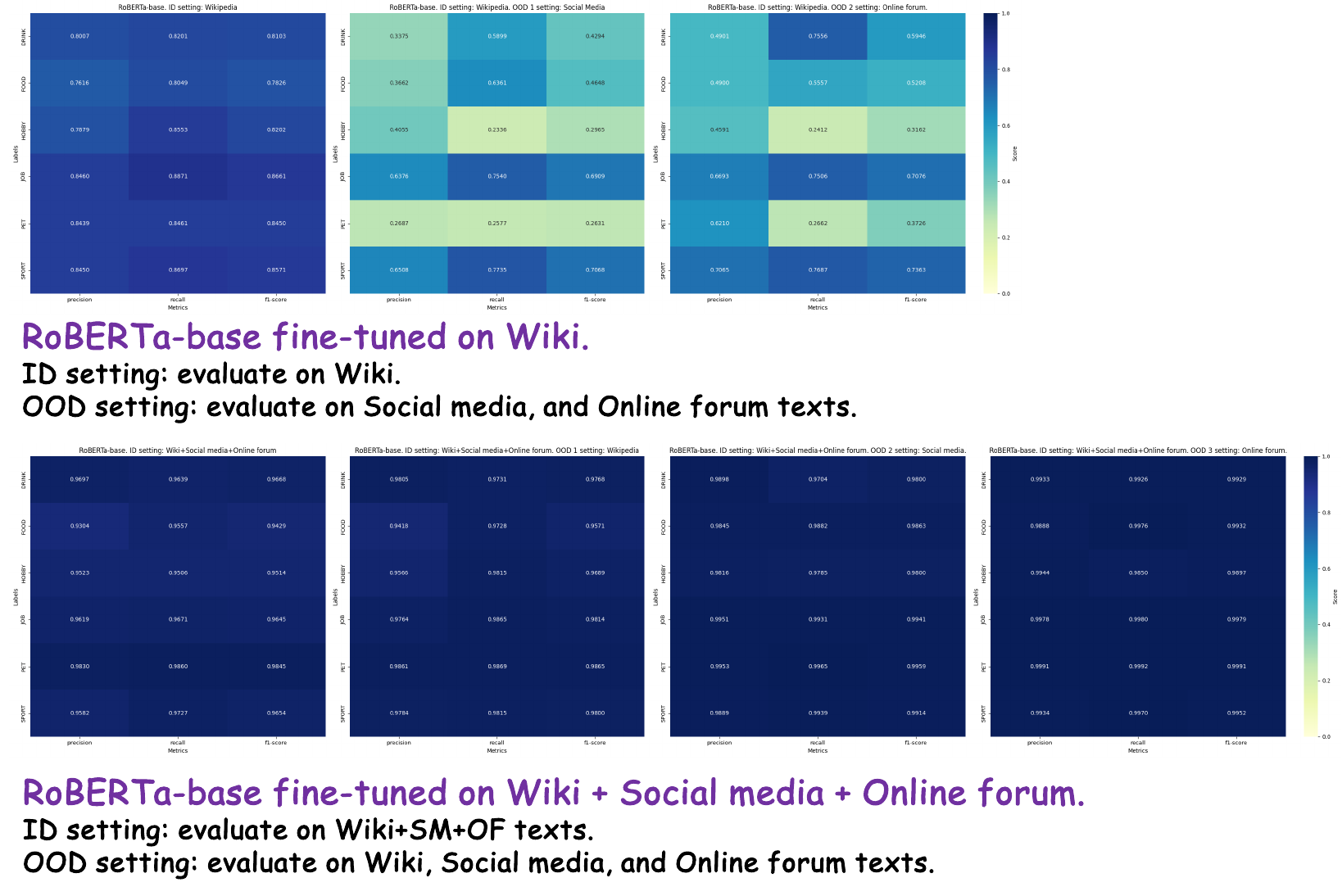}
\caption{Heatmaps showing the Precision, Recall, and F1-score across named entities; for both ID and OOD settings. The NER model is RoBERTa-base.}
\label{fig:RoBERTbaseViz}
\end{figure*}

\begin{table*}[!t]
\centering
\scriptsize
\begin{tabular}{p{0.95\textwidth}}  %
\toprule
\textbf{Examples of \textbf{Drink} hyponyms from Wikidata.} \\
\midrule
Fruitopia, Actimel, sima, Posca, Rivella, Sugarcane juice, Aquafina, Dortmunder Hansa, Brause, Tereré, Cole Cold, Qoo, Chun Mee tea, Arabic coffee, Tequila Sunrise, Cajuína, mors, ochsenblut, White Horse, Paddy Whiskey, gorbatschow, Black drink, Apotekarnes Cola, Perú Cola, A\&W Cream Soda, Blantons, Clamato, coco, sahti, Everclear, Dewars, Killians, Powers, Pepsi Twist, Molson Canadian, Monopolowa, Mug Root Beer, SunnyD, Laranjada, Yop, Frooti, Coca-Cola with Lime, Mocochinchi, Olvi Cola, Xan, Syta, Uzvar, 10 Cane, Pacífico, Absolwent, AdeS, Ancient Age, Anker Beer, Asahi Point Zero, Aspen Soda, Atholl brose, Bagpiper Whisky, Barton Premium Blend, Basil Haydens, Beams Eight Star, Bevo, Bibo, Bing, Black \& White, Bookers, Brisa drink, Buchanans, Bulleit Bourbon, Caffeine-Free Pepsi, Calvert Extra, Canadian Mist, Carlton Draught, Cassinelli, Champús, Ck vodka, Coca-Cola Raspberry, Colorado Native Lager, Concordia, Cooranbong, Cork Dry Gin, Country Time, Cracovia vodka, Crystal Light, Diet Coke Lime, Duke and Sons, Dunvilles Three Crowns, Dunvilles VR, Dutch Gold, Vio, Leonese lemonade, Export Cola, Ezra Brooks, Farmers Union Iced Coffee, Fentimans Curiosity Cola, Floats, Frisco, Fruit2O, George T. Stagg, Grand Valley Brewing Company, Grapico, Green Spot, Guinness Foreign Extra Stout, ayran of Susurluk, Yayık ayranı, Kentucky Tavern, Kia-Ora, Kirin Free, Kissui, Kola Inglesa, Leed, Lucid Absinthe, Maaza, Manzanita Sol, Melbourne Bitter, MiO, Mondo, Nativa, Nesher Malt, Noahs Mill, Nonino Grappa, OK Soda, OMORÉ, Old Charter, Old Crow, Old Ezra 101, Old Forester, Old Grand-Dad, Old Style Pilsner, Olde English 800, Olmeca Tequila, Oro, Papelón con limón, Pappy Van Winkles Family Reserve, Pendleton Whisky, Pepsi Wild Cherry, Pibb Xtra, Pilsen Trujillo, Pinnacle vodka, Pulp, Redbreast, Reyka, Rip It, Rowans Creek, Royal Tru, Seagrams Seven Crown, Shani, Skinny Blonde, Snapple, Sporade, Sprite Remix, TEHO, Tava, Teachers Highland Cream, Ten High, The Northern Lights, Tilt, Tooheys New, Topsia Cola, Town Branch, Tusker, Vernors, Vintage Bourbon, Virginia Gentleman, Viva, W. L. Weller, White Mischief, XXXX Gold, Yellowstone Bourbon, Yoo-hoo, Zima, DanUp, Midleton Very Rare, Mupy, Mer, KIRIN LEMON, Coca-Cola Plus, Lemon Milk, Sjus, Taar, Ciu, Bombora, Bonaqua, A\&W Root Beer, After Dark, Pópo, chamoyada, Caprio, Amita, Full Throttle, Carlton Cold, XXXX Bitter, Kessler Whiskey, Cola Couronne, Black Dog Scotch Whisky, Boddingtons Bitter, Antiquity, Blenders Pride, J2O, Directors Special, Imperial Blue, Jeremiah Weed, Knappogue Castle Irish Whiskey, McAfees Benchmark, Bradys Irish Cream, Celtic Honey, Mechitharine, Old Rip Van Winkle, Peter Scot, Red Knight, Rowsons Reserve, Royal Challenge, Royal Stag, Iron Brew, Worthingtons White Shield, Pure Blonde, Willett Pot Still Reserve, Maestro Dobel Tequila, Josta, Fullers ESB, London Pride, Organic Honey Dew, Fiji Bitter, Vailima, Carlton Midstrength, Passiona, Vodka O, Chibuku Shake Shake, Black Bull, Font Vella Levite, Lanjarón, Oxycrat, Agua de Solares, Unicorn Frappuccino, Pura Vida, RumChata, Club soda, Green Cola, bottled spring water, Grapefruit mate, Green iced tea, Brand 5, Lanique, Somó, Bignay wine, Tubâ, Blue milk, Green milk, Brukina, alcoholic beverage, drinking water, coffee, juice, milk, citronnade, citron pressé, Limonana, smoothie, slush, Affogato, Caffè corretto, Soma, apéritif, ayran, chicha morada, Hierbas, amazake, eggnog, Buttermilk koldskål, Leche Merengada, boza, salep, tomato juice, Belaya Rus vodka, Lassi, Colada morada, latte, kissel, Biała Dama, kykeon, nectar, Sotol, Hibiscus tea, socată, chal, ginger tea, Mecca-Cola, Caffè Americano, Tejate, Hoppelpoppel, Krupnik, milk substitute, kompot, Vin Mariani, Caudle, Cola de mono, Noche Buena, dairy drink, East Friesland tea blend, Sbiten, Tjolk, plant milk, Nalewka, Moretta, Horchata, chicha de jora, mixed drink, Postum, Oasis, Pastis Henri Bardouin, Suanmeitang, Keglevich, Neera, Şıra, Mocochinchi, Shedeh, Oxygen cocktail, Ruskova, Sairme, Sayani, Layered drink, Uzvar, Coke float, 5-hour Energy, Accelerade, Alazani, Ale-8-One, All Sport, Arctic Velvet, Arwa, Barbasol, Beep, Benjamin Prichards Tennessee Whiskey, Boilo, Boj, Bondi Blonde, Bong Spirit Vodka, Boodles British Gin, Bovonto, Brisbane Bitter, caffeinated drink, Café Rica, Cape North Vodka, Casa Dragones, Casa Noble, Cel-Ray, Chernihivske, Chicago Root Beer, Choc-Ola, Chuflay, Cirrus Vodka, Citadelle, Clique, Clix Malt Liquor, Coca wine, Cochlear Baha, Coco López, Colt 45, Cool mountain beverages, Crusha, Double Cross Vodka, Dr. Wells, Filberts Old Time Root Beer, flavored milk, Frïs Vodka, General Foods International, Glens Vodka, Golden Grain Alcohol, Gorki List, Governator Ale, Han Vodka, Harcos Laboratories, Hardaliye, Inka, Jobo, soft cider, Submarino, tizana, Jim Beam Rye, KEO, Kavaklıdere, Khardwi, Kola Escocesa, Kunu, Kübler Absinthe, La Croix Sparkling Water, Le Tourment Vert, Leon Beer, sugar water, Lion Red, syrup, Kurdish coffee, Lotus vodka, Malt beverage, Mega Sport, Mercy, Mickeys, No-Cal Soda, Nordic Mist, Nuvo, Old Taylor, Original New York Seltzer, peanut punch, Peardrax, Penta Water, Persa, Pimp Juice, Platinka, Sharbat, Polmos Łańcut, Polonaise, prairie oyster, Propel Fitness Water, Purl, RESQ, Rasna, Reál Sangria, Air sirap, relaxation drink, Rhythm, Rock Hill Farms Single-Barrel, Royal Salute whisky, Sams Choice, Sangria Señorial, Savvy Vodka, Schaefer Beer, Señor Río, Shustov vodka, Snow Queen Vodka, Sobieski Vodka, Speights Old Dark, Spike Shooter, St. Ides, Staminade, Tatratea, Templeton Rye, Tiky, Tooheys Extra Dry, Tooheys Old, Tres Agaves, Tres Generaciones, Triibe, Ultimat Vodka, V44, Very Old Barton, Vitasoy, Vodka 14, Vodka Perfect, White Mountain Cooler, Wisent, XUXU, Blue Moon, Frugo, Soplica, Cocaine, Hype Energy, non-alcoholic beverage, ED, love potion, Boyd \& Blair, carbonated beverage, beef tea, Navan liqueur, Wostok, Blue, 1519 Tequila, 4 Copas, Milo Dinosaur, Lucky Lager, Es kelapa muda, Avion tequila, Joose, Lohusa Şerbeti, Frozen alcoholic beverage, Laser (malt liquor), Crabbies, Eghajira, VÄD, Domaine Pinnacle Ice Cider, Castello Mio Sambuca, Anestasia Vodka, Jeffersons Bourbon, Nae Danger, Leblon Cachaça, Fortuna, DeLeón Tequila, Espolon, Gozio Amaretto, Wrexham Lager Beer, Mageu, Gulden Draak 9000 Quadruple, Mana Energy Potions, Meyenberg Goat Milk, National Bohemian, VKA Vodka, SoBe Mr. Green, Pallini Limoncello, Finley, Tokhm sharbati, Donat Mg, 3 A.M. Vodka, Xingu (beer), hot beverage, Spruce beer, Yagua, Farol, Suau, pinol, Jacqueline, fermented beverage, Swiss lemonade, Santera Tequila, Halls Beer Cheese, Black Cow Vodka, liquid water, oral rehydration solution, gamnip-cha, dohwa-cha, doncha, Supligen, Mead in Poland, Brexitovka, Mio Mio Mate, Schorle, El Namroud (Arak), coffee drink, Sucumbé \\
\bottomrule
\end{tabular}
\caption{Instances from the hyponyms of \textbf{Drink} used to construct \textit{NERsocial}. }
\label{Table:listOfHyponymsDrink} 
\end{table*}
\begin{table*}[!t]
\centering
\tiny
\begin{tabular}{p{0.95\textwidth}}  %
\toprule
\textbf{Examples of \textbf{Food} hyponyms from Wikidata.} \\
\midrule
bread, chorizo, emping, olive oil, Tempeh, Urtica dioica, ayran, satay, Gado-gado, kanafeh, tarte flambée, nachos, mochi, Vigna radiata, Cumberland sausage, refried beans, Brem, purée, Fischbrötchen, Youtiao, pork ball, Rojak, scallion, fricassee, BP-5 Compact Food, banana chip, Beni shōga, tzatziki, canning, Bakso, Coto Makassar, cendol, shahe fen, Soto ayam, Muktuk, petit four, beef bourguignon, Caldo gallego, Gözleme, Carnaroli, Kısır, Sōki, shabu-shabu, Corned beef, jajangmyeon, Cuajada, profiterole, Kilishi, Ghormeh sabzi, fruit preparation, Rendang, Skolebrød, Heumilch, Lontong, Lemang, kabsa, Syrniki, Soused herring, Cioccolato di Modica, Novel food, pissaladière, Nasi krawu, Sauce ravigote, Pastirma, tofu skin, Filipinos, Liege waffle, Lacquemant waffles, Fusarium venenatum, Lemper, sate kambing, Idiyappam, frog cake, Yufka, condensed milk, Salted duck egg, fried green tomatoes, Tklapi, Otak-otak, Nasi uduk, Nasi kucing, black-eyed pea, Nasi kuning, nasi campur, Jin deui, Neera, Ternasco de Aragón, sadya, Katmer, Finocchiona, es doger, Kidneys (meat), lumpia, mixed nuts, Gulai, Nagasari, Oncom, lawar, Pempek, Rawon, Jamu, Sega lengko, Bajigur, Lupis, Karedok, nasi kebuli, Urap, Bakwan, Mie celor, Geplak, Papeda, Amala, Tempoyak, Pide (flatbread), Bandrek, batagor, Betutu, Nasi ulam, red tortoise cake, Buntil, Cicchetti, Serabi, Esquites, Rengginang, Tapai, Pecel, Klepon, Getuk, Docang, Empal gentong, Es teler, Gaplek, Golden Nuggets, Enfrijoladas, Humanitarian daily ration, Iced VoVo, Imokilly Regato, Kuzu fırın, Dilber dudağı, Peynir helva, pumpkin dessert, Gavurdağı salatası, kuzu kapama, hoşmerim, Karadeniz pidesi, Kemplang, Kerak telor, keşkül, Konro, kaak al-Quds, paçanga, Mezcal worm, Nasi bogana, Nasi liwet, Pallubasa, Penne alla vodka, pig roast, Pizza Rolls, Kuih sepit, Roti buaya, Kue leker, Nisan, Sate Padang, Scrod, Soto Madura, Sega Jamblang, Tahu sumedang, Tahu gejrot, Swikee, Tekwan, Ting Ting Jahe, Wingko, kalburabastı, stuffed zucchini flower, börek, alphabet pasta, Sate Madura, Cenil, linguiça, Soto mie, Celimpungan, Dawet Ice, Es Gempol, Es bir, Kelan Antep, krupuk kulit, Kuluban, Mendoan, Rica-rica, Soto Betawi, Soto Sokaraja, Teh poci, Cicvara, Zerde, Chigirtma, Jizza, Luk chup, Pilus, satay maranggi, Nasi Kapau, Paripuvada, Dodol Garut, Géco, Ampo snack, smoked egg, cornetto, Corn tortilla, Shanxi knife-cut noodles, Mie ayam, Bubur ketan hitam, Bubur pedas, Kembang Goyang, Bandros (food), kuru fasulye, Es goyobod, Es buah, Kuih jelurut, Empal gepuk, Sate Bandeng, Costrada de Alcalá, kuih cucur, Kaz tiridi, Palm nut soup, supangle, Sate Lilit, Kue ape, Seblak, Pindang, Adrem (snacks), Vakfıkebir ekmeği, Pukis, zeytinyağlı enginar, Kuzu haşlama, Cennet çamuru, Sop saudara, Köftes in the Turkish cuisine, Kpekpele, Kue gapit, Chiroti, Tauge goreng, fu, breakfast burrito, Kue bugis, Kue satu, Pancong, Majadito, Pecel pitik, Keema Matar, rolex, chip, Fried Rice with Salted Fish and Chicken, Braised Chicken, Red peach cake, cacık, Çökertme, Bagiak, Cilok Goreng, Cilung, Es Goyang, Brown Sugar, Jaja Giling, Kritcu, Pepe, Red Rice, Marrow and Rice, Petis Bumbon, Pork Satay, Duck Satay, Sate Sapi Suruh, Soto Bogor, Kame Jush, Fugazzetta, Fugazza with cheesse, piccantino, Pizza al tegamino, Rusip, Food Lion, Kroger, Aldi, Commissary, Walmart, Tinee Giant \#147, Virginia Beach Farmers Market, Foodbank of Southeastern Virginia an the Eastern Shore, Smoked Roa Fish Sauce, Naturegg, Camin, Niva cheese, Dill, Laverbread, Crab melt, Llaucha, Tlačenka, Sándwich de chola, Quispiña, Mukuna, Pasankalla, longaniza de Aragón, Jam, wild rice, gari and beans, Chick-fil-A Waffle Potato Fries, Chick-n-Strips, Gari foto, Spleen as food, arayes, Mpoto Mpoto, Dilly Bar, revani, Mücendra pilavı, memek, Slavic ritual food and drinks, yassı kadayıf, oseng-oseng mercon, Turkish mantı, Pearl, Omelette Arnold Bennett, Mohanthal, Toros salad, Kabak çintmesi, pasta, margarine, borscht, escalope, speculaas, hamburger, milk, salad, Coussin de Lyon, meat, sugar, vegetable, andouillette, ladyfinger, kiwifruit, biscuit, shashlik, gnocchi, omelette, root vegetable, rotisserie, Zwiebelkuchen, batter, Shchi, Tarta de Santiago, drink, seed, soup, Funyuns, pineapple cake, strudel, garnish, paratha, Chenpi, fast food, egg, Christmas wafer, pemmican, Nam ngiao, oladyi, sausage, jeon, pan loaf, kama, dondurma, Zamorano cheese, legume, strawberry pie, malt, edible underwear, Halver Hahn, Stutenkerl, rijstevlaai, offal, Gugelhupf, millet, Easter egg, Monjayaki, Tortell, potato chip, king cake, dough, egg yolk, Turkish delight, dessert, Baklava, falafel, fondue, sashimi, seafood, Royal jelly, gazpacho, herb, guacamole, haggis, steak, dietary fiber, Dragonfruit, manna, hummus, galette, ciabatta, zōni, Poğaça, cocoa butter, crisp bread, Marillenknödel, cinnamon roll, powidl, Flädle, Spiesebraten, apple butter, curry puff, damper, apple strudel, Punschkrapfen, milk-cream strudel, Žemlovka, Kaiserschmarrn, Egg in the basket, chop suey, biryani, bagel, Çäkçäk, fried egg, broth, Mousefood, Khachapuri, Lo mein, Vetkoek, Poffertjes, fairy bread, spirulina, Spanisch Brötli, balut, vol-au-vent, Ragout fin, Pease pudding, Haddekuche, Ričet, tempura, joshpara, Bienenstich, Dolma, imambayıldı, Lecsó, bean, organic food, tahini, Capicola, Burgossan cheese, burrito, Acarajé, genetically modified food, Ajoblanco, ajvar, gastrique, Roncal cheese, Johnnycake, Fritule, omurice, Spritzkuchen, English muffin, oyster cracker, Leipziger Lerche, Majorero, pastry, Absnicli, Gyeongju bread, pilaf, blood, dried shrimp, hotteok, Amish Friendship Bread, au jus, sushki, Pozole, obara, Magiritsa, Bacalhau à Brás, Burrata, deviled egg, wonton, teacake, Bánh bao, Tacacá, cabbage soup, anpan, Antidoron, Cocido lebaniego, Rönttönen, Gim, Lokma, cretons, Hawaiian pizza, Pecorino Sardo, Mouna, Bigos, Thukpa, cardamom bread, famine food, okroshka, bruschetta, Ovelgönne bread roll, Fatoot, century egg, curd, mousse, Buldak, ice cream sandwich, coconut water, arepa, pain aux raisins, Kommissbrot, Solomon Gundy, shortbread, oden, pigs in blankets, scrambled eggs, kuku, bosintang, Bossche bol, gâteau Basque, Tteokguk, Birnenhonig, Papanași, Scaccia, fritter, Kruidnoten, Lardo, tea egg, sweetheart cake, Germknödel, Kimchi-jjigae, budae-jjigae, Obatzda, turnip cake, oyster omelette, Shin Ramyun, Yong tau foo, Miyeok guk, Seolleongtang, Suncake, brain, quiche, Maamoul, Sinseollo, staple food, Heart of palm, dish, tamagoyaki, finger food, snack, space food, frozen food, chapati, functional food, pollock roe, Sopaipilla, Graham bread, breakfast cereal, Pistou, Michetta, Biscotti, challah, egg white, baby food, Backmalz, pâté, aspic, Pork chop bun, Pane sciocco, soufflé, beef Stroganoff, Bresse chicken, junk food, Manduguk, panettone, kissel, Kompyang, steak and kidney pie, Geng, chicken, memil-buchimgae, Birnbrot, bisque, Cocido, Hirayachi, Kiritanpo, Scone, rice bread, bear claw, shell-shaped noodle, tapioca, Knieküchle, kifli, Hochzeitssuppe, Bitto, Nuns puffs, Borodinsky bread, Bosnian pot, Bougatsa, Uttapam, cracknel, Dongchimi, rose hip soup, guk, sundubu-jjigae, butterfat, edible fats and oils, Bêtise de Cambrai, Steckrübeneintopf, bhakri, Tiger bread, Sesame seed candy, Piadina, Cocido Montañés, Lekvar, Bakauke, laksa, Nasi pecel, Chakhchoukha, Bouillon cube, brown bread, comfit, Thenthuk, fried cauliflower, Mekitsa, food product, Podpłomyk, Pannenkoeken, Grit, Kulcha, Krotekake, Double ka meetha, grooms cake, pastila, mochi ice cream, queso flameado, Piggy bun, Bak kut teh, Wodzionka, Brunswick stew, tri-tip, Cabell dàngel, Cachapa, Caldo gallego, Carac, Galbitang, butter cookie, sugar cube, hot dog bun, Chakhokhbili, Orama, botvinia, pumpkin bread, Laobing, trofie, gooey butter cake, cocoa powder, Jjigae, boiled egg, tomato soup, Yukgaejang, Chermoula, Menu, Jeongol, Chicago-style pizza, Muk, chili powder, Chilled food, gomguk, kosher foods, Welsh cake, kkul-tarae, Tripe soup, Khuushuur, Ciorbă de perișoare, clam chowder, Tyurya, ukha, coca, Cocido madrileño, Coda alla vaccinara, cornbread, Kottu, mush, Allerheiligenstriezel, Succotash, Buddha Jumps Over the Wall, Corned beef, farro, Tonjiru, Istrian stew, Okinawa soba, cream pie, Country Captain, baozi, coconut cream, Semla, Dal, Danish pastry, red bean paste, pineapple bun, jajangmyeon, Crumpet, Crêpe Suzette 
\\
\bottomrule
\end{tabular}
\caption{Instances from the hyponyms of \textbf{Food} used to construct \textit{NERsocial}. }
\label{Table:listOfHyponymsFood} 
\end{table*}
\begin{table*}[!t]
\centering
\scriptsize
\begin{tabular}{p{0.95\textwidth}}  %
\toprule
\textbf{Examples of \textbf{Hobby} hyponyms from Wikidata.} \\
\midrule
Stamp collecting, Blog, blogging, Bonsai, Penjing, board games, book discussion club, book club, book restoration, bowling, Brazilian jiu jitsu, BJJ, jiu jitsu, bullet journal, calligraphy, candle making, candy making, car spotting, card game, cardistry, cheesemaking, chess, Home roasting, coffee roasting, coloring book,  computer programming, confectionery, cooking, cosplay, couponing, creative writing, crochet, crossword, cue sports, dance, decorative arts, digital art, distro hopping, diving, DJing, Do it yourself, DIY, drawing, editing, electronics, embroidery, engraving, fantasy sport, fashion, fashion design, feng shui, filmmaking, fingerpaint, fish farming, fishkeeping, floral design, flower arrangement, second language acquisition, SLA, gambling, betting, gaming, homebrewing, home distillation, hula hoop, jewellery design, jigsaw puzzle, juggling, karaoke, kendama, knitting, kombucha brewing, lapidary, leather crafting, lego, livestreaming, streaming, music, podcast, lock picking, machining, macramé, magic, cosmetics, manga, massage, maze, mechanics, meditation, memory improvement, metalworking, minimalism, simple living, model building, model engineering, nail art, manicures, origami, pen spinning, performance, pet sitting, philately, photography, pilates, planning, plastic arts, poetry, poi, pottery, public speaking, puppetry, pyrography, quilling, quilting, quiz, railway modelling, model railroading, rapping, reading, recreational drug use, refinishing, reiki, robot combat, role playing, Rubik's Cube, scrapbooking, scuba diving, sewing, shoemaking, singing, sketching, skipping rope, rope skipping, slot car racing, shopping, social media, speedrunning, stand up comedy, storytelling, stretching, string figure, Sudoku, tabletops, Taekwondo, tanning, tarot, tattoo, taxidermy, upcycling, video editing, Watching documentaries, Watching movies, Watching television, weaving, webtoon, weight training, wikiracing, wine tasting, winemaking, wood carving, woodworking, word search, writing, musical composition, yoga, zumba, Amateur geology, rock collecting, rockhounding, auto detailing, astronomy, backpacking, tenting, hiking, tramping, camping, hill walking, trekking, beachcombing, beekeeping, birdwatching, camping, canoeing, canyoning, canyoneering, car tuning, dog walking, fossicking, amusement park visiting, fruit picking, geocaching, ghost hunting, gold prospecting, graffiti, groundhopping, hiking, walking, hillwalking, tramping, letterboxing,  treasure hunting, magnet fishing, noodling, Renaissance fair, renovation, road cycling, roller skating, safari, Scouting, snorkeling, tanning, sun bathing, Tai chi,  thru hiking, travel, urban exploration, vacation, vegetable farming, videography, fishkeeping, amateur radio, Tyrosemiophilia, Antenne Bergstraße, Sucrology, birdwatching, DXing, model building, garage kit, filmmaker, philatelist, Squirrel fishing, book collecting, video blogger, Butterfly watching, Element collecting, MW DX, mermaiding, mineral collecting, Sneaker collecting, modelling with clay, sardana dancer, Cynophilia, art collecting, model builder, autistic special interest, doujin artist, hunting, gardening, caving, collecting, aircraft spotting, postcrossing, storm chasing, bus spotting, Hobby farm, philia, transporting \\ 
\bottomrule
\end{tabular}
\caption{Instances from the hyponyms of \textbf{Hobby} used to construct \textit{NERsocial}. }
\label{Table:listOfHyponymsHobby} 
\end{table*}
\begin{table*}[!t]
\centering
\tiny
\begin{tabular}{p{0.95\textwidth}}  %
\toprule
\textbf{Examples of \textbf{Job} hyponyms from Wikidata.} \\
\midrule
scientist, clown, visual effects supervisor, special effects supervisor, ninja, astronomer, medicine, astronaut, king, probation officer, spokesperson, naturalist, tenor, anesthesiologist, mayor, soprano, public administration, photographer, actor, writer, teacher, mime, Hello Girls, coach, webmaster, reporter, architect, businessperson, tsar, sailor, spiritual direction, military personnel, poet, podologist, domestic worker, hairdresser, intellectual, Kannushi, chancellor, paraveterinary worker, secretary, engineer, shepherd, computer scientist, geisha, politician, flight attendant, knight, barber, firefighter, race queen, Patrouillenreiter, beutler, fashion tailor, professor, First officer, audio engineer, management consulting, disc jockey, urban planner, farmer, entrepreneur, lay brother, entertainer, altar server, technology evangelist, dental administrative assistant, dental technician, sports agent, crossing keeper, cobbler, astrologer, paparazzi, cook, major general, waiter, watchmaker, non-commissioned officer, broker, baker, deacon, war correspondent, electrician, ombudsman, physicist, mathematician, art dealer, printer, speechwriter, burgomaster, mercenary, cowboy, barista, actuary, librarian, software developer, philanthropy, mufti, Zeitmilitär, temporary career soldier, newspaper delivery, antiquarian seller, economist, military officer, public health, sommelier, master of ceremonies, mangaka, diplomat, elevator operator, machinist, kit manager, privateer, historian, sniper, referee, greenskeeper, veterinarian, Combat Medic Specialist, Zimmerpolier, inventor, state-certified engineers specializing in building construction, Civilian employee, executioner, psychiatrist, goldsmith, civil servant, psychologist, freelancer, merchant, chauffeur, chauffeuse, qadi, customs broker, Zollrevisor, croupier, missionary, zootechnician, music publisher, prosector, flying ace, assistant referee, Japanese idol, Zuchtwart, Breacher, conductor, patternmaker, bicycle mechanic, shipbroker, muezzin, tailor, Zwischenmeister, comedian, customs officer, strongwoman, plumber, navigator, alterations specialist, medical director, medical psychotherapist, repentista, dievdirbys, lifeguard, chamberlain, medicine man, independent scholar, news presenter, condottiero, milkman, literary agent, mosaicist, insurance broker, house painter, general contractor, atomic spy, land surveyor, ice master, chocolatier, prison officer, funeral director, notary's assistant, ablader, zookeeper, polyglot, factor, naval surgeon, accountant, mechanic, mason, system administrator, stonemason, engraver, justice of the peace, paramedic, cantor, fisher, translator, Division manager, jeweller, emergency physician, account manager, zoologist, heilpraktiker, administrator intercalaris, type designer, nurse anesthetist, showman, church musician, copy editor, joint venture broker, police officer, rhetorician, doctoral advisor, aide-de-camp, ferryman, statesperson, mental calculator, boom operator, Apparatchik, racing driver, flight instructor, smokejumper, woman of letters, official, police officer, drummer, treasurer, Thai boxer, master of novices, airman, academician, acolyte, preacher, flight engineer, trainee care provider, weaver, conservator, guitar maker, horse trainer, garde champêtre, wet nurse, bellfounder, fusilier marin, maritime pilot, sailor, government veterinarian, artist, chief executive officer, analyst, pornographic actor, singer-songwriter, credit broker, dental assistant, professional golfer, Fakir, Master of ceremonies, lepidopterist, preceptor, musher, shipowner, chimney sweep, Sofer, truck driver, griot, inspector, Coureur des bois, Benshi, heraldist, real estate broker, geologist, boilermaker, illegal prospectors, gravedigger, corrector, dominatrix, pharmacy technician, quality specialist, animator, brother, sailmaker, bullfighter, fishmonger, physician writer, roofer, plant and apparatus builder, anlagenmechaniker, investor, carrier, dispatcher, bureaucrat, television producer, sheriff, perfumer, automatic line operation technician, editor-in-chief, mahout, draper, chemist, nurse practitioner, official, Hofmeister, traffic guard, application programmer, anesthesia technician assistant, fighter pilot, video journalist, apologist, paralegal, pharmacy personal, arabist, bounty hunter, application expert, seiyū, chief operating officer, ghostwriter, Copywriting, Hafiz, graphic designer, association football manager, war artist, library reference desk, bertsolari, occupational therapist, succentor, program director, architectural model builder, archivist, peddler, musician, illustrator, engineer of roads and bridges, hygienist, executor, extra, letterer, expert witness, glazier, bookmaker, equites, creative director, comptroller, mime artist, curator, driving instructor, guard, financial adviser, military bishop, drafter, salesperson, book design, stamp dealer, tax advisor, engineering technologist, automatician, concertmaster, claims adjuster, lineworker, miller, farrier, Kapo, commercial agent, art director, auto mechanic, chief of staff, Betel nut beauty, performing artist, lady-in-waiting, comics artist, official, referee, manager, independent financial adviser, prelate, laic, balloonist, Kohen, test pilot, radio officer, monk, conferencier, asphalt constructor, clerk, poster artist, platelayer, assistant for automation and computer technology, Apprentice in Electronics and Computer Science, assistant for health tourism, assistant for media technology, paratrooper, doula, astrophysicist, songwriter, respiratory therapist, storyteller, automatic installer, elevator mechanic, personal trainer, luthier, dom kapellmeister, breakdancer, arborist, oblate, art exhibition curator, personal shopper, machine expert, automobile salesperson, aoidos, badminton referee, ballet dancer, ballet master, bandleader, banker, bank teller, bartender, construction worker, site manager, nursery gardener, Baumwart, unloader, call girl, lighting master, project manager, mountain guide, miner, Bergmeister, ocularist, lyricist, bus driver, vocational teacher, career soldier, occupational physician, business manager, maid, butler, legal agent, welder, brewer, sporting director, script doctor, Metropolitan and Abbot, detective, hardware architect, futurist, jockey, ship captain, bodyguard, bululu, beekeeper, tinsmith, page, day trader, security guard, beer sommelier, association football referee, picture editor, case manager, air traffic controller, game programmer, biographer, biologist, biotechnologist, supermodel, baseball coach, tenant farmer, insurance broker in Austria, contributing editor, tiler, computer forensics, whitesmith, municipal clerk, law clerk, piano maker, bladesmith, Floorer, winegrower, violist, tutor, geographer, marine chemist, gyōji, business magnate, socionom, data entry clerk, clinical nurse specialist, Mistress of the Robes, carder, Posek, make-up artist, nanny, association football player, systems analyst, impresario, skald, television presenter, character actor, manga gensakusha, parson, Liedermacher, photojournalist, business broker, Da'i, child actor, educator, basket weaver, flag officer, executive, Grandmaster, director of audiography, money changer, well builder, delegate, Bridge tender, bookseller, illuminator, middle management, Meshimori onna, business analyst, Calator, call centre agent, painter, special education teacher, absentee landlord, tour guide, miko, Net idol, general director, motorcycle courier, Commissioner of Baseball, statutory auditor, cellarer, deputy, subdeacon, Certified Public Accountant, double agent, private investigator, Chanteuse, character designer, signalman, hazzan, managing editor, Chief Officer, keyboardist, millionaire, choir director, database administrator, columnist, financial planner, building superintendent, personal assistant, scuba diving, steeplejack, chartered accountant, colorist, climatologist, clinical psychologist, cartoonist, lumberjack, postdoctoral researcher, Mandarin, sporting director, Buddhist Abbot, Dragoman, seneschal, tax collector, creative producer, crooner, stripper, Csikós, glossator, patent attorney, pilot in command, fireman, impressionist, systems integrator, inker, kunoichi, weather presenter, food taster, correspondent, Dabbawala, director, Daoshi, data steward, muhaddith, Queen's Guard, decoration painter, toastmaster, Taikomochi, logothete, flatulist, spin doctor, university chaplain, switchboard operator, information scientist, senior lecturer, Landscape contracting, instrumentalist, mortgage broker, archdeaconry in Protestantism, polychromer, stonesetter, sculptor, gemcutter, hieromonk, page-turner, Dichter, Primicerius, music artist, artisan, dubbing dramaturge, dubbing director, handball referee, buyer, nähkne \\
\bottomrule
\end{tabular}
\caption{Instances from the hyponyms of \textbf{Job} used to construct \textit{NERsocial}. }
\label{Table:listOfHyponymsJob} 
\end{table*}
\begin{table*}[!t]
\centering
\scriptsize
\begin{tabular}{p{0.95\textwidth}}  %
\toprule
\textbf{Examples of \textbf{Pet} hyponyms from Wikidata.} \\
\midrule
CC, Muezza, Orangey, Mrs. Chippy, Catmando, Tama, Dewey Readmore Books, Socks, F. D. C. Willard, Nora, Hodge, Humphrey, Towser, Wilberforce, India, Simon, Larry, Winnie, Creme Puff, Sybil, Maru, Mr. Green Genes, Trim, Scarlett, Casper, Choupette, Dusty the Klepto Kitty, Fred the Undercover Kitty, Freya, Heed, Henri, Little Nicky, Luna the Fashion Kitty, Meow, Mike, Mr. Nuts, Oscar, Peter, the Lord cat, Prince Chunk, Room 8, Rusik, Scarlett Magic, Sockington, Stubbs, Grumpy Cat, Tiddles, Shironeko, Panther, Willow, Tsim Tung Brother Cream, Misuke, Emily, Kotora, Félicette, Lil Bub, Buurtpoes Bledder, Munich Mouser, Nelson, Peta, Peter III, Treasury Bill, Hamish McHamish, Colonel Meow, Dorofei, Smudge, Spangles, Poppy, Lorenzo the cat, Tara, Pepper the Cat, Ketzel, Hana, Filuś, Frank and Louie, FamousNiki, Tiffany Two, Corduroy, Sammy, Merlin, Bébert, Matroska, Freddy, Think Think, Scooter, Tibs the Great, Ah Tsai, Palmerston, Tomba, Pixel, Gladstone, Sister Cream, Bob, Musashi, Leo, Luca, Seri, Marble, Tombili, Nala, Mr. White, Gli, Orlando, Kiisu Miisu, Paddles, Hamilton, Toffee, Nutmeg, Liv, Crimean Tom, Peter II, Peter, Arcturus Aldebaran Powers, Nitama, Foss, Mostik, Rubble, Garfi, Brigadier Broccoli, Beerbohm, Moka, Blackie, Longcat, Sam the Cat has Eyebrows, Lóu lóu, Trick, Şero, Naro, Embassy Cat, Kasper, Jackie the Cat, Fatso the Cat, Dotty the catt, Souris Calle, Remy, Reggie, Gonzo, Archie, Rizzo, Mittens, kitten, Egyptian Mau, Donskoy cat, Serengeti cat, Toyger, Cheetoh, Dragon Li, Kurilian Bobtail, American Bobtail, Abyssinian, Aegean cat, American Curl, American Shorthair, American Wirehair, Antipodean, Arabian Mau, Asian, Asian Semi-longhair, Australian Mist, Balinese, Bambino, domestic short-haired cat, European Shorthair, Burmilla, British Longhair, Brazilian Shorthair, British Semi-longhair, Anatoli, domestic long-haired cat, German Rex, Exotic Shorthair, California Spangled, Cornish Rex, Birman cat, Bombay, Devon Rex, Bengal cat, Chartreux, Havana Brown, Maine Coon, Norwegian Forest Cat, manx cat, Ragdoll, Turkish Angora, Himalayan, Ceylon cat, black cat, tabby cat, stray cat, calico cat, feral cat, polydactyl cat, deaf white cat, ship's cat, therapy cat, Cyprus cat, Van cat, Hairless Cats, library cat, Tennessee Rex, Tibetan cat, Siamese, Persian, Maine Coon, British Shorthair, Scottish Fold, Sphynx, Ragdoll, Bengal, Russian Blue, Abyssinian, Birman, Burmese, Himalayan, Exotic Shorthair, American Shorthair, Devon Rex, Cornish Rex, Manx, Norwegian Forest Cat, Turkish Angora, Labrador Retriever, German Shepherd, Golden Retriever, French Bulldog, Bulldog, Beagle, Poodle, Rottweiler, Boxer, Siberian Husky, Chihuahua, Great Dane, Doberman Pinscher, Australian Shepherd, Yorkshire Terrier, Cavalier King Charles Spaniel, Shih Tzu, Pug, Border Collie, Dachshund, Goldfish, Betta, Neon Tetra, Guppy, Angelfish, Zebra Danio, Discus, Platy, Molly, Swordtail, Koi, Oscar, Plecostomus, Clownfish, Cichlid, Catfish, Barb, Rainbowfish, Cory, Gourami, Bearded Dragon, Leopard Gecko, Ball Python, Corn Snake, Red-Eared Slider, Box Turtle, Green Iguana, Blue-Tongued Skink, Chameleon, Tokay Gecko, Savannah Monitor, Tegu, Uromastyx, Kingsnake, Boa Constrictor, African Spurred Tortoise, Painted Turtle, Green Tree Python, Crested Gecko, Gargoyle Gecko, Budgerigar, Cockatiel, African Grey Parrot, Amazon Parrot, Macaw, Conure, Lovebird, Eclectus Parrot, Cockatoo, Quaker Parrot, Parrotlet, Pionus Parrot, Rosella, Caique, Parakeet, Ringneck Parrot, Lorikeet, Toucan, Dove, Finch, Bald Eagle, \\
\bottomrule
\end{tabular}
\caption{Instances from the hyponyms of \textbf{Pet} used to construct \textit{NERsocial}. }
\label{Table:listOfHyponymsPet} 
\end{table*}
\begin{table*}[!t]
\centering
\scriptsize
\begin{tabular}{p{0.95\textwidth}}  %
\toprule
\textbf{Examples of \textbf{Sport} hyponyms from Wikidata.} \\
\midrule
Golf, Basketball, Tennis, Cricket, Boxing, Hockey, Baseball, Wrestling, Chess, Swimming, Volleyball, Rugby, Athletics, Bowling, Figure skating, Gymnastics, Ice Hockey, Table Tennis, Polo, American Football, Softball, Cross Country, Surfing, Diving, Sprint Running, Sailing, Archery, Dressage, Motorcycle racing, Horse Racing, Badminton, Karate, Skeleton Sport, Triathlon, Kickboxing, Motocross, Judo, Taekwondo, Shunty, Fencing, Lacrosse, Snooker,  Rowing, Snowboarding, Weightlifting, Futsal, Squash, Handball, Target Shooting, Alpine Skiing, Australian Football, Biathlon, BMX, Bobsleigh, Canoeing, Carom Billiards, Checkers, Clay pigeon shooting, Cross-Country Skiing, Curling, Cyclo-cross, Decathlon, Equitation, Formula racing, Hapkido, Harness Racing, Hurdles, Hurling, Jujutsu, Kart racing, Kayak, Korfball, Luge, Marathon Running, Modern Pentathlon, Mountain Biking, Netball, Padel, Pelota, Pool Billiards, Racketlon, Rallycross, Road Cycling, Rock Climbing, Short Track, Skateboarding, Ski Jumping, Speed skating, Sumo, Track Cycling, Trampolining, Water Polo, Windsurfing, high kick, cheerleading, poling, pole sports, Inuit one foot high kick, one foot high kick, two foot high kick, Alaskan high kick, winter guard, color guards, Competition aerobatics, Cluster ballooning, hopper ballooning, gliding, hang gliding, human powered aircraft, Parachuting, human skydiving, cargo parachuting, BASE jumping, skysurfing, kin ball, Newcomb ball, beach volleyball, hooverball, quidditch, yukigassen, skibobbing, aikido, jujutsu, judo, sambo, sumo, kickboxing, Mixed martial arts, MMA, rowing, women rowing, sculling, swimming, motocross, baseball, modern baseball, basketball, lacrosse, Table squash, Triple jump, Shooting sports, wheelchair racing,  Aerial dance, rally raid, playboating, synchronized diving, Hornussen, bull-leaping, extreme ironing, tamburello, Camel racing, Buzkashi, motorcycle speedway, Ancient Greek boxing, Formula 5000, Chinlone, Woodball, artistic fencing, Indiaca, Paralympic association football, bodyboarding, powerbocking, Water Jousting, Professional baseball, Brännboll, inline hockey, Canicross, land sailing, Tractor pulling, wheelchair hockey, Grand Prix motor racing, Ice climbing, Jugger, Combined driving, roller soccer, street hockey, Tent pegging, Juego del Palo, Camel wrestling, knife throwing, Varzesh-e Bastani, pato, endurance racing, team pursuit, Calcio Fiorentino, Pencak Silat, Paralympic Judo, Belt wrestling, Ice canoeing, Shitō-ryū, sidecar rally, marathon speed skating, Hamster racing, pushball, Calva, Valencian pilota, Pole climbing, professional boxing, Ringo, iron arm, Jallikattu, Fiorentina Waterpolo, historical medieval battles, Pond hockey, Hōlua, Waveski, Medieval football, TeamGym, Tour skating, kho kho, Wallyball, Khmer traditional wrestling, hunting in Italy, Fitbox, Mondio Ring, pallone col bracciale, Free flight, FIA Formula 4, blind soccer, angleball, Kırkpınar Oil Wrestling Tournament, Australian handball, blind cricket, Blue, Egham Regatta, equitation, FIK BFG Fana, flickerball, Foot hockey, Footbag net, recreational fishing, Rally de La Nucía-Mediterráneo - Trofeo Costa Blanca, pigeon racing, Locksport, Logrolling, Minkey, Para-equestrian, Pedestrianism, Trugo, Workers' sport, indoor athletics, Neppis, Kemari, Pala (pelota game), Pacu jawi, fast draw, snatch, sport in Yugoslavia, Elle (sport), throwing the stone, gukgung, Fenerbahçe Swimming, Peteca, Solar car racing, Vert skating, Goat racing, ufoball, crossbow shooting, Ice derby, paralympic cycling, Catalan bowling, Pigeon-shooting, Handball at the 2016 South Asian Games, automobile endurance racing, Ultra-distance cycling, Frontball, ice sport, Moto-Football in Greece, Foam Tennis, Gena, One leg rowing, sand skiing, kite flying, Nordic shooting with cross-country running, Road rallying, dirt track motorcycle racing, Chakuna, World Sports Network (WSN), motorsport, swimming, dog sports, gymnastics, running, archery, Kroatischer Sport, equestrianism, winter sport, shooting sport, kung fu, Olympic sport, team sport, recreational fishing, esports, extreme sport, strength sport, Hybrid sport, parkour, recreational sport, Tag rugby, Diving platform, paralympic sports, youth sports, workers' sport in Germany, Basque rural sports, air sports, dancesport, axe throwing, disabled sport, mountain sport, ball game, equestrian sport, boardsport, women's sports, casting, outdoor recreation, cat agility, national sport, ballroom dance, Francombat, chariot racing, mind sport, slacklining, professional sports, Fire sport, Fierljeppen, ferreting, fun sport, knife throwing, competitive sport, flowriding, mixed-sex sports, Kronum, cycle sport, contact sport, sport in 's-Hertogenbosch, racket sport, sport in Dordrecht, agility sport, Rabbiting, Sheaf toss, traditional Breton games, cue sports, outdoor sports, sport in ancient Greece, sports in Detroit, recreational mountaineering, Surr, sport in ancient Rome, spectator sport, sports in Moscow Oblast, adventure sport, military sports, net sport, sports in New York, sports in Seattle, indoor athletics, deaf sport, Taigi, Freestyle relay, athletic culture, Truck pulling, hiking, trekking, beach sport, street workout, skating, endurance sport, skiing, throwing sport, amateur sports, risk sport, international sport, Drone racing, sport in a geographic region, men's sports, multisport sport, mixed sports discipline, RESEE, summer sport, Stijldans \\
\bottomrule
\end{tabular}
\caption{Instances from the hyponyms of \textbf{Sport} used to construct \textit{NERsocial}. }
\label{Table:listOfHyponymsSport} 
\end{table*}

\end{document}